\documentclass[sn-nature]{sn-jnl}

\usepackage{microtype}
\usepackage{caption}
\usepackage[labelformat=simple] {subcaption}
\usepackage{hyperref}

\usepackage{graphicx}%
\usepackage{multirow}%
\usepackage{amsmath,amssymb,amsfonts}%
\usepackage{amsthm}%
\usepackage{mathrsfs}%
\usepackage[title]{appendix}%
\usepackage{xcolor}%
\usepackage{textcomp}%
\usepackage{manyfoot}%
\usepackage{booktabs}%
\usepackage{algorithm}%
\usepackage{algorithmicx}%
\usepackage{algpseudocode}%
\usepackage{listings}%
\usepackage{float}%
\usepackage{placeins}
\usepackage{amsmath}
\usepackage{amssymb}
\usepackage{mathtools}
\usepackage{amsthm}

\usepackage[capitalize,noabbrev]{cleveref}

\theoremstyle{plain}
\newtheorem{theorem}{Theorem}[section]

\theoremstyle{definition}

\theoremstyle{remark}

\usepackage[utf8]{inputenc} 
\usepackage[T1]{fontenc}    
\usepackage{hyperref}       
\usepackage{url}            
\usepackage{booktabs}       
\usepackage{amsfonts}       
\usepackage{nicefrac}       
\usepackage{microtype}      
\usepackage{xcolor}         
\usepackage{comment}
\usepackage{amsbsy,amsfonts,amsmath,amsthm,commath,microtype,dsfont}
\usepackage{graphicx}
\usepackage[colorinlistoftodos,textsize=tiny]{todonotes}
\usepackage{cleveref}

\usepackage{enumitem}
\usepackage{mathtools}

\usepackage{lineno}

\newcommand\E{\mathbb{E}} 
\renewcommand\P{\mathbb{P}} 
\newcommand\R{\mathbb{R}} 
\newcommand\N{\mathbb{N}} 
\newcommand\argmin{\operatorname*{\arg\min}} 
\newcommand\ind[1]{\mathds{1}_{\{#1\}}} 
\usepackage{xspace}

\newcommand\risk{\mathcal{R}_{0/1}}
\newcommand\mse{\mathcal{R}_{\operatorname{sq}}}
\newcommand\eq[1]{Eq.~(\ref{#1})}

\begin{document}

\title[AI without networks]{AI without networks}


\author*[1]{\fnm{Partha P.} \sur{Mitra}}\email{parthaxmitra@gmail.com}

\author*[2]{\fnm{Cl\'ement} \sur{Sire}}\email{clement.sire@univ-tlse3.fr}

\affil*[1]{\orgname{Cold Spring Harbor Laboratory}, \orgaddress{\street{1, Bungtown Road}, \city{Cold Spring Harbor}, \postcode{11724}, \state{NY}, \country{USA}}}

\affil*[2]{\orgdiv{Laboratoire de Physique Th\'eorique}, \orgname{CNRS \& Universit\'e de Toulouse III -- Paul Sabatier}, \city{Toulouse}, \postcode{31062}, \country{France}}


\abstract{Contemporary Artificial Intelligence (AI) stands on two legs: large training data corpora and many-parameter artificial neural networks (ANNs). The data corpora are needed to represent the complexity and heterogeneity of the world. The role of the networks is less transparent due to the obscure dependence of the network parameters and outputs on the training data and inputs. This raises problems, ranging from technical-scientific to legal-ethical. We hypothesize that a transparent approach to machine learning is possible without using networks at all. By generalizing a parameter-free, statistically consistent data interpolation method, which we analyze theoretically in detail, we develop a network-free framework for AI incorporating generative modeling. We demonstrate this framework with examples from three different disciplines - ethology, control theory, and mathematics. Our generative Hilbert framework applied to the trajectories of small groups of swimming fish outperformed state-of-the-art traditional mathematical behavioral models and current ANN-based models. We demonstrate pure data interpolation based control by stabilizing an inverted pendulum and a driven logistic map around unstable fixed points. Finally, we present a mathematical application by predicting zeros of the Riemann Zeta function, achieving comparable performance as a transformer network. We do not suggest that the proposed framework will always outperform networks as over-parameterized networks can interpolate. However, our framework is theoretically sound, transparent, deterministic, and parameter free: remarkably, it does not require any compute-expensive training, does not involve optimization, has no model selection, and is easily reproduced and ported. We also propose an easily computed method of credit assignment based on this framework, to help address ethical-legal challenges raised by generative AI.}

\keywords{Statistically Consistent Interpolation, Hilbert Kernel, Behavioral Modeling}



\maketitle

Contemporary AI continues to have a large impact across many domains of activity~\cite{lecun2015deep}, encompassing technological, scientific, and creative domains. While the applications are diverse in computational architecture, they are all based on two basic elements: \textit{(i)}~functions approximated by artificial neural networks, which are nonlinearly parameterized by many parameters~\cite{Goodfellow-et-al-2016}, and \textit{(ii)}~large sets of data relevant to the application domain, collected from the real world~\cite{zhao2023survey} or simulated computationally~\cite{silver2017mastering}. Although these applications have been pragmatically successful in specific domains due to high performance in certain complex tasks~\cite{brown2020language,silver2017mastering,jumper2021highly}, there are significant challenges~\cite{noauthor_ai_2023}. These stem from the lack of transparency and interpretability of the networks~\cite{rudin2019stop}, particularly important in scientific~\cite{van2023ai} or biomedical applications~\cite{world2021ethics,babic2021beware,shad2021designing}, as well as in ethical-legal issues related to authorship and ownership of the output of generative AI~\cite{cohan23}. 
Other related problems are difficulties in reproducing results and porting code~\cite{hutson2018artificial}, due to stochastic elements in training and sensitivity to network details.   

These challenges arise fundamentally from the non-transparent, nonlinear relation between the data corpora and the network parameters, determined using various online and offline optimization procedures (learning algorithms). The centrality of this parameter optimization step is clear -- the overall subject area is called machine learning. Additionally, the multi-layer nonlinear parameterization of the functions involved means that the input-output relationships of the optimized (trained) networks are also non-transparent. This has led to attempts to develop interpretable network architectures~\cite{rudin2022interpretable}, but ANNs are complex non-linear parameterizations and will generally not bear a transparent relation to the training data. One possible solution is offered by recent work~\cite{belkin2018overfitting,belkin2018reconciling,mitra_fitting_2021} showing that the networks may be acting as data interpolation devices, thus prioritizing the data corpora over the networks. However, recent work in this area has focused on conditions under which the data interpolation provides good performance (“benign interpolation”~\cite{bartlett2020benign}), rather than solving the issues above. We suggest that a fundamentally transparent approach to machine intelligence is possible for sufficiently large data corpora that eliminates the networks entirely, thus cutting the Gordian knot of interpreting the network parameters and eliminating conceptual confusions that may arise from the complexity of the networks. This radically empirical approach -- all data and no model -- also automatically solves the problem of understanding network outputs in terms of the inputs and facilitates reproducibility and portability. 

The basis of our approach is an interpolation methodology, which remarkably has no tunable parameters. The Hilbert kernel interpolator~\cite{devroye1998hilbert}, a classical Nadaraya-Watson~\cite{nadaraya1964estimating} regression scheme using a singular weight function, was previously shown to be statistically consistent~\cite{devroye1998hilbert} (producing Bayes optimal results for large data sets), but has not been widely used. We show that this interpolator belongs to a large class of interpolation methods, which we term the generalized Hilbert kernel. We perform a rigorous and detailed theoretical analysis, including a closed form analytical expression for the asymptotic convergence rate, which we show is strikingly independent of data dimension. We propose a general framework, based on the generalized Hilbert kernel and inclusive of generative modeling, and apply it to problems of a scientific and theoretical nature from three disciplines, ethology, control theory, and mathematics. In a study of collective animal behavior, the framework applied to groups of 2 and 5 rummy-nose tetra fish outperforms both the current best data-driven behavioral models~\cite{calovi2014swarming,calovi2018disentangling,escobedo2020data} and ANN models~\cite{papaspyros2023biohybrid,papaspyros2023ML} for this species. In a control theory example, we replace the system model normally used in a model-predictive control (MPC) \cite{camacho2007constrained,richalet1978model}
framework with a generalized Hilbert predictor, to control an inverted pendulum as well as a chaotic map. Finally, the framework is used to predict zeros of the Riemann zeta function, with similar performance as a transformer trained on the same task,  both predictors having presumably reached the Bayes limit. 

While we demonstrate good performance in these applications, this is not our main take-home message, as performance is not the only criterion with which to evaluate such an approach, particularly in scientific and biomedical applications where understandability, transparency, and theoretical grounding is of great importance. Overparameterized networks can interpolate and may also provide good performance. There is however a major difference: the Hilbert approach involves no learning, in contrast with all network-based AI. The Hilbert framework resolves the issues of transparency, reproducibility, and credit assignment, while demonstrating performance in scientifically sound settings. The framework is theoretically well-grounded, deterministic and parameter free: it does not require compute-expensive training, does not involve stochastic optimization, does not include model selection, and is easy to reproduce. These are critical considerations for scientific applications of machine learning, which we consider our primary target in this work, although we expect that the utility of the approach will extend to engineering applications as well, particularly given legal issues surrounding copyright and data provenance. The Hilbert generative framework also shows interesting phenomena such as copying portions of training data, which have been recently observed in the context of Large Language Models~\cite{nasr2023scalable}, and provides a theoretical basis for understanding such behavior.

We first discuss Statistically Consistent Interpolation (SCI) and present a summary of our analytical and rigorous results for the behavior of the Hilbert kernel interpolator, together with numerical illustrations of selected results. Next, we propose a framework for interpolative generative and system modeling of scalar or vector time series, using the generalized Hilbert kernel. We present three applications of this framework. The first application is to the collective behavior of groups of 2 and 5 rummy-nose tetra fish, for an in-depth demonstration of utility and performance in the wide context of collective animal phenomena. The second application is to control theory and engineering, where we demonstrate performance in a canonical example, the stabilization of an inverted pendulum, using a Hilbert predictor to replace the system model. The third application is to a mathematics problem, where we exploit the Hilbert scheme to predict the zeros of the Riemann zeta function, and compare its performance to that of a transformer network trained on the same task. 

We discuss the ramifications of our results, for the applications addressed here, as well as in more general contexts. Finally, we propose an approach for credit assignment that helps address ethical-legal challenges raised by generative AI and has implications for copyrighted training data, and discuss other ramifications of a network-free, data interpolation-based transparent approach to AI. 

\section*{Results and discussion}\label{sec2}

\subsection*{Statistically consistent interpolation and the Hilbert kernel}\label{subsec2}

Data interpolation and statistical regression are both classical subjects, but have been largely disjoint until recently. Scattered data interpolation techniques~\cite{wendland2004scattered} are generally used for clean data. On the other hand, when supervised learning or statistical regression techniques are applied to noisy data, smoothing or regularization methods are usually applied to prevent training data interpolation, as the latter is believed to lead to poor generalization \cite{james2013introduction}. However, empirical evidence from overparameterized deep networks has shown that data interpolation (equivalently, zero error on the training set) does not automatically imply poor generalization \cite{zhang2017understanding,belkin2018understand}. This has in turn given rise to much theoretical work to understand how and why noisy data interpolation can still lead to good generalization \cite{cutler2001pert,wyner2017explaining,belkin2018overfitting,rakhlin2018consistency,ongie2019function,belkin2019reconciling,liang2018just,bartlett2019benign,montanari2019generalization,karzand2019active,xing2018statistical}. Note, however, that this body of work is largely within the framework of parameterized networks, linear or nonlinear, and generally involves a model fitting/selection step, in contrast with our network and parameter-free approach. 

A key notion in this regard is the phenomenon of statistically consistent interpolation (SCI) \cite{mitra_fitting_2021}, {i.e.}, regression function estimation that interpolates training data but also generalizes as well as possible by achieving the Bayes limit for expected generalization error (risk), when the sample size becomes large. The Hilbert kernel \cite{devroye1998hilbert} is a Nadaraya-Watson style estimator \cite{nadaraya1964estimating,watson1964smooth} with the unique property that it is fully parameter-free and hence does not have any bandwidth or scale parameter. It is global and uses all or batches of the data points for each estimate: the associated kernel is a power law and thus scale-free. Although statistical consistency  of this estimator for $d$-dimensional input data was proven when it was proposed~\cite{devroye1998hilbert}, there has been no systematic analysis of the associated convergence rates and asymptotic finite sample behavior, and it is generally not used in applications. We show that the Hilbert kernel is a member of a large class of SCI methods that have the same asymptotic convergence rates. 

The only other interpolation scheme we are aware of, that is proven to be statistically consistent in arbitrary dimensions under general conditions, is the weighted interpolating nearest neighbors method (wiNN)~\cite{belkin2018overfitting}, which is also a NW estimator of a similar form but with two important differences: a finite number of neighbors $k$ is utilized (rather than all data points), and the power-law exponent $\delta$ of the NW kernel is a tunable parameter satisfying $0<\delta<d/2$, rather than the fixed value $\delta=d$ for the Hilbert kernel. To achieve consistency, $k$ has to scale appropriately with sample size. Despite the superficial resemblance, the wiNN and the generalized Hilbert Kernel estimators have quite different convergence rates. In scattered data interpolation~\cite{wendland2004scattered}, the focus is generally on the approximation error (corresponding to the “bias'' term in our analysis below). Interestingly, the risk for generalized Hilbert kernel interpolation is dominated by the noise or “variance” term in the asymptotic regime. For wiNN, note that the bias-variance tradeoff is still active and determines the optimal choice of $k$ as a function of the number of data. In contrast with wiNN or Hilbert kernel interpolation, other interpolating learning methods, such as simplex interpolation \cite{belkin2018overfitting} or ridgeless kernel regression~\cite{liang2018just}, are generally not statistically consistent in fixed finite dimension~\cite{rakhlin2018consistency}.

Given a dataset $\{x_i,y_i,~i=0,...,n\}$, the Hilbert kernel estimator $\hat{f}(x)$ of the regression function $f(x)$ (see Methods for problem setup and notation definitions) is a Nadaraya-Watson style estimator employing a singular kernel, 
\begin{eqnarray}
    \hat{f}(x) &=& \sum_{i=0}^n w_i(x)y_i,  \label{WeightDef}\\
    w_i(x) &=& \frac{\|x-x_i\|^{-d}}{\sum_{j=0}^n \|x-x_j\|^{-d}},
    \label{HilbertDef}
\end{eqnarray}
where $w_i(x)$ is the weight of the $i$-th data in the prediction $\hat{f}(x)$, for a given $d$-dimensional input $x$. Note that when $x=x_i$ is an input vector from the dataset, $w_i(x=x_i)=1$, $w_j(x=x_i)=0$, for all $j\ne i$, and the prediction exactly coincides with the dataset result, $\hat{f}(x=x_i) =y_i$ (true interpolation).

\begin{figure}[h]
     \centering
     \begin{subfigure}[t]{0.31\textwidth}
        \includegraphics[width=\textwidth]{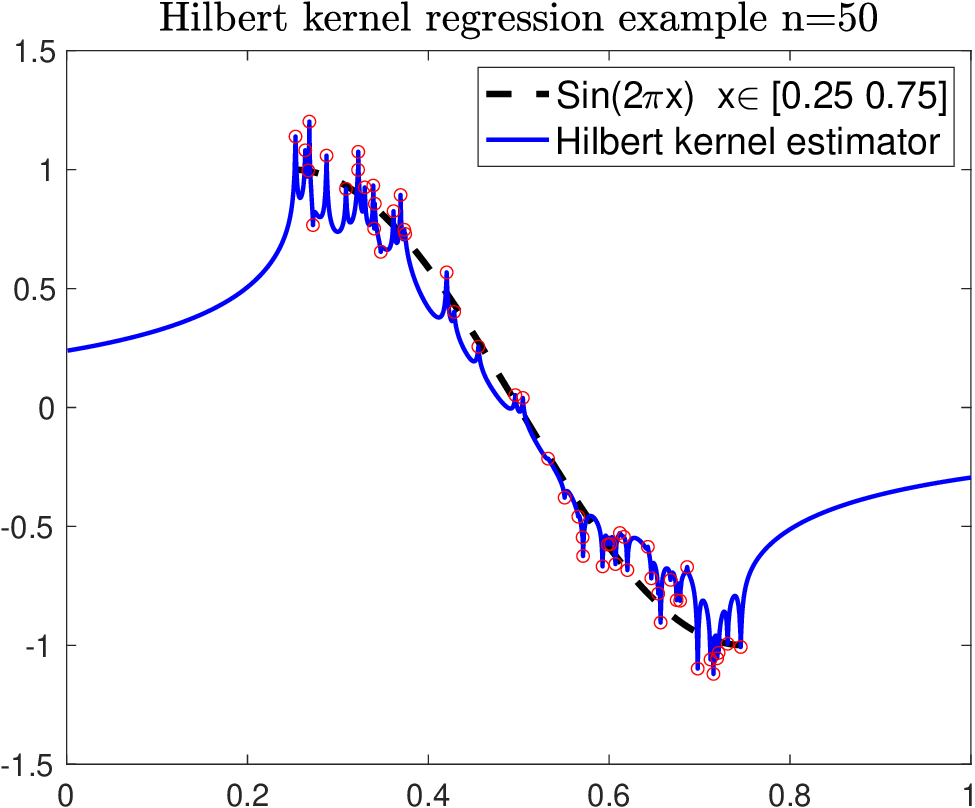}
         \caption{\small Hilbert regression}
         \label{fighilbert}
     \end{subfigure}
     \hfill
     \begin{subfigure}[t]{0.34\textwidth}
         \includegraphics[width=\textwidth]{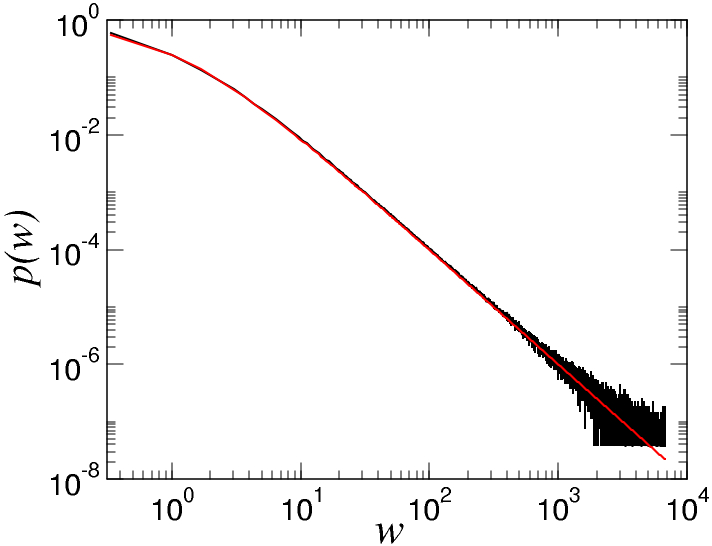}
         \caption{\small Distribution of weights}
         \label{figdist}
     \end{subfigure}
     \hfill
     \begin{subfigure}[t]{0.31\textwidth}
         \includegraphics[width=\textwidth]{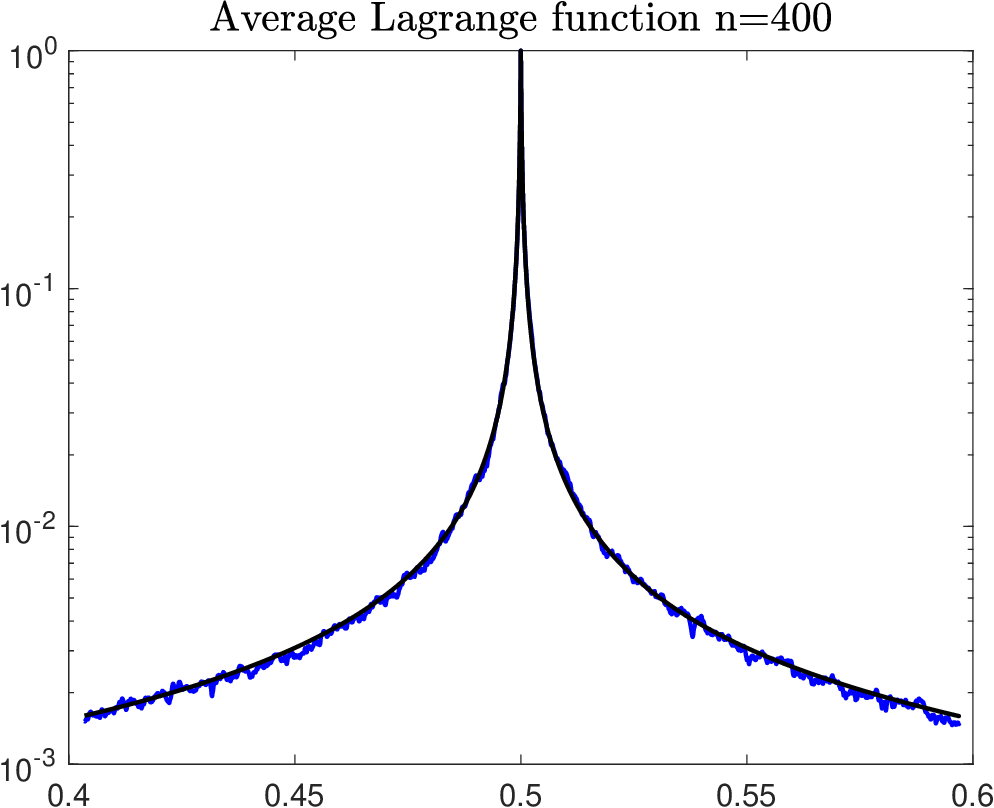}
         \caption{\small Lagrange function}
         \label{fighilbertlag}
     \end{subfigure}
        \caption{\small \textbf{Illustration of the Hilbert interpolation scheme in one dimension.} {\bf a},~an example is shown of the Hilbert kernel regression estimator in $d=1$, both within and outside the input data domain. A total of 50 samples $x_i$ were chosen uniformly distributed in the interval $[0.25 ~~0.75]$ and $y_i=\sin(2\pi x_i)+n_i$ with the noise $n_i$ chosen {\it i.i.d.} Gaussian distributed $\sim N(0,0.1)$. The sample points are circled, and the function $\sin(2\pi x)$ is shown with a dashed line within the data domain. The solid line is the Hilbert kernel regression estimator. Note the interpolation behavior within the data domain and the extrapolation behavior outside the data domain. {\bf b}, we plot the results of numerical simulations for the distribution $p$ of the scaling variable $w=\frac{W}{W_n}$, with
		$W_n\approx\frac{1}{n\ln(n\ln(n))}$, and for $n=65536$ (black line). This is compared to $p(w)=\frac{1}{(1+w)^2}$ (red line), which has the predicted universal tail $p(w)\sim w^{-2}$ for large $w$. {\bf c}, numerical simulation of the expected value of the Lagrange function of the Hilbert kernel regression estimator in one dimension for a uniform distribution, as in Fig.~\ref{fighilbert}. $n=400$ samples $x_i$ were chosen uniformly distributed in the interval $[0,1]$ for 100 repeats, and the Lagrange function evaluated at $x_0=0.5$ was averaged across these 100 repeats (blue curve). The black curve shows the asymptotic form $(1+Z)^{-1}$ with $Z=2|x-x_0|/W_n$. 
        }
        \label{fig:three graphs}
\end{figure}

We show in Fig.~\ref{fighilbert} an example of the Hilbert kernel regression estimator in one dimension, $d=1$ (an application to classification by thresholding a soft classifier is shown in Extended Data Fig.~\ref{fighilbertclassification}). Both the bias and the variance of the estimator can be visually seen, as well as the extrapolation behavior outside of the data domain. Note that in higher dimensions, the sharp peaks would have rounded tops, which increasingly flatten with dimension.

A generalized Hilbert kernel can be obtained by a change in coordinates, $x\rightarrow \Phi(x)$, where $\Phi$ is an arbitrary diffeomorphism. This results in a large class of weighting schemes for the input data, each of which produces a statistically consistent estimate of the regression function with the same asymptotic convergence rates (see Extended Data Fig.~\ref{extfig:coordinatetransform} for an example of the generalization using a coordinate transformation). Unlike ANNs or kernel regression, the change in coordinates does not necessitate a data-dependent parameter optimization step. Another generalization is obtained by replacing $y_i$ by a continuous function $g(y_i)$ in the estimator. This way, an estimate of $E[g(y)|x]$ may be obtained, for example moments of $y$ may be estimated in this manner. These generalizations considerably broaden the original method. 

We also propose a new leave-one-out estimator for the noise variance, which can be used to construct confidence limits to the regression function: 

\begin{eqnarray}
    \widehat{\sigma^2}(x) &=& \sum_{i=0}^n w_i(x)\widehat{\sigma^2}_i,\\
    \widehat{\sigma^2}_i &=& (y_i-\sum_{j\neq i} y_j w_{j(i)}(x_i) )^2,
    \label{HilbertVar}
\end{eqnarray}
where $w_{j(i)}(x_i)$ are weights defined like in Eq.~(\ref{HilbertDef}) but omitting the $i$-th sample.

We proved that the excess risk of the generalized Hilbert estimator decays logarithmically with the sample size $n$, under broad conditions and independently of the dimension $d$, and that the risk is dominated by the variance term in a bias-variance decomposition, with the bias term being generically sub-leading (Theorems~\ref{theovar2}-\ref{theobias}, {i.e.}, no bias-variance tradeoff). We carried out a detailed characterization of the moments of the weight functions across sample realizations and conjecture that the full distribution has a scaling form showing power-law behavior. We also characterized the extrapolation behavior of the estimator outside the training data domain. The problem setup, precise theorem statements and proof outlines are presented in the Methods section, and detailed proofs are available in the Supplementary Materials for all theorems referred to here, under the corresponding theorem numbers. 

\begin{itemize}
    \item We show (Theorem \ref{regressionrisk}) under general conditions that the excess regression risk at the point $x$ is asymptotically equivalent to the local variance $\sigma^2(x)$ reduced by the logarithm of the sample size (note: no unknown constants, no dimension dependence), 
\begin{equation}
    \E[(\hat{f}(x)-f(x))^2]\underset{n\to +\infty  }{\sim} \frac{\sigma^2(x)}{\ln(n)}.
\end{equation}
Although all $n$ samples are used in the estimate, they contribute with quite different weights. As discussed below, the effective number of degrees of freedom ({\it d.o.f.}) contributing to the estimate can be shown to be $\ln(n)$. Thus, this formula may intuitively be interpreted as a reduction of the variance by the effective {\it d.o.f.}.
\item For excess classification risk $\delta\risk(x)$ we show (Theorem \ref{theoclass}) that for any $\varepsilon>0$, there exists a constant $N_{x,\varepsilon}$, such that for any $n\geq  N_{x,\varepsilon}$, 
\begin{equation}
    0\leq \delta\risk(x)\leq 2(1+\varepsilon) \frac{\sigma(x)}{\sqrt{\ln(n)}},\\
\end{equation}
\item Where the data distribution has nonzero density, we show (Theorem \ref{theo1}) that the $\beta-$moments of the weights $\beta>1$ satisfy: 
\begin{equation}
    \E\left[w_0^\beta(x)\right]\underset{n\to +\infty  }{\sim}\frac{1}{(\beta-1) n\ln(n)}.
\end{equation}
In particular, $\E\left[w_0^2(x)\right]\sim [n\ln(n)]^{-1}$ implies $\E [\sum_i w_i^2(x)] \sim \ln(n)^{-1}$, so that the effective {\it d.o.f} defined by $\E [\sum_i w_i^2(x)]^{-1} \sim \ln(n)$. We also prove that the information entropy,
$S(x) =\E[-\sum_i w_i(x)\ln(w_i(x))] \sim \ln(\sqrt{n})$. We present heuristic arguments for a universal $w^{-2}$ power-law behavior of the probability density of the weights in the large $n$ limit (see Fig.~\ref{figdist} for numerical evidence, details in the appendix).

\item The expected value of the weight function around a fixed data point is called the Lagrange function, and plays an important role in the classical literature on interpolation. For a point $x$ within the support of the data distribution (assumed continuous), we are able to derive a limiting form of the Lagrange function (Theorem \ref{theolag}) by rescaling the Lagrange function with a local scale proportional to $[\rho(x)n\log(n)]^{-1/d}$. In this limit 
(denoted by $\lim_Z$), $n\to+\infty$, $\|x-x_0\|^{-d}\to +\infty$  (i.e., $x_0\to x$), such that $V_d\rho(x)\|x-x_0\|^d n\log(n)\to Z$, the Lagrange function $L_0(x)=\E_{X|x_0} [w_0(x)]$ converges to a proper limit (see Fig.~\ref{fighilbertlag} for a numerical simulation),
\begin{equation}
    \lim_{Z} L_0(x)= \frac{1}{1+Z}.
\end{equation}
\item Asymptotic convergence rates are left unaltered with a smooth coordinate transformation $x\rightarrow \Phi(x)$. The asymptotic form of the Lagrange function for the transformed case may be obtained using the substitutions $x\rightarrow X=\Phi(x)$ and $\rho(x)\rightarrow R(X)=\rho(\Phi^{-1}(X))|\partial x/\partial X|$ ({i.e.}, the density is transformed with the appropriate Jacobian). An optimal choice of $\Phi$, which governs subleading terms in the convergence, will depend on the details of the density $\rho(x)$ and the function $f(x)$. However, the bias term depends inversely on $\rho$ (see Eq.~(\ref{biaskne0})), so a good heuristic for the coordinate transformation would be to make the density of $X$ as uniform as possible over its support. 

\item The lack of data-dimension dependence of the asymptotic convergence rates is remarkable, compared with other estimators where asymptotic risk worsens with increasing dimensions (curse of dimensionality). We expect the performance of the Hilbert estimator to improve with increasing dimensions compared with conventional estimators. 

\item Interpolating function estimators also {\it extrapolate} outside the support of the data distribution (see also the application to fish behavior modeling below). We characterize the extrapolation behavior of the Hilbert estimator, which can be seen in Fig.~\ref{fighilbert}, in theorem \ref{theoout}. Far outside the support, the estimate tends to the average value of the function over the data distribution.  
\end{itemize}

\renewcommand{\thesubfigure}{\bf\alph{subfigure}}

A fundamental question in machine learning using parametric models such as ANNs is the model complexity, or the effective number of parameters in the model. Larger data sets may intuitively be expected to support more complex parametric models. Since the Hilbert scheme has no parameters, the usual notions of model complexity or parameter counting are irrelevant. However, one could ask how many independent sample points contributed effectively to the estimate at a given point $x$, corresponding to the notion of degrees of freedom in classical statistics. Thus, we define a heuristic, sample-dependent, and local estimate of an inverse number of degrees of freedom as, $({\it d.o.f.})^{-1}=\sum_i w_i^2(x)$. If all of the weights are the same and concentrated over ${\cal N}$ data, then $w_i=1/{\cal N}$ for these data (and 0 otherwise),  and we indeed find ${\it d.o.f.} ={\cal N}$. In particular, when only one data point contributes to the prediction, ${\it d.o.f.}\,=1$. Thus, the Hilbert scheme resembles an adaptive weighted nearest neighbor technique, with typically $\ln(n)$ nearby points contributing on average to the estimator, as defined by the inverse ${\it d.o.f.}$. Indeed, and in contrast with $k$NN-style methods, the actual number of data points contributing to an estimate can adapt to the provided input $x$. 

\subsection*{Interpolative generative modeling of time series}

Interpolative regression, {e.g.}, the Hilbert kernel estimator as defined above, applies directly to supervised learning and classification with labeled samples. Although stated for scalar $y$ or binary classification, the generalization to vectors (element-wise) is straightforward, and also to multi-class classification using a one-hot vector representation of the classes. Here, we propose applying interpolative regression to generative modeling, which is a major of area of current interest. 

Generative AI architectures contain as core components ANNs that are subject to supervised learning using labeled examples (e.g., next token prediction in language models~\cite{vaswani2017attention}, or predicting added noise in diffusion-based image generation models~\cite{dhariwal2021diffusion}). These ANNs can be replaced by Hilbert kernel interpolators (other statistically consistent interpolating learners such as the wiNN algorithm~\cite{belkin2018overfitting} could be similarly used; in this paper we focus on the Hilbert kernel). This approach eliminates highly parameterized black boxes central to the generative architectures in favor of a parameter-free, pure-data approach, that brings the important benefits of transparency and ease of credit assignment. Here, we focus on generative or predictive modeling of time series, with applications to behavioral time series in animal behavior, feedback control of nonlinear dynamical systems, as well as series constructed from Zeta function zeros, as worked examples which we can examine in detail. Applying these ideas more broadly to generative AI is a rich area for future research. 

Consider the generative modeling of time series ({e.g.}, positions and velocities of fish, outputs of a driven dynamical system, tokens in language models, etc.). Parameterized ANNs (RCNNs, LSTMs, Transformers, etc.) have been previously used to predict the next time point given a past time window in an autoregressive manner. Instead, we propose Hilbert autoregression (note that $x_i$ could be a scalar or vector)
\begin{equation}
    x_{t}= f_H(x_{t-1},x_{t-2},..,x_{i-T}),
    \label{eq:ARmodel}
\end{equation}
 where $f_H$ is a generalized Hilbert estimator of the prediction function from a lag-window of length $T$, which sets the dimension of the Hilbert estimator (more generally, a weighted average of such estimators weighted a power of $T$ could be considered). Once training data is used to establish $f_H$,  the prediction model can then be used to generate samples of the time series $x_t$ from an initial segment of length $T$. Eq.~(\ref{eq:ARmodel}) constitutes an autonomous dynamical system fully determined by the training data, and does not involve any model parameter fitting or learning, in contrast with existing approaches to autoregressive modeling. There is precedent to data-driven autoregressive prediction for nonlinear systems (e.g. \cite{sugihara1990nonlinear}), but these approaches still have adjustable parameters and do not have theoretical guarantees of statistical consistency of the prediction function. 
 
 Examples of this procedure on simple stochastic processes are shown in the Supplementary Materials. We observed steady state, periodic (Extended Data Fig.~\ref{extfig:autoregression2}), and occasionally intermittent behaviors, as would be expected from dynamical systems theory. In addition, we observed “copying” behavior, where the dynamics copying parts of the training data after an initial transient (Extended Data Fig.~\ref{extfig:autoregression1}) or as part of periodic episodes (Extended Data Fig.~\ref{extfig:autoregression2}). Such copying behavior has been reported in Large Language Models, in the context of extraction of snippets of training data using appropriate prompts~\cite{nasr2023scalable}, and is of interest in the context of “stochastic parrot” style phenomena, {i.e.} the idea that the machine learning algorithm is essentially repeating the content of the data with some added noise due to limitations of the algorithm~\cite{lindholm2022machine}. The extent of copying can be monitored using the entropy of the predictive weights as a function of time. In the real data example considered in the next section, where the training data consists of several time series fragments, we observe short episodes of near-copying behavior, but this does not dominate the generated time series. 
 
 The same procedure can also be used to predict one signal from another, or to predict the output of a driven, non-autonomous dynamical system if paired training data is available -- {e.g.}, for a pair of time series $x_t,y_t$, the predictive equation 
\begin{equation}
    y_{t}= f_H(y_{t-1},y_{t-2},..,y_{t-T};x_{t-1},x_{t-2},..,x_{i-T}),
    \label{eq:ARMAmodel}
\end{equation}
can be used, where $f_H$ is constructed from a paired training set $x_t,y_t$. During the prediction phase, only $x_t$ are observed, and the samples $y_t$ are generated recursively with a suitable initial condition. Modeling input-output relations or transfer functions is central to system identification~\cite{ljung1998system} in control theory. Thus, we expect that the Hilbert framework will be applicable to the control of complex systems where system identification using parameterized models is challenging, and a direct data-driven approach provides an alternative to using black-box ANNs for the same purpose. We illustrate this with examples below to stabilize an inverted pendulum and controlling a driven chaotic map. 


\subsection*{Hilbert generative model for realistic animal collective motion}

Collective behaviors, and in particular, collective motions, are ubiquitous in nature, and are for instance observed in bacteria colonies, insects, bird flocks, or fish shoals~\cite{camazine2001self,sumpter+2010,sumpter2006principles}.
The traditional approach to studying collective animal motion -- e.g., collective swimming of fish schools -- aims at producing data-driven mathematical models which exploit the inference of the social interactions between individuals~\cite{lukeman2010inferring,herbert2011inferring,katz2011inferring,gautrais2012deciphering,calovi2014swarming,calovi2018disentangling,escobedo2020data,lei2020computational}. Such approaches can lead to the direct implementation of the measured social interactions in analytical models, in good agreement with experiments~\cite{calovi2018disentangling,escobedo2020data,lei2020computational}. Deciphering the social interactions between individuals and their interactions with obstacles (e.g., the tank wall, in the fish context) offers in-depth understanding and direct interpretation of these social interactions and of the resulting collective behavior of fish, even when modifying the lighting of their environment~\cite{Xuefishlight2023,Lafoux2023fishlight}. As an application, the data-driven mathematical model of~\cite{calovi2018disentangling} was recently implemented to command a robot fish interacting in closed-loop with 1 or 4 real fish~\cite{papaspyros2023biohybrid}, a valuable tool to investigate the behavior of a fish group submitted to a controlled perturbation.

A more recent approach benefits from the progress in the applications of machine learning (ML). The social behavior of animal groups is directly learned by an ANN by using experimental collective trajectories as the training data~\cite{heras2018deep,cazenille2019automatic,costa2020automated,papaspyros2023ML,papaspyros2023biogap}. In~\cite{papaspyros2023ML}, a deep learning model was designed, which quantitatively reproduces the long-term statistical properties of the motion of pairs of fish. This work \cite{papaspyros2023ML} also emphasized the need to measure a set of stringent observables to properly assess the agreement between a generative model for animal trajectories and the short-term \textit{and} long-term experimental dynamics. The deep learning model was shown to be in comparable agreement with experiments as the state-of-the-art mathematical model for the considered species~\cite{calovi2018disentangling}. Moreover, the deep learning model of~\cite{papaspyros2023ML} was also applied to another fish species, with similar success, and \textit{without retraining}. This illustrates one of the main advantages of ML models, which can be reused without retraining to other species with similar motion patterns, contrary to mathematical models for which the social interactions must be specifically measured for each considered species~\cite{escobedo2020data}. On the other hand, ML models lack interpretability and do not provide detailed insight about the actual social interactions at play. In particular, and contrary to data-driven analytical models derived from the reconstruction procedure of~\cite{calovi2018disentangling,escobedo2020data}, ML models are unable to disentangle the contributions of the different interactions leading to the instantaneous behavior of an individual: attraction/repulsion and alignment with other individuals and avoidance of obstacles. Note that the ML model of~\cite{papaspyros2023ML} was also exploited to command a robot fish interacting in closed-loop with one real fish~\cite{papaspyros2023biogap}.

In the following, we show that the Hilbert interpolation scheme, exploiting experimental data for the dynamics of animal groups, constitutes a light and powerful generative model to produce realistic long-term trajectories of such groups. Hence, the Hilbert scheme provides a valuable alternative to other generative models for collective animal behavior, which we now explicitly describe in the context of the collective motion of small fish groups. 

\subsubsection*{General implementation of the Hilbert scheme in the collective behavior context}

We have implemented the Hilbert interpolation scheme as a generative model by exploiting the experimental trajectories for $N=2$ and $N=5$ rummy-nose tetra fish (\textit{Hemigrammus rhodostomus}) using previously published data~\cite{papaspyros2023biohybrid,papaspyros2023biogap}. Adult rummy-nose tetra, of typical body length 3--3.5\,cm are social fish swimming in groups and can exhibit a highly coordinated behavior: highly polarized schools or vortex/milling schools. 

We used 2.6 hours of data for $N=2$ fish (18 experiments with different pairs of fish), and 9.4 hours of data for $N=5$ fish (20 experiments), with all fish swimming in a circular arena of radius $R=25$\,cm. Details about the data set are provided in the Methods section.


\begin{figure}[ht]
     \centering
        \includegraphics[width=\textwidth]{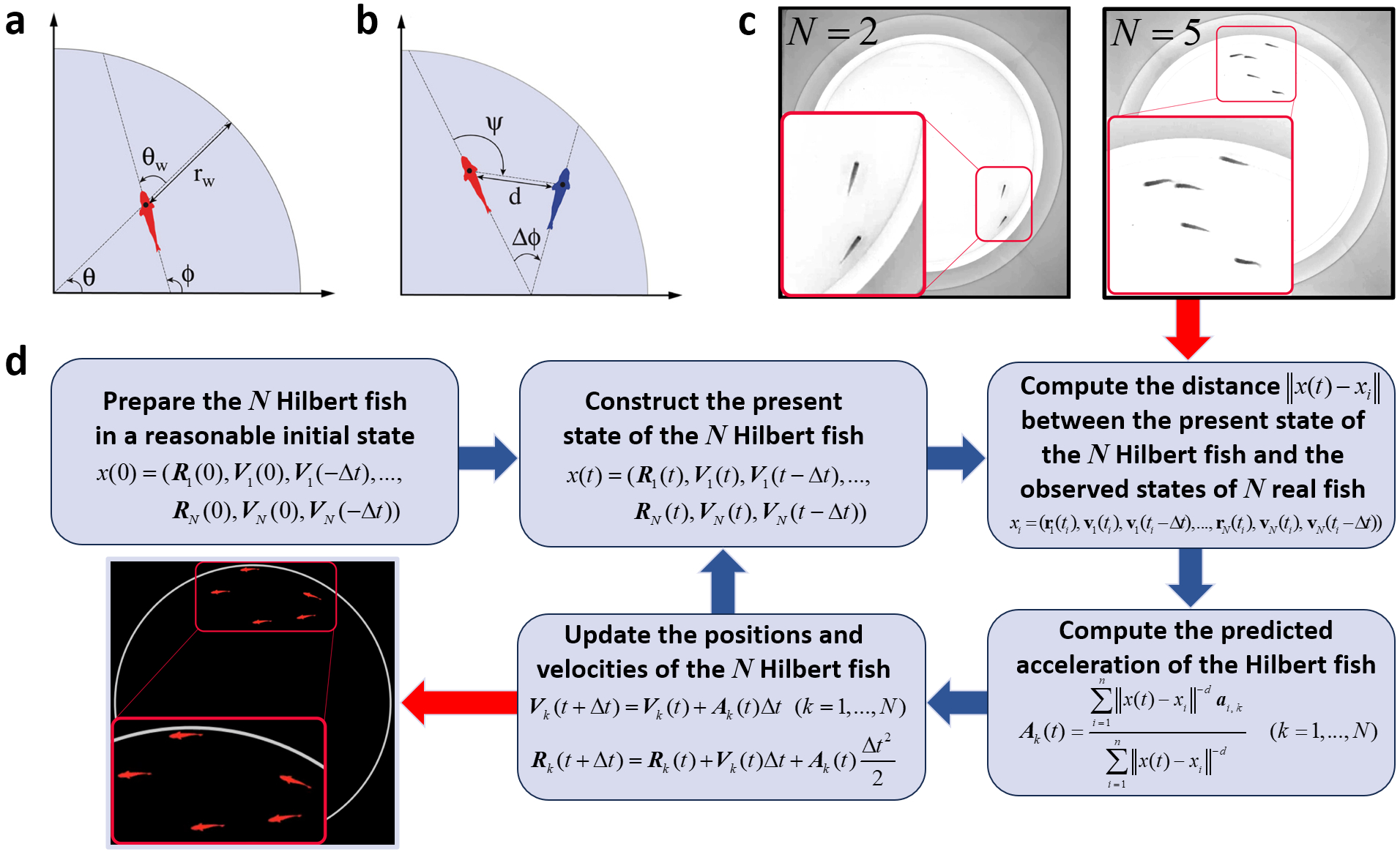}
         \caption{\small \textbf{Relevant fish variables and flow chart of the Hilbert interpolation scheme.} {\bf a},~Relevant variables for an individual: azimuthal angle, $\theta$; heading angle, $\phi$; heading angle relative to the normal to the wall, $\theta_{\rm w}=\phi - \theta$; distance to the wall, $r_{\rm w}$. {\bf b},~Relevant variables for a pair of individuals: distance between the individuals, $d$; relative heading angle between the 2 individuals, $\Delta\phi$; viewing angle at which the red focal individual perceives the other individual $\psi$.
         {\bf c},~Snapshots from experimental videos from~\cite{papaspyros2023biohybrid}, for $N=2$ and $N=5$ fish. {\bf d},~Flowchart describing the implementation of the Hilbert interpolation scheme (for a memory $M=1$) as a generative model for realistic fish trajectories.}
         \label{fig:flowchart}
\end{figure}

We now detail the implementation of the Hilbert interpolation scheme in the present context, which is summarized in Fig.~\ref{fig:flowchart}. We define the 2-dimensional positions, velocities, and accelerations of the $N$ Hilbert agents at a given time $t$ as, respectively, ${\bf R}_k(t)$, ${\bf V}_k(t)$, ${\bf A}_k(t)$, $k=1,...,N$. Similarly, the experimental positions, velocities, and accelerations of the $N$ real fish for the $i$-th data point (occurring at time $t_i$ in some experimental trajectory) are denoted in lowercases: ${\bf r}_k(t_i)$, ${\bf v}_k(t_i)$, ${\bf a}_k(t_i)$, $k=1,...,N$. The positions of the fish being measured with respect to the center of the tank at position $(0,0)$, and the tank being circular and bounded, the theoretical mean of ${\bf r}_k(t)$ and ${\bf v}_k(t)$ is zero, and we define the mean modulus of these quantities, $\sigma_r= \langle\|{\bf r}_k\|\rangle$ and $\sigma_v= \langle\|{\bf v}_k\|\rangle$, also averaged over the fish $k$. We then define the rescaled and dimensionless positions and velocities for the fish and the Hilbert agents by ${\bf \hat r}_k(t_i)= {\bf r}_k(t_i)/ \sigma_r$, ${\bf \hat v}_k(t_i) = {\bf v}_k(t_i)/ \sigma_v$, ${\bf \hat R}_k(t_i)= {\bf R}_k(t_i)/ \sigma_r$, ${\bf \hat V}_k(t_i) = {\bf V}_k(t_i)/ \sigma_v$.

We now introduce the memory $M$ which will characterize the input vector $x$ and the data vectors $x_i$ in the definition of the Hilbert weights, in Eqs.~(\ref{WeightDef},\ref{HilbertDef}). Indeed, for each Hilbert agent or real fish, its configuration at time $t$ is defined by its position at the current time, and its velocity at times $t$, $t-\Delta t$,..., $t-M\Delta t$:
\begin{eqnarray}
    x(t)=&&\left( {\bf \hat R}_1(t), {\bf \hat V}_1(t), {\bf \hat V}_1(t-\Delta t),...,{\bf \hat V}_1(t-M\Delta t),..., \right. \nonumber
    \\&&\left. {\bf \hat R}_N(t), {\bf \hat V}_N(t), {\bf \hat V}_N(t-\Delta t),...,{\bf \hat V}_N(t-M\Delta t)       \right),\label{input} \\
    x_i=&& \left( {\bf \hat r}_1(t_i), {\bf \hat v}_1(t_i), {\bf \hat v}_1(t_i-\Delta t),...,{\bf \hat v}_1(t_i-M\Delta t),..., \right. \nonumber
    \\&&\left. {\bf \hat r}_N(t_i), {\bf \hat v}_N(t_i), {\bf \hat v}_N(t_i-\Delta t),...,{\bf \hat v}_N(t_i-M\Delta t)       \right).\label{inputdata}
\end{eqnarray}
In Eqs.~(\ref{input},\ref{inputdata}), the input and data vectors involve rescaled positions and velocities so that all coordinates have the same typical magnitude of order unity. We note that these input and data vectors are of dimension,
\begin{equation}
    d = N\times 2\times [1+(M+1)] =  2N(M+2),\label{dimension}
\end{equation}
where the factor 2 comes from dealing with 2-dimensional position and velocity vectors, and the factor $M+2$ arises from taking into account the current position and $M+1$ present and past velocities of each agent/fish.
Then, following the definition of the Hilbert scheme in Eqs.~(\ref{WeightDef},\ref{HilbertDef}), and using Eqs.~(\ref{input},\ref{inputdata}), the predicted accelerations, ${\bf A}_k(t)$ ($k=1,...,N$), for the Hilbert agents are obtained as a weighted average of the $n$ experimental accelerations:
\begin{eqnarray}
    \left( {\bf A}_1(t),..., {\bf A}_N(t)  \right)&=& \sum_{i=1}^n w_i(x(t))  \left( {\bf a}_1(t_i),..., {\bf a}_N(t_i)  \right), \label{WeightFish}\\
    w_i(x(t)) &=& \frac{\|x(t)-x_i\|^{-d}}{\sum_{j=1}^n \|x(t)-x_j\|^{-d}}.
    \label{HilbertFish}
\end{eqnarray}

A subtlety arises in matching the identity of the $N$ Hilbert agents with the identity of the $N$ real fish. A natural procedure consists in using the best matching permutation of the identities that have the closest match to a training data point (see the discussion in Methods). 

Once the predicted accelerations, ${\bf A}_k(t)$ $(k=1,..., N)$, of the Hilbert agents have been evaluated using Eqs.~(\ref{WeightFish},\ref{HilbertFish}), their positions and velocities are updated according to the integration scheme,
\begin{eqnarray}
{\bf V}_k(t+\Delta t) &=& {\bf V}_k(t) + {\bf A}_k(t)\Delta t ,\label{updateV}\\
{\bf R}_k(t+\Delta t) &=& {\bf R}_k(t) + {\bf V}_k(t)\Delta t + {\bf A}_k(t)\frac{\Delta t^2}{2},\label{updateR}
\end{eqnarray}
and the time is then updated, $t \to t+\Delta t$. The procedure using Eqs.~(\ref{input},\ref{inputdata},\ref{WeightFish},\ref{HilbertFish},\ref{updateV},\ref{updateR}) and summarized in Fig.~\ref{fig:flowchart} is repeated to generate long trajectories of groups of $N$ Hilbert agents. Finally, we have implemented a rejection procedure enforcing the agent to strictly remain in the arena (see the discussion in Methods). Below, we also evaluate the performance of the Hilbert scheme when this constraint is released.

\begin{figure}[ht]
     \centering
        \includegraphics[width=\textwidth]{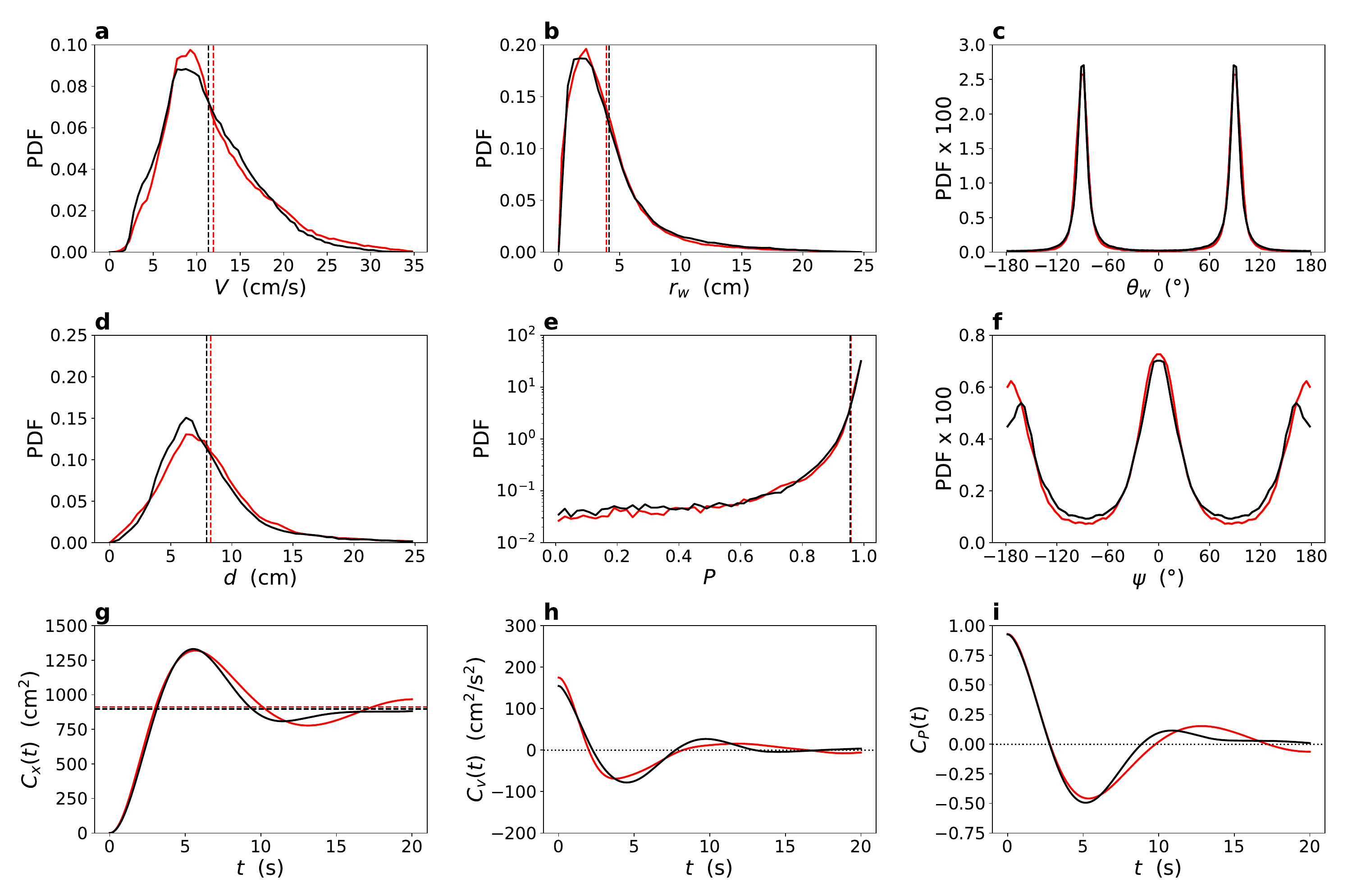}
         \caption{\small \textbf{Comparison of the behavior of real and Hilbert fish for $N=2$ individuals.} The different panels show the 9 observables used to characterize the individual ({\bf a}-{\bf c}) and collective ({\bf d}-{\bf f}) behavior, and the time correlations in the system ({\bf g}-{\bf i}): {\bf a},~PDF of the speed, $V$; {\bf b}, PDF of the distance to the wall, $r_{\rm w}$; {\bf c}, PDF of the heading angle relative to the normal to the wall, $\theta_{\rm w}$; {\bf d}, PDF of the distance between the pair of individuals, $d$; {\bf e}, PDF of the group polarization, $P=\left|\cos({\Delta\phi}/{2})\right|$, where $\Delta\phi $ is the relative heading angle; {\bf f},~PDF of the viewing angle at which an individual perceives the other individual, $\psi$. See Fig.~\ref{fig:flowchart}a and b for a visual representation of these different variables. {\bf g}, Mean squared displacement, $C_x(t)$, and its asymptotic limit, $C_x(\infty) =2\langle r^2\rangle\approx 900$\,cm$^2$ (dotted lines); {\bf h}, Velocity autocorrelation,  $C_v(t)$; {\bf i}, Polarization autocorrelation, $C_P(t)$.         
         The black lines correspond to experiments, while the red lines correspond to the predictions of the Hilbert generative model for a memory $M=2$ and for 4~hours of simulated trajectories. Vertical dashed lines correspond to the mean of the corresponding PDF (see also Extended Data Table~\ref{Table1}).}
         \label{fig:2fish_PDF}
\end{figure}

\subsubsection*{Hilbert scheme for the collective dynamics of $N=2$ fish} 

For $N=2$ fish, and following~\cite{calovi2018disentangling}, we quantify the individual behavior of the Hilbert agents or real fish by measuring  the probability density functions (PDF) of \textit{(1)}~their speed, \textit{(2)}~their distance to the wall, $r_{\rm w}$, and \textit{(3)}~their heading angle relative to the normal to the wall, $\theta_{\rm w}$ (see Fig.~\ref{fig:flowchart}a for a graphical representation of these variables). The collective behavior is quantified by the PDF of \textit{(4)}~the distance $d$ between the pair of individuals, \textit{(5)}~the group polarization,  $P=\|{\bf P}\| = \frac{1}{N}\left\lVert\sum_{i=1}^N (\cos\phi_i,\sin\phi_i)\right\rVert$, which reduces to $P=\left|\cos({\Delta\phi}/{2})\right|$, for $N=2$, where $\Delta\phi $ is the relative heading angle, and \textit{(6)}~the viewing angle at which an individual perceives the other individual, $\psi$ (see Fig.~\ref{fig:flowchart}b for a graphical representation of these variables). Finally, we quantify temporal correlations by \textit{(7)}~the mean squared displacement, $C_x(t)=\langle ({\bf r}_k(t+t')-{\bf r}_k(t'))^2\rangle$, \textit{(8)}~the velocity autocorrelation, $C_v(t)=\langle {\bf v}_k(t+t')\cdot{\bf v}_k(t')\rangle$, and \textit{(9)}~the polarization autocorrelation, $C_P(t)=\langle {\bf P}(t+t')\cdot{\bf P}(t')\rangle$, where the average is over the reference time~$t'$, and over the individual $k$ (only for $C_x(t)$ and $C_v(t)$). Time correlation functions are notably difficult to reproduce and represent a stringent test of a model~\cite{papaspyros2023ML}. 

The comparison of these 9 observables for the experiment and the Hilbert interpolation scheme for $M=2$ is presented in Fig.~\ref{fig:2fish_PDF}. We also report the mean and the standard errors for each PDF in Extended Data Table~\ref{Table1}. We find an excellent overall agreement between the Hilbert and experimental results, with a {significantly better accuracy} than the reported results for the data-driven mathematical model~\cite{calovi2018disentangling} or for the machine learning approach~\cite{papaspyros2023ML,papaspyros2023biogap}. Both real fish and Hilbert agents move at a typical speed of 11.5\,cm/s. They also remain close to the wall (mean distance $r_{\rm w}$ of 4\,cm). This is a consequence of the burst-and-coast swimming mode of rummy-nose tetra~\cite{calovi2018disentangling,Xuefishlight2023}, consisting in a succession of short acceleration periods (kicks/bursts of typical duration 0.1\,s), each followed by a longer gliding period (typically 0.5\,s) in a near straight line, ultimately preventing the fish to effectively escape the concavity of the tank wall. This is confirmed by the sharp peaks near -- but slightly below \cite{calovi2018disentangling} -- $\pm 90^\circ$ in the PDF of the heading relative to the normal to the wall, $\theta_{\rm w}$, showing that fish and Hilbert agents are often almost parallel to the wall. Finally, the motion of the pairs of fish and Hilbert agents is highly coordinated: both individuals remain at a typical distance of 8\,cm and the pairs are highly polarized ($P=0.95$). Finally, the PDF of the viewing angle $\psi$ at which an individual perceives the other quantifies the relative placement of the 2~individuals. The wide peaks near $\psi =0^\circ$ and $\psi = \pm 180^\circ$ reveal that an individual tends to stay in front or behind the other one. 

Study of the temporal correlations clearly shows the superiority of the Hilbert generated trajectories in mimicking the real behaviors over both the analytical and ML models. The mean squared displacement, $C_x(t)$, starts with a quadratic (ballistic) regime, $C_x(t)\approx \langle v^2\rangle t^2$, followed by a short linear (diffusive) regime, $C_x(t)\approx D t$, interrupted due to the bounded tank. Finally, $C_x(t)$ exhibits rapidly damped oscillations before leveling at $C_x(\infty)= 2\langle r^2\rangle$, when the two positions at times $t+t'$ and $t'$ become uncorrelated. The Hilbert model reproduces quantitatively the ballistic and diffusive regimes, and the main oscillation, as well as the asymptotic level. The velocity autocorrelation, $C_v(t)$, decays with damped oscillations from its maximal value, $C_v(t=0) = \langle v^2\rangle$. Note that the ML model of~\cite{papaspyros2023ML} and, to a lesser extent, the behavioral mathematical model of~\cite{calovi2018disentangling}, fail at reproducing the experimental $C_x(t)$ and $C_v(t)$. Finally, the model also reproduces the initial decay of the polarization autocorrelation, $C_P(t)$, from $C_P(t=0) = \langle P^2\rangle\approx 0.94$, and its main oscillation.
We illustrate these results in SI Movie~S1, which shows a 2-minute collective dynamics of 2 Hilbert agents. 

We have also studied the impact of the memory $M$. We find that using no memory ($M=0$) for the Hilbert input and data vectors of Eqs.~(\ref{input},\ref{inputdata}) would not produce coordinated trajectories. Our analysis below of the dynamics without the rejection procedure enforcing the presence of the wall will confirm that a memory $M=0$ fails to reproduce the correct fish behavior. We found that $M=1$, $M=2$, and $M=3$ lead to comparable results, while the agreement with experiments is degrading for a memory in the range $M=4$--$10$. A possible reason could be the increase of the dimension $d$ of the input and data vectors with $M$ (see Eq.~(\ref{dimension})), resulting in a scarcity of data in the $d$-dimensional configuration space ($d=16$ for $M=2$; $d=48$ for $M=10$). 

In addition, we have tested the stability of the Hilbert scheme when not implementing the rejection procedure strictly enforcing the presence of the tank wall (see Methods). For $M=0$ (no memory), the 2~agents quickly escape the tank, become independent, and wander several meters away from the tank, confirming that the Hilbert scheme for $M=0$ is unable to capture the social interactions between fish nor the presence of the wall. We ascribe the failure of the $M=0$ Hilbert model to the fact that the input and data vectors $x$ and $x_i$ in Eqs.~(\ref{input},\ref{inputdata}) do not encode enough information to evaluate the similarity of a fish configuration and a Hilbert agent's configuration. In fact, when $M>0$, these vectors now implicitly encode an estimate of the acceleration, through the knowledge of the velocity at time $t$, but also at past times $t-\Delta t,...,t-M\Delta t$. Indeed, for $M>0$, we find that the Hilbert agents remain cohesive, stay inside the tank for the vast majority of the simulations (87\,\% for $M=2$), and only make brief and small (1.3\,cm away from the tank wall, on average) excursions outside the tank. This is illustrated for $M=2$ in the Extended Data Fig.~\ref{extfig:2fish_nowall}, which corresponds to Fig.~\ref{fig:2fish_PDF} but without the rejection procedure. The deep learning models of~\cite{papaspyros2023ML,papaspyros2023biogap} present a similar stability without the rejection procedure, showing that both Hilbert (for $M>0$) and ML models can implicitly capture and learn the presence of the wall from the data. In fact, note that the deep learning models of~\cite{papaspyros2023ML,papaspyros2023biogap} also exploit a memory of the past, encoded in the input data (with an effective $M=5$) and in their architecture involving LSTM layers. SI Movie~S2 shows a 2-minute collective dynamics of 2~Hilbert agents without enforcing the presence of the wall, confirming that the Hilbert interpolation scheme has implicitly learned the presence of the wall from the data.

A valuable advantage of the Hilbert interpolation scheme compared to ML methods is that it explicitly gives the weight $w_i(x)$ of a given data $x_i$ involved in the current prediction, for a given input $x$ (see also the final section about credit assignment). To quantify the effective number of data, ${\cal N}$, involved in this prediction, we define the entropy,
\begin{equation}
    S(t)= -\sum_{i=1}^n w_i(x(t))\log_2[w_i(x(t))],
\end{equation}
where the weights $w_i(x(t))$ are defined in Eq.~(\ref{HilbertFish}). If the weights at a given time $t$ are equidistributed over ${\cal N}$ data, one obtains $S=-{\cal N}\times 1/{\cal N}\log_2(1/{\cal N}) =\log_2({\cal N})$, and $2^S={\cal N}$ indeed represents the number of contributing data. 
In particular, in the presence of “copying” behavior, ${\cal N}=1$ and $S=0$.
In the Extended Data Fig.~\ref{extfig:2fish_entropy_time}, we show a 2-minute time series for $S(t)$ exhibiting very short periods when $S\approx 0$ (${\cal N}\approx 1$), meaning that the Hilbert prediction is essentially using/copying a single experimental data/configuration to compute the predicted accelerations. On the other hand, this short time series also presents 3 periods when $S>6$ (${\cal N}> 64$). In the Extended Data Fig.~\ref{extfig:2fish_entropy_time}, we plot the PDF of $S(t)$, $\rho(S)$, which presents an integrable divergence near $S=0$ ($\rho(S)\sim S^{-1/2}$), and a Gaussian tail for large $S$. Over 3~effective hours of simulations, we recorded entropies as large as $S\sim 15$, corresponding to an effective number of data involved in the prediction of order ${\cal N}\sim 2^{15}\sim 32768$. Compared to $k$NN methods, the Hilbert interpolation scheme is able to adapt the effective number of data points used for the prediction to the properties of the input vector. 

It is important to note that the close match between the observables and correlations from the Hilbert generated trajectories and the experimental observations cannot simply be attributed to copying behavior. A study of the weight entropy (Extended data Fig.~\ref{extfig:2fish_entropy_time}) shows only brief episodes of near-copying behaviors, shorter than the timescales involved in the correlation functions and too infrequent to explain the results.

\begin{figure}[ht]
     \centering
        \includegraphics[width=\textwidth]{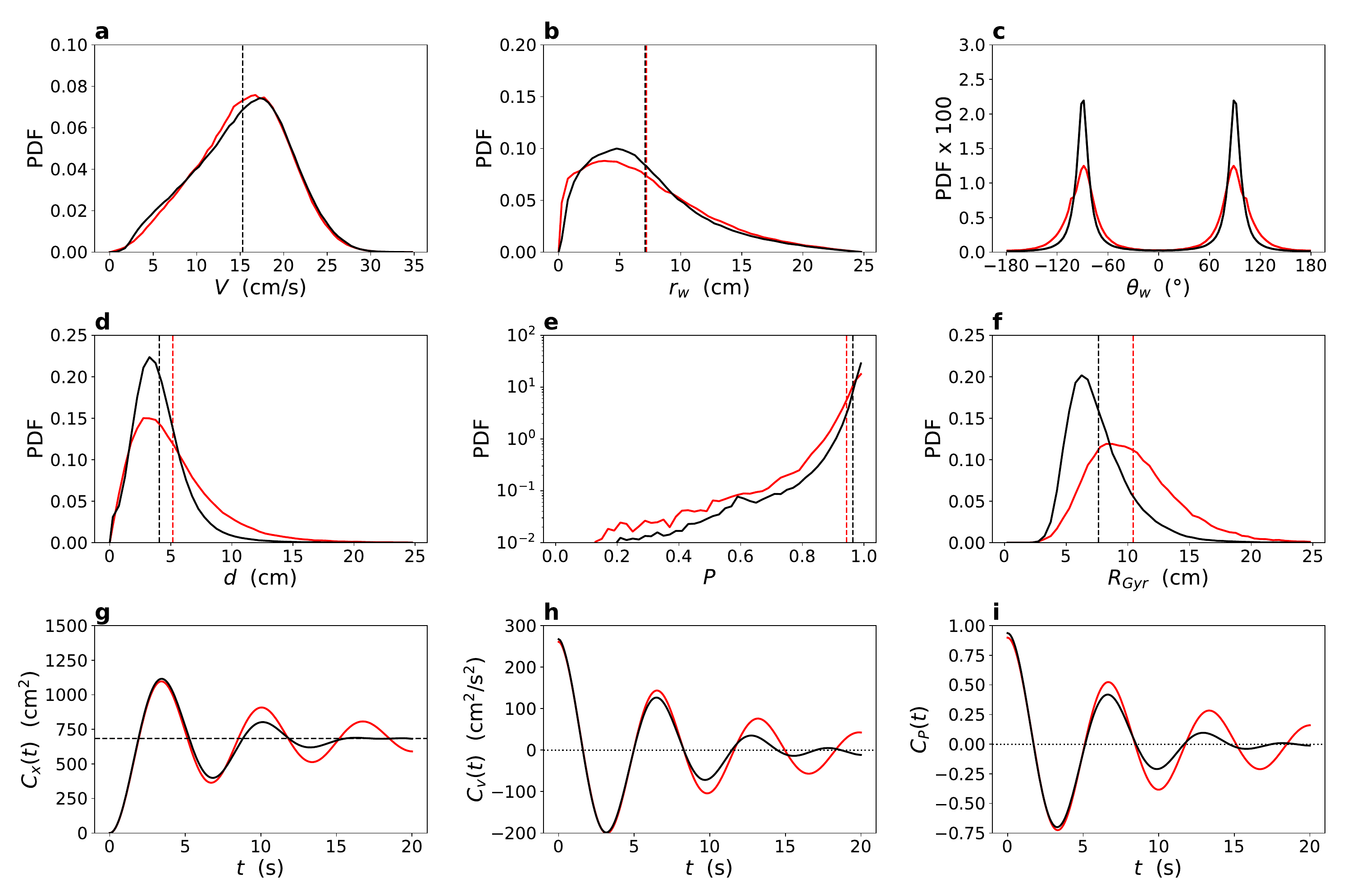}
          \caption{\small \textbf{Comparison of the behavior of real and Hilbert fish for $N=5$ individuals.} The different panels show the 9 observables used to characterize the individual ({\bf a}-{\bf c}) and collective ({\bf d}-{\bf f}) behavior, and the time correlations in the system ({\bf g}-{\bf i}): {\bf a},~PDF of the speed, $V$; {\bf b}, PDF of the distance to the wall, $r_{\rm w}$; {\bf c}, PDF of the heading angle relative to the normal to the wall, $\theta_{\rm w}$; {\bf d}, PDF of the distance between an individual and its closest neighbor, $d$; {\bf e}, PDF of the group polarization,~$P$; {\bf f}, PDF of the radius of gyration of the group, $R_{\rm Gyr}$. See Fig.~\ref{fig:flowchart}a and b for a visual representation of the main variables. {\bf g},~Mean squared displacement, $C_x(t)$, and its asymptotic limit, $C_x(\infty) =2\langle r^2\rangle\approx 680$\,cm$^2$ (dotted lines); {\bf h}, Velocity autocorrelation,~$C_v(t)$; {\bf i}, Polarization autocorrelation,~$C_P(t)$. The black lines correspond to experiments, while the red lines correspond to the predictions of the Hilbert generative model for a memory $M=2$ and for 2~hours of simulated trajectories. Vertical dashed lines correspond to the mean of the corresponding PDF (see also Extended Data Table~\ref{Table1}). The horizontal and vertical scales of each graph are the same as that of the corresponding graph in Fig.~\ref{fig:2fish_PDF}, for $N=2$. }
         \label{fig:5fish_PDF}
\end{figure}

\subsubsection*{Hilbert scheme for the collective dynamics of $N=5$ fish} 

For $N=5$ fish, we quantify the individual behavior and the temporal correlations of the Hilbert agents or real fish with the same 6 observables as for $N=2$, but we adapt 2~of the 3 collective observables to the larger fish group~\cite{lei2020computational,Xuefishlight2023}: PDF of \textit{(4')}~the distance $d$ between an individual and its nearest neighbor, and \textit{(6')}~the radius of gyration of the group, $R_{\rm Gyr}=\sqrt{\frac{1}{N(N-1)}\sum_{i\ne j}\lVert {\bf R}_i-{\bf R}_j\rVert^2}$, which is a measure of the size of the group. 

The comparison of the 9 observables for the experiment and the Hilbert interpolation scheme for $M=2$ is presented in Fig.~\ref{fig:5fish_PDF}, where a fair general agreement is obtained, comparable to that of the data-driven mathematical model~\cite{lei2020computational,Xuefishlight2023}. We also report the mean and the standard errors of these different observables in Extended Data Table~\ref{Table1}. The PDF of the speed and distance to the wall are in good agreement with experiments and show that groups of 5 fish and Hilbert agents are moving significantly faster than pairs (mean speed 15.3\,cm/s for $N=5$ vs. 11.5\,cm/s for $N=2$), and significantly farther from the wall than pairs (mean distance to the wall 7\,cm for $N=5$ vs. 4\,cm for $N=2$). The PDF of the angle relative to the wall is less peaked than for $N=2$, showing that 5 fish or Hilbert agents are less often parallel to the wall than corresponding pairs. However, for $N=5$, the Hilbert agents are slightly less coordinated than real fish, with a larger distance between nearest neighbors (4\,cm for fish vs. 5\,cm for Hilbert agents). Note that in both cases, individuals are, on average, closer to each other than for $N=2$ (mean distance of 8\,cm)~\cite{lei2020computational,Xuefishlight2023}. Accordingly, the typical size of the group (quantified by the radius of gyration) is significantly smaller for fish than for Hilbert agents (7.6\,cm vs. 10.4\,cm), which represents the worst agreement observed so far. Yet, groups of $N=5$ fish and Hilbert agents are extremely and almost equally polarized ($P=0.96$ for fish vs. $P=0.94$ for Hilbert agents). Still, considering that the trajectories of the 5~Hilbert agents are integrated independently, it is remarkable that the Hilbert model produces such cohesive and highly polarized groups.
The Hilbert model reproduces fairly well the 3 experimental correlation functions, albeit with a longer persistence of the oscillations than for fish.
The mean squared displacement $C_x(t)$ presents a smaller amplitude than for $N=2$ (see Fig.~\ref{fig:2fish_PDF}), since fish and Hilbert agents are, on average, moving farther from the wall tank for $N=5$. The period of the oscillations is shorter for $N=5$, since fish and Hilbert agents move faster than for $N=2$, and these oscillations are better marked than for $N=2$. The velocity autocorrelation, $C_v(t)$, and the polarization autocorrelation, $C_P(t)$, also display larger, faster, and more persistent oscillations than for $N=2$. These results indicate that 5~fish or Hilbert agents maintain strong temporal correlations for larger traveled distance than corresponding pairs.
We illustrate these results in SI Movie~S3, which shows a 1-minute collective dynamics of 5 Hilbert agents.  

In this section, we have applied the generative Hilbert scheme to the modeling of animal collective motion. The statistical properties of the generated trajectories closely match experimental observations as quantified by 9~observables probing the instantaneous individual and collective behavior, as well as the temporal correlations. These observables offer a stringent test of the model, and in the present fish context, their analysis indicates that the Hilbert scheme provides a new and complementary approach to data-driven analytical modeling and ANN-based modeling, with equally good or better predictive capabilities. The analytical fish models of~\cite{calovi2014swarming,calovi2018disentangling} follow a canonical tradition in scientific theorizing, using simple and interpretable low-order (continuous or discrete) ordinary differential equations. On the other hand, both the ANN-based and Hilbert generative models are purely data-driven and neither offer a similar degree of interpretability as mathematical models. However, we posit that the Hilbert scheme nevertheless offers significant advantages over the ANN-based model. 

The Hilbert generative model leads to a straightforward implementation, consisting of a few lines of code  (see Fig.~\ref{fig:flowchart}). There is no training, parameter tuning, or model selection, and it offers strict reproducibility, portability, and transparency, which is particularly important for scientific applications. This framework also permits the reuse of generated data in real time to augment the experimental dataset. Moreover, the relative contribution of training data points contributing to a specific prediction is transparently characterized using the corresponding weights, and the procedure flexibly adapts those weights across training data points for any given prediction. 

In conclusion, the present work shows that the Hilbert scheme constitutes an advantageous alternative to ANN-based generative approaches in the context of the very active field of collective animal motion and behavior.

\subsection*{Hilbert predictive control of nonlinear dynamical systems}

Control systems are central to modern technology. The goal of control engineering is to ensure that systems have behaviors suited to the task at hand. Active control of a system generally involves two steps: modeling the response of the system state to controlled external inputs (system identification), and using this model to design appropriate inputs to make the system state follow a desired dynamical trajectory (controller design). In complex modern applications, the system models are increasingly data-driven, and are often exploiting ANNs. As with other applications of black box networks, the precise role of the network-based system models remains unclear. 

In keeping with the central idea in this work, we show that parametric system models are not required and that dynamical control can be achieved by simply interpolating previous observations of system responses to inputs. Our approach is similar in spirit to the behavioral approach to control due to Willems~\cite{willems1997introduction}. However, this earlier framework exploits the linear dependency between sequences of lag vectors of the system outputs and inputs, which only exist for linear dynamical systems. In contrast, our approach to system modeling based on Eq.~(\ref{eq:ARMAmodel}) is applicable more generally to nonlinear systems. 

\renewcommand{\thesubfigure}{\bf\alph{subfigure}}

\begin{figure}[h]
     \centering
     \begin{subfigure}[t]{0.32\textwidth}
        \includegraphics[width=\textwidth]{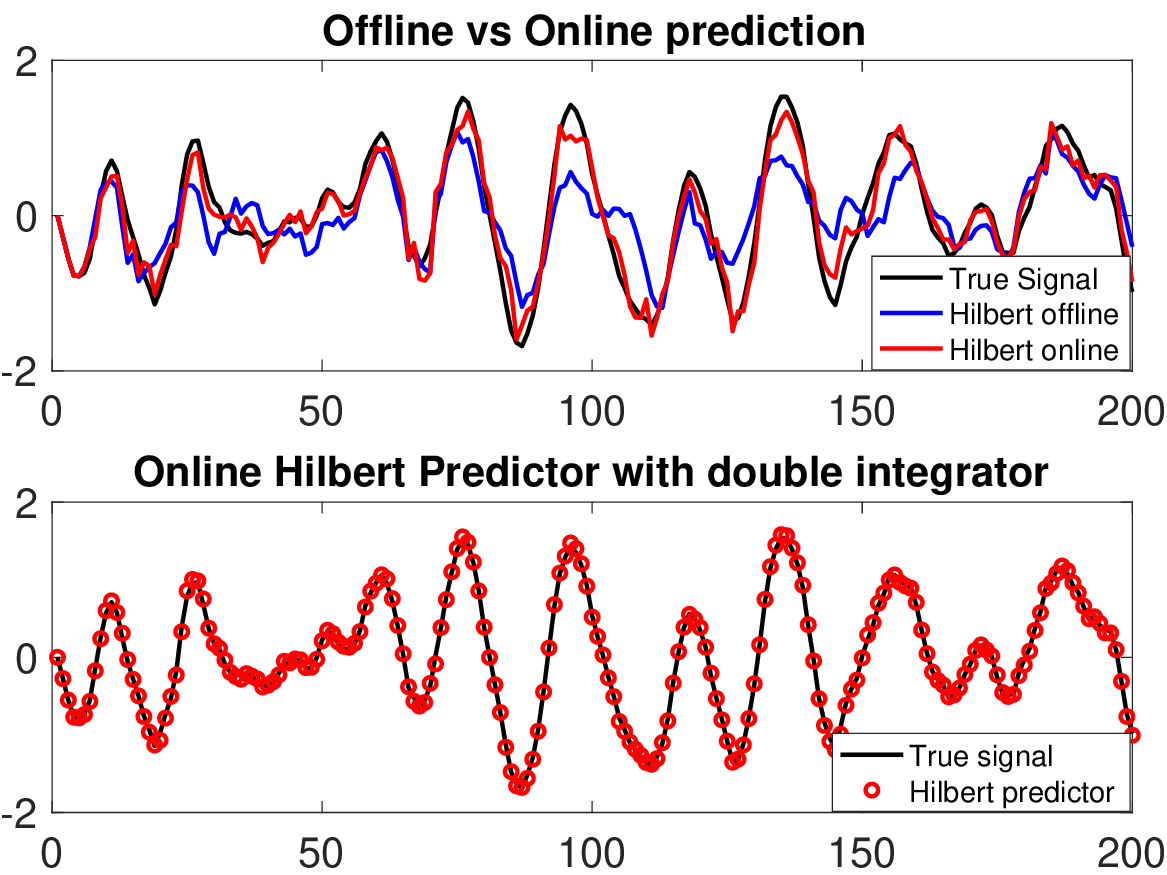}
         \caption{\small Damped oscillator}
         \label{figcontrol1a}
     \end{subfigure}
     \hfill
     \begin{subfigure}[t]{0.32\textwidth}
         \includegraphics[width=\textwidth]{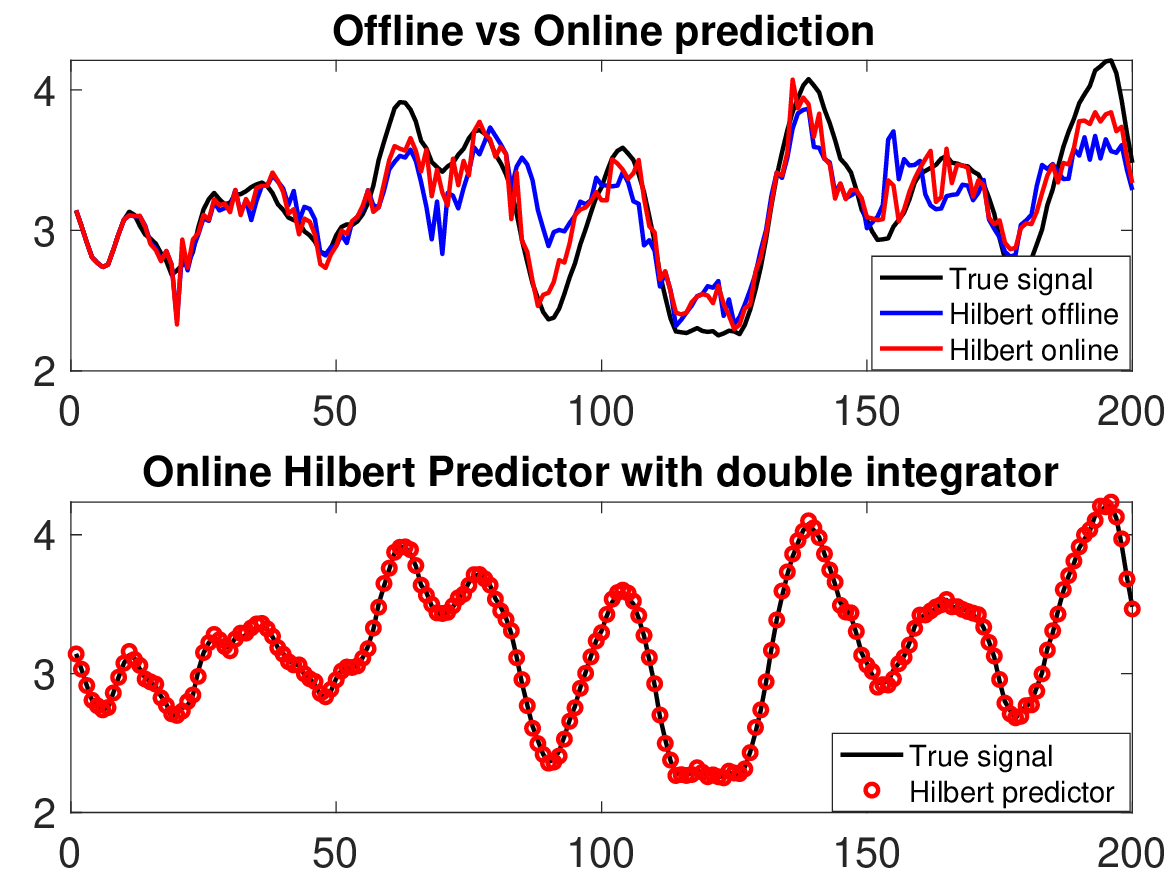}
         \caption{\small Inverted pendulum}
         \label{figcontrol1b}
     \end{subfigure}
     \hfill
     \begin{subfigure}[t]{0.32\textwidth}
         \includegraphics[width=\textwidth]{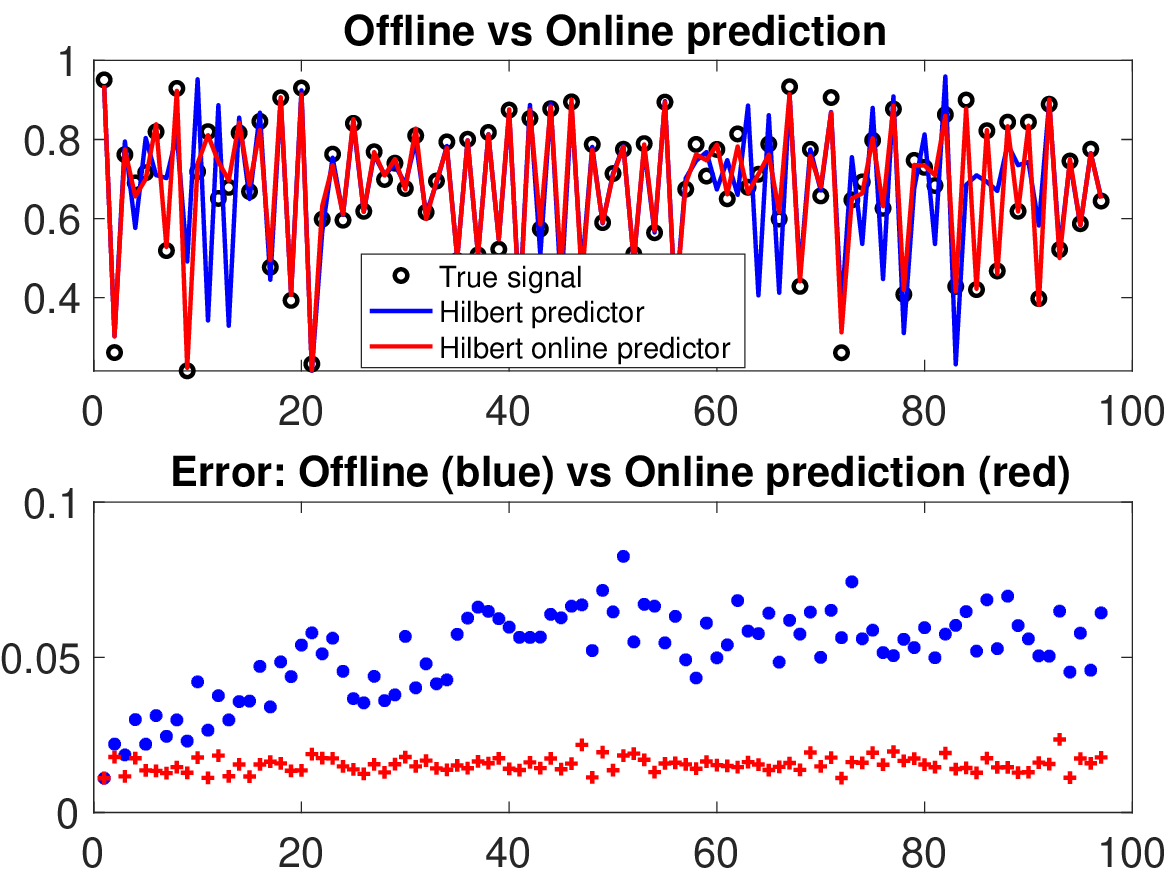}
         \caption{\small Driven logistic}
         \label{figcontrol1c}
     \end{subfigure}
        \caption{\small \textbf{Input-output prediction using Hilbert autoregression (Eq.~(\ref{eq:ARMAmodel})).} Fig.~1a shows offline and online prediction of an input driven, linear two-state system, as well as online prediction with a double integrator where the acceleration is predicted and integrated twice to obtain the time series prediction. The variance explained by the offline and online autoregressive prediction methods are 66.18\% and 94.20\%. When the acceleration is predicted, and integrated twice to obtain the signal prediction, 99.87\% of the variance is explained. Fig 1b shows similar predictions for a driven inverted pendulum, where $\theta=\pi$ corresponds to the pendulum positioned vertically down. The variance explained by the offline and online autoregressive prediction methods are 65.46\% and 87.10\%, and for the online double-integration based prediction, 99.87\% of the variance is explained. Note that the variance explained is similar for the linear and the nonlinear dynamical system examples. Fig 1c shows predictions of a driven logistic map (Eq.~(\ref{eq:DrivenLogistic}) with $\lambda=3.99, K=0.25$) using offline and online autoregressive prediction. The prediction is fairly accurate even for this highly nonlinear map: the offline predictor has a residual of 5.46\% (median absolute deviation) and the online predictor has a residual of 1.54\%. The top panel of 1c shows one representative instance, whereas the normalized median absolute deviations below are computed from ten independent runs with different random seeds. Notably, the offline prediction starts with a similar error as the online prediction, but over time the offline prediction deteriorates somewhat.}
        \label{fig:ControlsFig1}
\end{figure}

\begin{figure}[h]
     \centering
     \begin{subfigure}[t]{0.44\textwidth}
        \includegraphics[width=\textwidth]{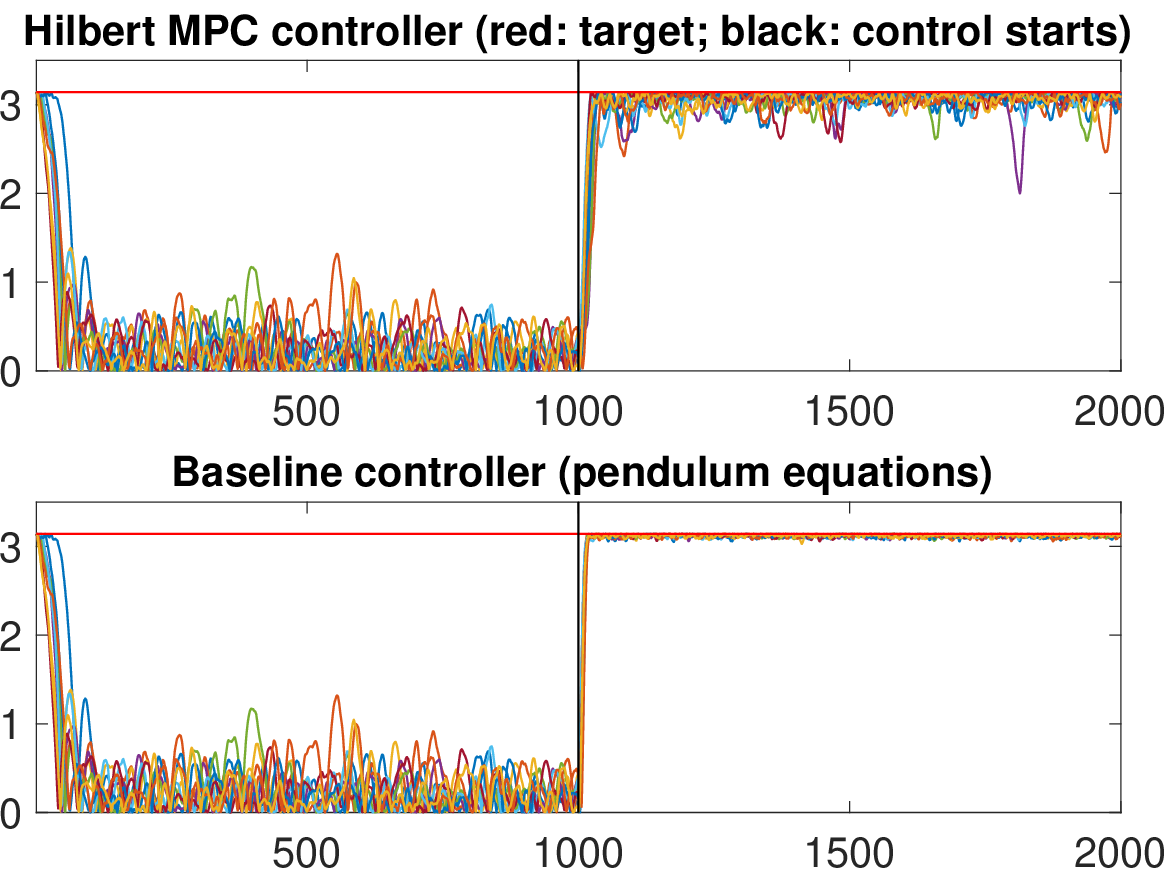}
         \caption{\small Inverted pendulum control}
         \label{figcontrol2a}
     \end{subfigure}
     \hfill
     \begin{subfigure}[t]{0.44\textwidth}
         \includegraphics[width=\textwidth]{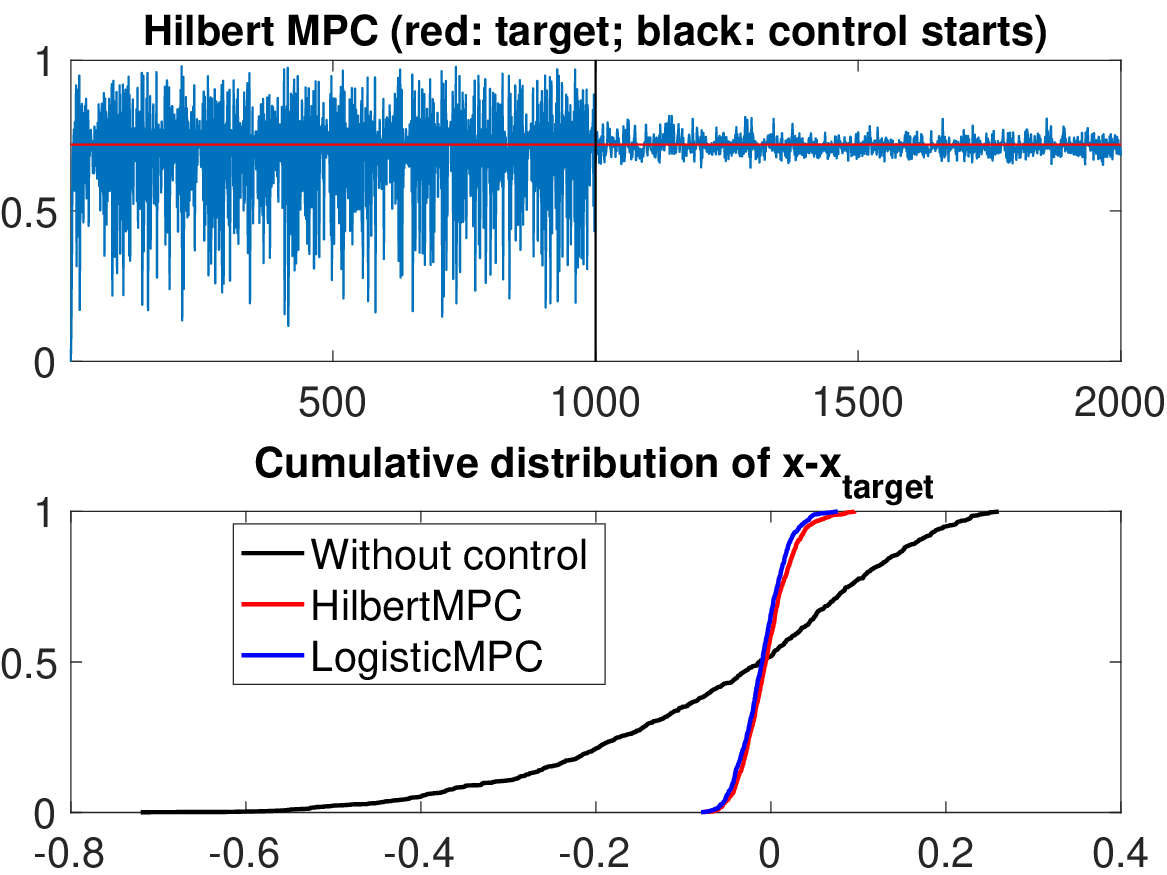}
         \caption{\small Driven logistic map control}
         \label{figcontrol2b}
     \end{subfigure}
        \caption{\small \textbf{Hilbert Predictive Control of nonlinear dynamical systems.} The left panels show that an inverted pendulum can be controlled to its unstable equilibrium with the MPC approach, where the driven system dynamics is predicted using Hilbert interpolation illustrated in Fig.~\ref{fig:ControlsFig1} rather than a parametric model. For an initial period the pendulum is driven by Gaussian white noise input to generate training data. The controller is activated at the black vertical line, following which the pendulum is driven to its unstable equilibrium position, and maintained there with small fluctuations (top left). As a comparison, the true pendulum equations used to generate the dynamics are used in the same MPC framework to show a baseline controller (bottom left). The plots show $\arccos(cos(\theta))$ with the angle $\theta=\pi$ corresponding to the vertically up configuration, for ten independent runs with different random number seeds. The RMS error achieved in steady state for the Hilbert MPC approach is 0.0129 radians, compared to a baseline RMS error using the true system model of 0.0027 radians. Similarly, the top right panel shows the Hilbert predictive controller approach applied to the control of a driven logistic map Eq.~(\ref{eq:DrivenLogistic}) around an unstable fixed point of the driven system for a constant input ($x_{target}=0.72$, $u_{target}=0.4668$). Uniformly distributed white noise is input during the training period and the controller is turned on at the black line. The controller is effective in stabilizing the system around the unstable fixed point (the control is turned on at the black vertical line). The control is statistically indistinguishable from a baseline case where the underlying driven logistic map is used instead of the Hilbert prediction. We quantify this using the standard deviation of $x-x_{target}$ normalized by the corresponding standard deviation when the controller is not turned on. This ratio, which should ideally be zero, is $0.1434\pm 0.0063$ (sd) for the Hilbert case and $0.1479\pm 0.0078$ (sd) for the baseline case, where the standard deviations are computed with ten independent runs with different random number seeds. 
        }
        \label{fig:ControlsFig2}
\end{figure}

We use training data consisting of past input-output observations to predict the future input-output behavior of the system using the Hilbert-based transfer function predictor Eq.~(\ref{eq:ARMAmodel}). We work within the framework of Model Predictive Control (MPC) where the predictive model is replaced by Eq.~(\ref{eq:ARMAmodel}). In MPC, a cost function is optimized over a future horizon over all possible inputs to find the optimal input over that horizon, then only the first input time step is implemented, following which the entire procedure is iterated. We perform this optimization step using a stochastic gradient descent like method where only cost function evaluations are utilized (see Methods). We apply this approach to two nonlinear control problems: stabilizing an inverted pendulum around the unstable vertical equilibrium (Fig.~\ref{figcontrol2a}), and stabilizing a driven logistic map (Fig.~\ref{figcontrol2b}) defined by 
\begin{equation}
    x_{t+1}=(1-K)\times\lambda\, x_t (1-x_t) + K\times u_t
    \label{eq:DrivenLogistic}
\end{equation}
also around an unstable fixed point. 

In Fig.~\ref{fig:ControlsFig1}, the effectiveness of the Hilbert-based transfer function is shown for three increasingly nonlinear cases: a linear two-state system whose impulse response is a damped oscillation (Fig.~\ref{figcontrol1a}), a torque-driven pendulum (Fig.~\ref{figcontrol1b}) and a driven logistic map (Fig.~\ref{figcontrol1c}). In each case, the top row exhibits the output of the autoregressive predictor (Eq.~(\ref{eq:ARMAmodel})) in blue superposed on the true signal in black (using lag windows of size $5,10$ and $3$ respectively). This case is labeled Hilbert offline, since the prediction is carried out using a fixed past episode of training data. The red curves correspond to the same procedure applied in an online manner (see Methods), in which at each time step, a measurement of the true system output is added back to the training data together with the control input applied. It can be seen that the online predictor performs better than the offline predictor. For the smooth dynamical systems (Fig.~\ref{figcontrol1a} and \ref{figcontrol1b}), the result is even better if the acceleration is predicted (see Methods), rather than the signal directly, and the acceleration integrated twice to obtain the prediction in an online manner. In this case, the error as quantified by the unexplained variance is $\sim 0.1\%$ and the black curve (original) and prediction (red symbols) are visually indistinguishable. For the driven logistic map case, the dynamics is non-smooth and predicting the acceleration does not improve performance, but the online predictor has a residual of only $1.54\%$. The offline prediction slightly deteriorates  over time, but still has a residual variance of $5.46\%$. It is important to re-emphasize that in each case, precisely the same (short) code together with the training data is used for the predictive step: there is no system modeling involved. 

\renewcommand{\thesubfigure}{\bf\alph{subfigure}}

Armed with the input-output predictors, we can proceed to the next step of controlling the outputs to a desired target. To achieve this goal, we utilize the MPC framework, which is an iterative process in which at each time step, a prediction is made over some time horizon $T$ for command inputs which minimize a suitable cost (we chose constant targets and a quadratic cost on the output signal). In the existing MPC framework, a parametric system model trained on previous data is used to make the prediction. In our case, the parametric model is replaced by the Hilbert-based predictor. Not having a parametric model means that cost function optimization methods that utilize analytically computed gradient information cannot be used. We used a stochastic gradient descent style optimization method that requires only function evaluations, in which a directional gradient is estimated in a randomly chosen direction \cite{nesterov2017random} (see Methods) and an appropriately scaled step is taken in that direction. This approach works for stabilizing two very different nonlinear dynamical systems: an inverted pendulum and a driven logistic map (Fig.~\ref{fig:ControlsFig2}). The proposed procedure for data-driven control is new and could be termed Hilbert Predictive Control (HPC), although note that any other SCI scheme would work, so an alternative name could be Interpolating Predictive Control. 

Fig.~\ref{fig:ControlsFig2} shows that HPC can stabilize nonlinear dynamical systems around unstable fixed points. In Fig.~\ref{figcontrol2a} a pendulum is controlled to its unstable equilibrium, pointing upwards. The pendulum is initially subjected to a Gaussian white noise input to generate training data. The controller is then turned on at a time marked by the black line, causing the pendulum to swing up and remain near the unstable equilibrium point. The top panel in Fig.~\ref{figcontrol2a} shows $10$ independent runs with different random number seeds, with an RMS (root mean square) departure from target of 0.0129 radians. The bottom panel in Fig.~\ref{figcontrol2a} shows ten independent runs with the same seeds, where the true pendulum model is used in place of the Hilbert predictor, retaining the same derivative-free optimization scheme, giving an RMS departure from target of 0.0027 radians. In Fig.~\ref{figcontrol2b} the top panel shows the result of applying HPC to the driven logistic map Eq.~(\ref{eq:DrivenLogistic}). After an initial period of uniformly distributed white noise input (confined to $[0,1]$) to generate training data, the control is turned on and the output quickly collapses to the target under the HPC approach, where the system model is replaced by the Hilbert predictor Eq.~(\ref{eq:ARMAmodel}). In this case, we find that if the true system model is used instead of the Hilbert predictor (keeping the controller design optimization method the same), the results are indistinguishable. The cumulative distribution function of the deviation from the target position of the Hilbert controller (Fig.~\ref{figcontrol2b}, bottom panel red curve; normalized standard deviation $0.1434\pm 0.0063$) is statistically indistinguishable from the corresponding deviations after applying the MPC controller using the true system model (Fig.~\ref{figcontrol2b}, bottom panel blue curve, normalized standard deviation $0.1479 \pm 0.0078$). 

In conclusion, it is possible to control nonlinear dynamical systems around unstable fixed points using the Hilbert scheme adapted to input-output modeling. Notably, the control algorithm for the two widely divergent cases is identical and does not involve parametric models of any sort. To the best of our knowledge, this is unprecedented in control theory and engineering, where parametric models dominate and no such universal controller with broad applicability exists. 

\subsection*{Hilbert predictor for the zeros of the Riemann Zeta function}

The Riemann zeta function, defined by $\zeta(s) =\sum_{n=1}^{+\infty} \frac{1}{n^s}$ (for $|s|>1$; and by its analytical continuation elsewhere), plays a significant role in number theory due to its close relation with the distribution of prime numbers~\cite{edwards2001riemanns}. Central to this connection is the Riemann hypothesis, which conjectures that all nontrivial zeros of $\zeta(s)$  are of the form $s=\frac{1}{2} + i E$, and are hence all located on a line in the complex plane. If confirmed, this conjecture would provide validation for a plethora of theorems in number theory that have been proven under the assumption of the Riemann hypothesis. Even more intriguing is the stronger conjecture that the imaginary part of the zeroes, $E$ (that we will assimilate to the zeroes of $\zeta$), are the eigenvalues (i.e., energy levels) of a quantum Hamiltonian and are hence aligned on the real line~\cite{Berry1999}. 

\begin{figure}[ht]
    \centering
    \includegraphics[width=\textwidth]{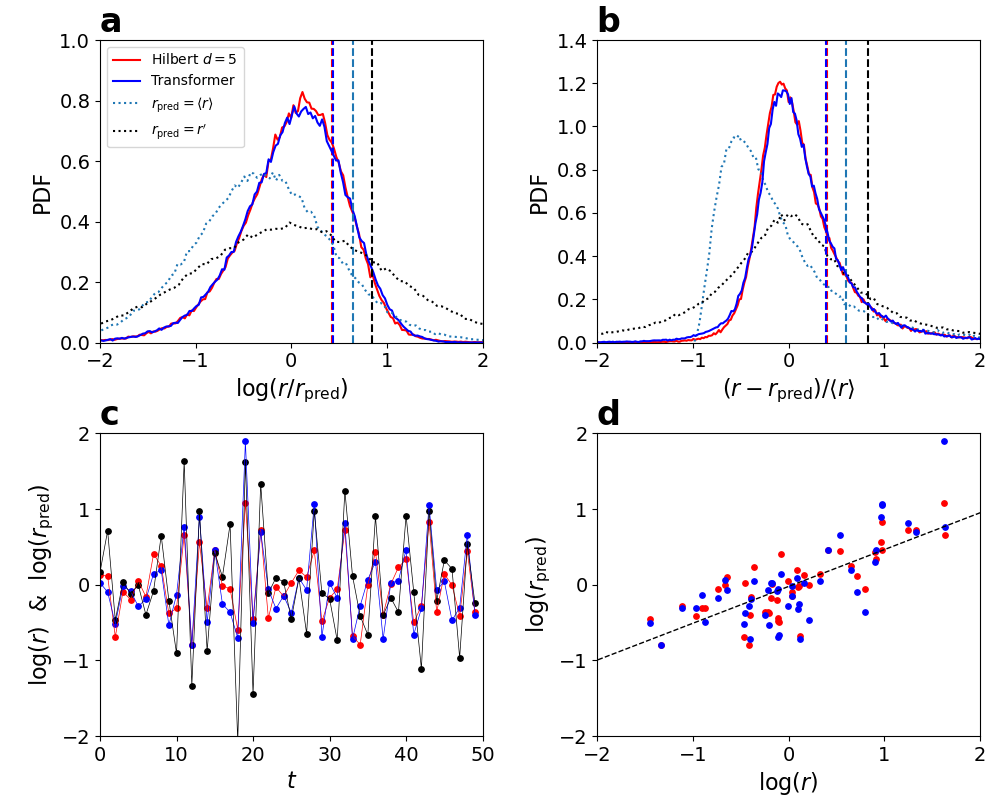}
    \caption{\small \textbf{Predicting zeros of the Riemann function.} {\bf a}, PDF of $\log(r/r_{\rm pred})$ (similar to a signed relative error), where $r$ is the actual observed ratio, and $r_{\rm pred}$ is the prediction of the Hilbert ($d=5$; red line) and Transformer (blue line) predictors (200\,000 predictions). For reference, we also plot the corresponding PDF for two other natural predictors: $r_{\rm pred}=\langle r \rangle$ (the expected value of the ratio; light blue dots) and $r_{\rm pred}=r'$ (a ratio randomly drawn in the dataset; black dots). The vertical dashed lines are the corresponding mean error, $\langle | \log(r/r_{\rm pred})|\rangle$.  {\bf b}, PDF of the normalized absolute error $(r -r_{\rm pred})/ \langle r \rangle$, and using the same notations. In both graphs, the PDF for the Hilbert and Transformer predictors are nearly identical and result in similar errors, significantly lower than for the two other simple predictors. Panel {\bf c} shows a sequence of 50 successive predictions of the Hilbert (red dots) and Transformer (blue dots) predictors, along with the actual observed values of $\log(r)$ (black dots). {\bf d} shows the same 50  predictions as a function of the actual observed log-ratio, along with the linear correlation measured between the 200\,000 predictions and observed values for the log-ratio (black dashed line), corresponding to a Pearson correlation coefficient $R=0.7$.}
    \label{fig:Riemann1}
\end{figure}

This last conjecture is numerically confirmed by the fact that the high-energy levels of non-integrable quantum Hamiltonians and large ``Riemann energies'' $E$ share the same statistical properties as the eigenvalues of large Hermitian random matrices with Gaussian entries~\cite{keating2000RMT}. In particular, the spacings between successive eigenvalues (for quantum or random Gaussian operators) or Riemann energies, normalized by the local mean spacing, obey a universal statistics. More precisely, in the context of Riemann energies, we define $S_i = (E_{i+1} -E_i)/\Delta E$, as the difference between the $(i+1)$-th and $i$-th zero of $\zeta$, normalized by the mean local spacing $\Delta E =\frac{2\pi}{\log(E_i)}$~\cite{edwards2001riemanns}. Then, for large $i$, the variable $S_i$ effectively behaves like a true random variable following the universal GUE statistics given by random matrix theory, with a PDF very well approximated by the Wigner surmise, $P(S) = \frac{32}{\pi^2}S^2\exp(-\frac{4}{\pi}S^2)$. Random matrix theory also makes precise predictions for  the correlation function between different energy levels, which are numerically verified for the zeros of the Riemann function. In order to avoid the need to normalize the spacings by the local mean spacing, it is customary to introduce the ratio between two successive spacings, $r_i = (E_{i+1} -E_i)/(E_i -E_{i-1})$, which now behaves like a quasi stationary random variable. Again, the PDF of the ratio $r$ for Riemann zeros is in excellent agreement with the prediction of random matrix theory~\cite{Atas2013ratio}.

By assimilating the index $i$ of a zero to a time, we  have exploited the Hilbert scheme to predict the next ratio $y=r(t)$, knowing $x=(r(t-1),~r(t-2),...,r(t-d))$, the latter $d$-dimensional vector constituting the input for the Hilbert kernel in Eq.~(\ref{eq:ARmodel}). As discussed in the Methods section, since values of the ratio $r$ span several orders of magnitude, we chose to deal with the log-ratio $\log(r)$ instead of $r$. In parallel, we have trained a Transformer~\cite{vaswani2017attention} to achieve the same goal (see details in Methods). For the training data, we have used 100\,000 (log-)ratios starting at the 1\,000\,000th ratio (to lie deep in the ``high-energy'' universal regime; table of zeros retrieved from~\cite{odlyzko_zeros}). Obviously, the Hilbert and Transformer predictors cannot truly predict the actual next term of a sequence of ratios, since the sequence of zeroes behaves effectively like a true random variable. However, they can both exploit the strong correlations between zeroes to offer a better estimate than, say, the average ratio 
$\langle r \rangle$, while not beating the Bayes limit. For the Hilbert predictor, the results naturally depend on the length $d$ of the input sequence. $d=1$ and $d=2$ lead to a mediocre performance, while the performance saturates rapidly with $d$, presumably due to the limited range of the correlations between Riemann zeros or ratios. In the following, we discuss our results for $d=5$.
\FloatBarrier

\begin{figure}[ht]
    \centering
    \includegraphics[width=\textwidth]{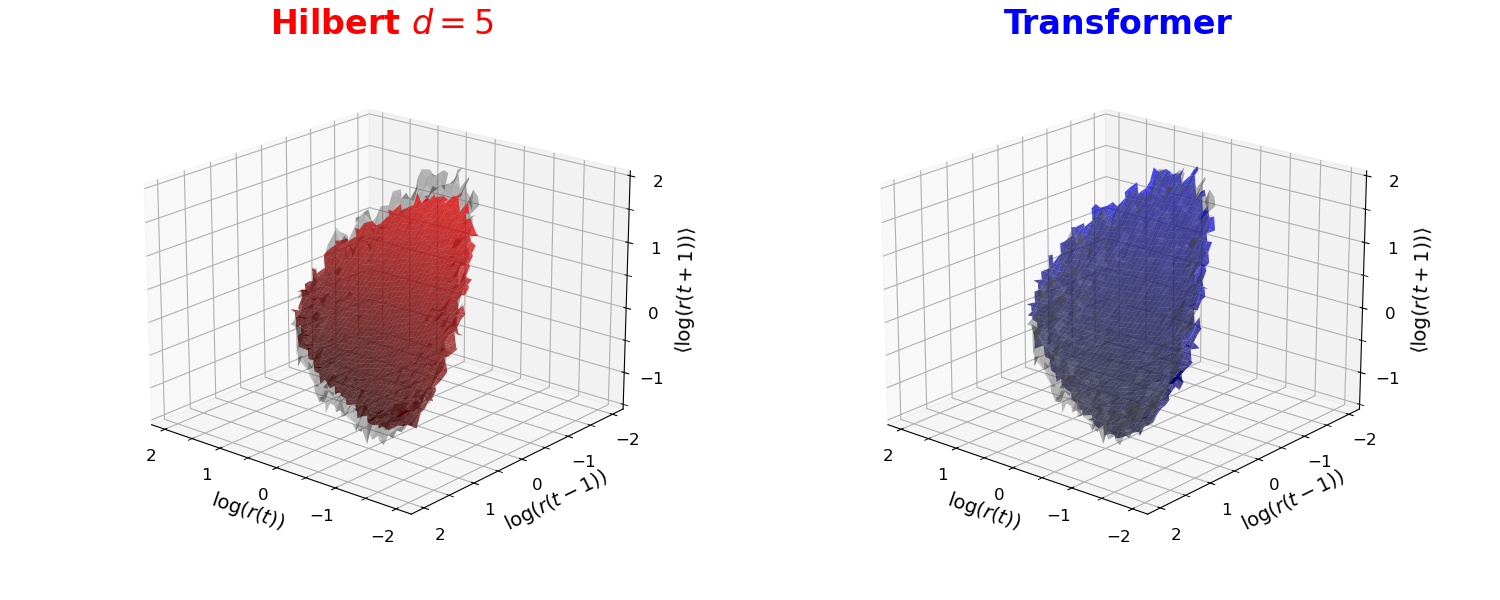}
    \caption{\small \textbf{Correlations of the zeros of the Riemann function.} For the Hilbert ($d=5$; red surface in left panel) and Transformer (blue surface in right panel) predictors, we plot the mean value of the prediction for the next log-ratio, $\langle \log(r(t+1)) \rangle$, obtained for all observed sequences with a given $r(t)$ and $r(t-1)$ (up to the size of the binning box). We have only retained bins containing at least 20 data when exploiting the 200\,000 predictions whose PDF is shown in Fig.~\ref{fig:Riemann1}. The gray surface corresponds to the actual mean of the observed next log-ratio, and is almost indistinguishable from the Hilbert and Transformer predictions. We provide a python code (see Data availability) which allows to reproduce and interact with this figure to change the angle of view (synchronized between both panels). }
    \label{fig:Riemann2}
\end{figure}

In Fig.~\ref{fig:Riemann1}, we show that, although the individual predictions of the Hilbert and Transformer predictors are different, they share the same statistical properties and a strong correlation with the observed ratio. Both predictors produce a much smaller mean error (and a narrower PDF of the error) than when using the mean ratio $\langle r \rangle$ or a ratio randomly drawn in the dataset as a simple predictor for the next ratio. Moreover, the fact that the PDF of the relative and absolute error for the Hilbert and Transformer predictors coincide suggests that both have reached the Bayes limit. In addition, the Pearson normalized correlation coefficient between 200\,000 observed log-ratios (outside the training range) and the corresponding Hilbert and Transformer predictions is found to be $R=0.7$ for both predictors.

Finally, in  Fig.~\ref{fig:Riemann2}, in order to visualize the correlations between ratios captured by both predictors, we plot  the mean value of the prediction for the next log-ratio, $\langle \log(r(t+1)) \rangle$, obtained for all observed sequences with a given $r(t)$ and $r(t-1)$. Again, the results for the Hilbert and Transformer predictors are very similar and are in good agreement with the actual  mean value of  the next observed log-ratio.

In conclusion, the Hilbert and Transformer predictors applied to the zeros of the Riemann function exhibit a very similar performance, and have presumably both reached the Bayes limit.

\FloatBarrier

\subsection*{Credit assignment in interpolative generative modeling}

An interpolating regression function is by construction linear in the labels, with the output being a weighted sum of training data labels. Furthermore, in the Hilbert kernel, the weights do not have any adjustable parameters. This provides a transparent route to credit assignment among different training data fragments. If an autoregressive generative model is used to create a time series $x_t$ (e.g., a musical wave form), then for any given time point $t$, the contribution of each training data point $i$ is given by the corresponding weight $w_i$. For an entire generated piece, the weights of each training data fragment could be averaged to obtain the quantitative contribution of that data point (which could then be used, for example, to apportion a revenue share). Our considerations show that the weight distribution is expected to be long-tailed, with a few fragments contributing large weights and many fragments contributing small weights. A suitable cutoff could be used to truncate the number of contributing samples. This process would also show which fragments of the generated piece are truly innovative, since those fragments will not utilize training data fragments with large weights. 

\section*{Conclusion}

In this work, we have made the case, with rigorous mathematical proofs as well as in-depth applications in three widely different disciplines (behavioral biology, control engineering, and number theory), that a fundamental rethink of the current tool-set of artificial intelligence is possible, where highly parameterized network models are complemented by parameter-free data interpolation. In particular, in some AI applications where an ANN module is currently trained using paired input-output training data, the ANN module could be replaced by a Hilbert kernel. While it is out of the scope of the present work to explore the many ramifications of such an approach, it shows the potential for an entire field of research and applications with a parameter-free interpolation centered approach. 

The potential benefits of replacing hard-to-interpret black-boxes with a theoretically simple and highly transparent approach are manifold. The estimates are straightforward to define, deterministic, and easily reproducible, and do not require any optimization or model selection. This should greatly benefit scientific and biomedical applications by removing the obscuring factors of multi-million parameter black-box networks that investigators have to work hard to design their appropriate architecture and then interpret. It also has ethical-legal utility in permitting proper credit assignment to training data fragments in generative AI, and addressing the concerns of content-creators whose works may be used as training data in generative models. 

In contrast with ANNs, the Hilbert estimator does not require an expensive and lengthy training phase. However, function evaluation is expensive as every training data point is included in every evaluation. We suggest that the function computation may be substantially accelerated. First, the method is suited to GPU parallelization as samples may be distributed among threads for computation of the distance from the evaluation point, and the dot product involved in the distance computation is a vector operation. Secondly, the full sum over the data can be replaced by a sum over smaller random batches drawn from the data, as commonly done during the training process of ANNs. Finally, since  the number of data contributing with significant weight is of order $\ln(n)$, computations at an evaluation point $x$ can be carried out exactly for the nearest $O(\ln(n))$ samples and the further away samples treated approximately by evaluating at the sample location $x_i$ closest to $x$. The requisite quantities corresponding to the distant samples may be pre-computed for each sample location and stored. This is an important topic for future research in both software and hardware architecture for machine learning. 

The success of a purely empirical, data-driven and completely model-free approach in predicting phenomena with statistically optimal performance raises philosophical questions about scientific understanding. Making good predictions is universally accepted to be a fundamental criterion in judging scientific models. If it is possible to make good predictions simply based on past observations, without models, other criteria such as simplicity and insight, while difficult to quantify, must be brought to the forefront in judging the validity and utility of scientific models of complex phenomena. 

\section*{Methods}\label{sec11}

\subsection*{Definitions, assumptions, and problem setup for the Hilbert scheme}

\noindent{\bf Notation, Definitions, Statistical Model} We model the labeled training data set $(x_0,y_0),\dotsc,(x_n,y_n)$ as $n+1$ {\it i.i.d.} observations of a random vector $(X,Y)$ with values in $\R^d \times \R$ for regression, and with values in $\R^d \times \{0,1\}$ for binary classification. Due to the independence property, the collection $X_0,\dotsc,X_n$ has the product density, $\prod_{i=0}^n\rho(x_i)$. We will denote by $\E$ an expectation over the collection of $n+1$ random vectors and by $\E_X$ the expectation over the collection $X_0,\dotsc,X_n$. An expectation over the same collection while holding $X_i=x_i$ will be denoted $\E_{X|x_i}$. 
The regression function $f \colon \R^d \to \R$ is defined as the conditional mean of $Y$ given $X=x$, $f(x) := \E[Y \mid X = x]$ and the conditional variance function is $\sigma^2(x):=\E[|Y-f(X)|^2|X=x]$. $f$ minimizes the expected value of the mean squared prediction error (risk under squared loss), $f = \argmin \mse(h)$ where $\mse(h) := \E[(h(X) - Y)^2]$. Given any regression estimator $\hat{f}(x)$, the corresponding risk can be decomposed as $\E[\mse(\hat{f}(X))] = \mse(f) + \E[ (\hat{f}(X) - f(X)^2]$. The excess risk is given by $\mse(\hat{f})-\mse(f) = \E[(\hat{f}(X) - f(X))^2]$. For a consistent estimator, this excess risk goes to zero as $n\rightarrow \infty$, and we are interested in characterizing the {\it rate} at which it goes to zero with increasing $n$ (note that our sample size is $n+1$ for notational simplicity, but for large $n$, this does not change the rate). 

In the case of binary classification, $Y\in \{0,1\}$ and $f(x) = \P[Y=1 \mid X=x)]$. Let $F \colon \R^d \to \{0,1\}$ denote the Bayes optimal classifier, defined by $F(x) := \theta(f(x) - 1/2)$ where $\theta(\cdot)$ is the Heaviside theta function. This classifier minimizes the risk $\risk(h) := \E[\ind{h(X) \neq Y}] = \P(h(X) \neq Y)$ under zero-one loss. Given the regression estimator $\hat{f}$, we consider the plugin classifier $\hat{F} (x)=\theta(\hat f(x)-\frac{1}{2})$. The classification risk for the plugin classifier $\hat{F}$ is bounded as $\E[\risk(\hat F(x))] - \risk(F(x)) \leq 2\E[|\hat f(x) - f(x)|]\leq 2\sqrt{\E[(\hat{f}(x)-f(x))^2]}$. 

Finally, we define two sequences $a_n,b_n>0$, $n\in\N$, to be asymptotically equivalent for $n\to+\infty$, denoted $a_n\sim_{n\to+\infty} b_n$, if the limit of their ratio exists and $\lim_{n\to\infty} a_n/b_n = 1$.

In summary, our work focuses on the estimation of asymptotic equivalents for $\E[(\hat{f}(x)-f(x))^2]$ and other relevant quantities, as this determines the rate at which the excess risk goes to zero for regression, and bounds the rate at which the excess risk goes to zero for classification.

\vskip 0.2cm
\noindent{\bf Assumptions.} We define the support $\Omega$ of the density $\rho$ as $\Omega=\{x\in \R^d /\rho(x)>0\}$, the closed support $\bar\Omega$ as the closure of $\Omega$, and $\Omega^\circ$ as the interior of $\Omega$. Our results will not assume any compactness condition on $\Omega$ or $\bar\Omega$. The boundary of $\Omega$ is then defined as $\partial\Omega=\bar\Omega\setminus\Omega^\circ$. We assume that $\rho$ has a finite variance $\sigma_\rho^2$. In addition, we will most of the time assume that the density $\rho$ is continuous at the considered point $x\in \Omega^\circ$, and in some cases, $x\in \partial\Omega\cap\Omega$. 

For the regression function $f$, we will obtain results assuming either of the following conditions
\begin{itemize}
\item $C_{\rm Cont}^f$: $f$ is continuous at the considered $x$,
\item $C_{\rm Holder}^f$: for all $x\in\Omega^\circ$, there exist $\alpha_x>0$, $K_x>0$, and $\delta_x>0$, such that  \\
$x'\in\Omega$ and  $\|x-x'\|\leq \delta_x\implies |f(x) - f(x')| \leq K_x\,\|x-x'\|^{\alpha_x}$ \\(local H\"older smoothness condition),
\end{itemize}
where condition $C_{\rm Holder}^f$ is obviously stronger than $C_{\rm Cont}^f$.
In addition, we will always assume a growth condition for the regression function $f$: 
\begin{itemize}
\item $C_{\rm Growth}^f$: $\int \rho(y)\frac{f^2(y)}{1+\|y\|^{2d}}\,d^dy<\infty$.
\end{itemize}

As for the variance function $\sigma$, we will obtain results assuming either that $\sigma$ is bounded or satisfies a growth condition similar to the one above:
\begin{itemize}
\item $C_{\rm Bound}^\sigma$: there exists $\sigma_0^2\geq 0$, such that, for all $x\in \Omega$, we have $\sigma^2(x)\leq\sigma_0^2$,
\item $C_{\rm Growth}^\sigma$: $\int \rho(y)\frac{\sigma^2(y)}{1+\|y\|^{2d}}\,d^dy<\infty$.
\end{itemize}
When we will assume condition $C_{\rm Growth}^\sigma$ (obviously satisfied when  $\sigma^2$ is bounded), we will also assume a continuity condition $C_{\rm Cont}^\sigma$ for $\sigma$ at the considered $x$.

Note that all our results can be readily extended in the case where $x\in \partial\Omega=\bar\Omega\setminus\Omega^\circ$ but keeping the condition $\rho(x)>0$ (i.e., $x\in\partial\Omega\cap\Omega$), and assuming the continuity at $x$ of $\rho$ as seen as a function restricted to $\Omega$, i.e., $\lim_{y\in \Omega\to x}\rho(y)=\rho(x)$. Useful examples are when the support $\Omega$ of $\rho$ is a $d$-dimensional sphere or hypercube and $x$ is on the surface of $\Omega$  (but still with $\rho(x)>0$). To guarantee these results for $x\in \partial\Omega\cap\Omega$, we need also to assume the continuity at $x$ of $f$, and assume that $\Omega$ is smooth enough near $x$, so that there exists a strictly positive local solid angle $\omega_x$ defined by
\begin{eqnarray}
\omega_x&=&    \lim_{r\to 0}\frac{1}{V_d\rho(x)r^d}\int_{\|x-y\|\leq r}\rho(y)\,d^dy\\
&=& \lim_{r\to 0}\frac{1}{V_dr^d}\int_{y\in\Omega/\|x-y\|\leq r}\,d^dy,
\end{eqnarray}
where $V_d=S_d/d=\pi^{d/2}/\Gamma(d/2+1)$ is the volume of the unit ball in $d$ dimensions, and the second inequality results from the continuity of $\rho$ at $x$. 
If $x\in \Omega^\circ$, we have $\omega_x=1$, while for $x\in \partial\Omega$, we have $0\leq \omega_x\leq 1$. For instance, if $x$ is on the surface of a sphere or on the interior of a face of a hypercube (and in general, when the boundary near $x$ is locally a hyperplane), we have $\omega_x=\frac{1}{2}$. If $x$ is a corner of a hypercube, we have $\omega_x=\frac{1}{2^d}$. From our methods of proof presented in the appendix, it should be clear that all our results for $x\in \Omega^\circ$ perfectly generalize to any $x\in \partial\Omega\cap\Omega$ for which $\omega_x>0$, by simply replacing $V_d$ whenever it appears in our different results by $\omega_x V_d$.

\vskip 0.2cm
\noindent{\bf Hilbert kernel interpolating estimator and Bias-Variance decomposition.} The Hilbert Kernel estimator at a given point, is a linearly weighted sum of the training labels (see Eq.~(\ref{HilbertDef})). The weights $w_i(x)$ are also called Lagrange functions in the interpolation literature and satisfy the interpolation property $w_i(x_j)=\delta_{ij}$, where $\delta_{ij}=1$, if $i=j$, and 0 otherwise. At any given point $x$, they provide a partition of unity so that $\sum_{i=0}^n w_i(x)=1$. The mean squared error between the Hilbert estimator and the true regression function has a bias-variance decomposition (using the {\it i.i.d} condition and the earlier definitions). Indeed, we have
\begin{eqnarray}
\hat{f}(x)-f(x) =& \sum_{i=0}^n& w_i(x)[f(x_i)-f(x)] + \nonumber\\ &\sum_{i=0}^n& w_i(x)[y_i-f(x_i)],
\end{eqnarray}
leading to
\begin{eqnarray}
    \E[(\hat{f}(x)-f(x))^2] = {\cal B}(x) + {\cal V}(x),\quad\quad\quad\quad\\
    (Bias)~~{\cal B}(x)= \E_X\Bigl[\Bigl(\sum_{i=0}^n w_i(x) [f(x_i)-f(x)]\Bigr)^2\Bigr],\\
    (Variance)~~{\cal V}(x)= \E\Bigl[\sum_{i=0}^n w_i^2(x) [y_i-f(x_i)]^2\Bigr],\quad \\
    = \E_X\Bigl[\sum_{i=0}^n w_i^2(x) \sigma^2(x_i) \Bigr].\quad\quad\quad
\end{eqnarray}

The present work derives asymptotic behaviors and bounds for the regression and classification risk of the Hilbert estimator for large sample size $n$. These results are derived by analyzing the large-$n$ behaviors of the bias and variance terms, which in turn depend on the behavior of the moments of the weights or the Lagrange functions $w_i(x)$. For all these quantities, asymptotically equivalent forms are derived. The proofs exploit a simple integral form of the weight function and details are provided in the appendix, while the body of the paper provides the results and associated discussions. 

\subsection*{Rigorous results for Hilbert kernel interpolation}

\renewcommand{\thetheorem}{\arabic{theorem}}
\subsection*{The weights, variance, and bias terms}

\subsubsection*{Moments of the weights: large $n$ behavior}
\label{subsubsection:weights}
In this section, we consider the moments and the distribution of the weights $w_i(x)$ at a given point $x$. The first moment is simple to compute. Since the weights sum to 1 and $X_i$ are {\it i.i.d}, it follows that $\E_{X}[w_i(x)]$ are all equal and thus 
$\E_{X}[w_i(x)]=(n+1)^{-1}$. The other moments are much less trivial to compute, and we prove the following theorem in the appendix~\ref{prooftheo1}: 

\begin{theorem}\label{theo1}
For $x\in\Omega^\circ$ (so that $\rho(x)>0$), we assume $\rho$ continuous at $x$. Then, the moments of the weight $w_0(x)$ satisfy the following properties:
\begin{itemize}
    \item For $\beta>1$: 
\begin{equation}
    \E\left[w_0^\beta(x)\right]\underset{n\to +\infty  }{\sim}\frac{1}{(\beta-1) n\ln(n)}.
\end{equation}
\item The expected value of the entropy $S(x)$ (moment for ``$\beta=1^-$'') behaves as,
\begin{equation}
    \E\left[S(x)\right] = (n+1)\, \E\left[-w_0(x)\ln(w_0(x)) \right]\underset{n\to +\infty  }{\sim}\frac{1}{ 2}\ln(n).
\end{equation}
\item For \,$0<\beta<1$: \,defining
$	\kappa_\beta(x):=\int \frac{\rho(x+y)}{||y||^{\beta d}}\,d^dy<\infty$, 
we have
\begin{eqnarray}\label{eqbeta1}
\E\left[w_0^\beta(x)\right]\underset{n\to +\infty  }{\sim}\frac{\kappa_\beta(x)}{(V_d\rho(x) n\ln(n))^\beta}.
\end{eqnarray}
\item For $\beta<0$: \,all moments for  $\beta\leq -1$ are infinite, and the moments of order $-1<\beta<0$ satisfy 
\begin{equation}
\E\left[w_0^\beta(x)\right]\leq 1+n\,\kappa_{|\beta|}(x)\kappa_{\beta}(x),
\end{equation}
so that a sufficient condition for its existence is $\kappa_{\beta}(x)=\int {\rho(x+y)}{||y||^{|\beta|d}}\,d^dy<\infty$.
\end{itemize}
\end{theorem}

Heuristically, the behavior of these moments is consistent with the random variable $W=w_0(x)$ having a probability density function satisfying a scaling relation $P(W)=\frac{1}{W_n} p\left(\frac{W}{W_n} \right)$, with the scaling function $p$ having the universal tail (i.e., independent of $x$ and $\rho$), $p(w)\underset{w\to +\infty  }{\sim}w^{-2}$, and a scale $W_n$ expected to vanish with $n$, when $n\to+\infty$. Exploiting Theorem~\ref{theolag} below, we will later give an additional heuristic argument suggesting this scaling and universal power-law tail for the probability density function of $W=w_0(x)$. 
With this assumption, we can determine the scale $W_n$ by imposing the exact condition $\E[W]=1/(n+1)\sim 1/n$:
\begin{eqnarray}
\E[W]&=&\frac{1}{W_n}\int_0^1 p\left(\frac{W}{W_n} \right)W\,dW,\\&=&W_n \int_0^{\frac{1}{W_n}}p(w)w\,dw,\\
&\sim& W_n\int_1^{\frac{1}{W_n}} \frac{dw}{w}  \sim-W_n \ln(W_n)\sim \frac{1}{n},
\end{eqnarray}
leading to $W_n\sim \frac{1}{n\ln(n)}$. Then, the moment of order $\beta>1$ is given by
\begin{eqnarray}
	\E[W^{\beta}]&=&\frac{1}{W_n}\int_0^1 p\left(\frac{W}{W_n} \right)W^{\beta}\,dW,\\&\sim& W_n\int_0^1W^{\beta-2}\,dW,\\&
	\underset{n\to +\infty  }{\sim}&\frac{1}{(\beta-1) n\ln(n)},\label{heuri1}
\end{eqnarray}
which indeed coincides with the first result of Theorem~\ref{theo1}.

An interesting question which naturally arises concerns the number, ${\cal N}$, of relevant data configurations $x_i$ which are “typically'' involved in the prediction, given an input $x$ (effective number of degrees of freedom). To that end, we define
\begin{equation}
    {\cal N}(x)= \frac{1}{\sum_{i=0}^n w_i^2(x)}.
\end{equation}
If the weights are equidistributed over ${\cal N}$ data, one indeed obtains ${\cal N}(x)=1/({\cal N}\times 1/{\cal N}^2)={\cal N}$. Theorem~\ref{theo1} proves that we have,
\begin{equation}
\E\left[\frac{1}{{\cal N}(x)}\right] =(n+1)\,\E[W^2] \underset{n\to +\infty}{\sim}\frac{1}{\ln(n)}.
\end{equation}
This exact result suggests that the typical number of data contributing to a prediction is on average $\ln(n)$, at least when evaluated via $\E\left[\frac{1}{{\cal N}(x)}\right] $. However, since the distribution of the weights has a fat tail going down to the small scale  $W_n$, we expect that $\E\left[{\cal N}(x)\right]$ may grow faster with $n$ than $\ln(n)$.
An alternative way to quantify the effective number of data involved in a Hilbert prediction exploits the information entropy, 
\begin{equation}
    S(x)= -\sum_{i=0}^n w_i(x)\log[w_i(x)].
\end{equation}
If the weights are equidistributed over ${\cal N}$ data, one obtains $S=-{\cal N}\times 1/{\cal N}\log(1/{\cal N}) =\log({\cal N})$, and ${\rm e}^S={\cal N}$ indeed represents the number of contributing samples. By using the present heuristic argument for the distribution of the weights, one can easily evaluate the expectation value of the entropy, 
\begin{equation}
\E[S(x)]\underset{n\to +\infty} = -(n+1)\,\E[W\log W] \approx -(n+1)\int_{W_n}^1 \frac{W_n}{W^2}\times W\log W\, dW {\sim}\frac{1}{2}\ln(n),
\end{equation}
which recovers the second part of Theorem~\ref{theo1}.
Hence, using the entropy to evaluate the typical number of data contributing to a prediction now suggests a scale ${\cal N}\sim \sqrt{n}$ (times subleading terms), which grows much faster with $n$ than the previous scale $\ln(n)$ evaluated from $\E\left[\frac{1}{{\cal N}(x)}\right] $. Again, the occurrence of these two different scales is due to the wide power-law distribution of the weights. Yet, we will see that in terms of the variance term decay, the relevant scale for the number of weights effectively contributing to the prediction is indeed  ${\cal N}\sim \ln(n)$ (see Theorem~\ref{theovar2}). On the other hand, the entropy provides the information associated with the weight distribution.

Our heuristic argument also suggests that in the case $0<\beta<1$, we have
\begin{eqnarray}
	\E[W^\beta]&=&\frac{1}{W_n}\int_0^1 p\left(\frac{W}{W_n} \right)W^{\beta}\,dW,\\ 
    &\sim& W_n^\beta \int_0^\frac{1}{W_n} p\left(w \right)w^\beta \,dw,\\
	&\underset{n\to +\infty}{\sim}& \frac{\int_0^{+\infty} p\left(w \right)w^\beta \,dw}{(n\ln(n))^\beta} ,\label{heuri2}
\end{eqnarray}
where the last integral converges since $p(w)\underset{w\to +\infty  }{\sim}w^{-2}$ and $\beta<1$. This result is perfectly consistent with \eq{eqbeta1} in Theorem~\ref{theo1}, and suggests that $\int_0^{+\infty} p\left(w \right)w^\beta \,dw=\frac{\kappa_\beta(x)}{(V_d\rho(x))^\beta}$. Interestingly, for $0<\beta<1$, and contrary to the case $\beta>1$, we find that the large $n$ equivalent of the moment is not universal and depends explicitly on $x$ and the density $\rho$.
As for moments of order $-1<\beta<0$, we conjecture that they are still given by \eq{eqbeta1} (and equivalently, by \eq{heuri2}) provided they exist, and that the sufficient condition for their existence $\kappa_{\beta}(x)<\infty$ is hence also necessary, since $\kappa_{\beta}(x)$ also appears in \eq{eqbeta1}. The fact that moments for $\beta\leq -1$ do not exist strongly suggests that $p(0)>0$. In fact, \eq{heuri2}) also suggests that all moments for $-1<\beta<0$ exist if and only if  $0<p(0)<\infty$. In Fig.~\ref{figdist} (see also the last section of the appendix~\ref{prooftheo1}), we present numerical simulations confirming our scaling ansatz, the fact that $p(w)\underset{w\to +\infty  }{\sim}w^{-2}$, and the quantitative prediction for $W_n$.

It is shown in Devroye {\it et al.}~\cite{devroye1998hilbert} that the Hilbert kernel regression estimate does not converge almost surely ({\it a.s.}) by giving a specific example. Insight can be gained into this lack of almost sure convergence by considering the weight function $w_0(x)$, for a sequence of independent training sample sets of increasing size $n+1$. 
Let the corresponding sequence of weights be denoted as $\omega_n\in[0,1]$. From Theorem~\ref{theo1}, it is clear that $\omega_n$ converges to zero in probability, since the following Chebyshev bound holds (analogous to the bound on the regression risk):
\begin{equation}\label{omegan}
    \P(\omega_n>\varepsilon)\leq \frac{1+\delta}{\varepsilon^2n\ln(n)},
\end{equation} 
for arbitrary $\varepsilon>0$ and $\delta>0$, and for $n$ larger than some constant $N_{x,\delta}$. Alternatively, one can exploit the fact that $\E[\omega_n]=\frac{1}{n+1}$, leading to
$\P(\omega_n>\varepsilon)\leq \frac{1}{\varepsilon\,  n}$,
which is less stringent than \eq{omegan} as far as the $n$-dependence is concerned, but is more stringent for the  $\varepsilon$-dependence of the bounds.

Let us show heuristically that $\omega_n$ does not converge {\it a.s.} to zero. Consider the infinite sequence of events ${\cal E}_n\equiv \{\omega_n>\varepsilon\}$, $n\in \N$, and the corresponding infinite sum $\sum_n\P({\cal E}_n) = \sum_n\P(\omega_n>\varepsilon)$. Exploiting our previous heuristic argument for the scaling form of the distribution of weights, we obtain
\begin{eqnarray}
    \P(\omega_n>\varepsilon)&=&\int_{\varepsilon}^1 \frac{1}{W_n}p\left(\frac{W}{W_n}\right)\, dW,\\&\sim& \int_{\varepsilon\,  n\ln n}^{n\ln n} \frac{d w}{w^2} \sim \frac{1-\varepsilon}{\varepsilon\, n\ln(n)}.
\end{eqnarray}
Since $\sum_{n=2}^{N} \frac{1}{n\ln (n)}\sim\ln(\ln(N))$ is a divergent series, a Borel-Cantelli argument suggests that an infinite number of the events ${\cal E}_n$ (i.e., $\omega_n>\varepsilon$) must occur, which implies that $\omega_n$ does not converge {\it a.s.} to 0. 
Note that the weights are equal to 1 at the data points due to the interpolation condition, so that large weights occasionally occur, causing the lack of {\it a.s.} convergence.

\subsubsection*{Lagrange function: scaling limit}

Due to the {\it i.i.d.} condition, the indices $i$ are exchangeable, and we set $i=0$ for the computation of the expected Lagrange function $L_0(x)=\E_{X|x_0} [w_0(x)]$. Thus, the sample point $i=0$ at $x_0$ is held fixed, and the other ones are averaged over in computing the expected Lagrange function. 
For $x_0\ne x$ kept fixed, we  have $\lim_{n\rightarrow \infty}L_0(x)= 0$. However, we show in the appendix~\ref{prooflag}  that $L_0(x)$ takes a very simple form when taking a specific scaling limit:

\begin{theorem}
\label{theolag}
For $x\in\Omega^\circ$, we assume $\rho$ continuous at $x$. 
Then, in the limit 
(denoted by $\lim_Z$), $n\to+\infty$, $\|x-x_0\|^{-d}\to +\infty$  (i.e., $x_0\to x$), and such that $z_x(n,x_0)=V_d\rho(x)\|x-x_0\|^d n\log(n)\to Z$, the Lagrange function $L_0(x)=\E_{X|x_0} [w_0(x)]$ converges to a proper limit,
\begin{equation}
    \lim_{Z} L_0(x)= \frac{1}{1+Z}.
\end{equation}
\end{theorem}

Exploiting Theorem~\ref{theolag}, we can use a simple heuristic argument to estimate the tail of the distribution of the random variable $W=w_0(x)$.
Indeed, approximating $L_0(x)$ for finite but large $n$ by its asymptotic form $\frac{1}{1+z_x(n,x_0)}$, with $z_x(n,x_0)=V_d\rho(x) n\log(n)\|x-x_0\|^d$, we obtain 
\begin{eqnarray}
\int_W^{1}P(W')\,dW' &\sim&\int \rho(x_0)\,\theta\left(\frac{1}{1+V_d\rho(x) n\log(n)\|x-x_0\|^d}-W\right)\,d^dx_0,\\
    &\sim& V_d\rho(x)\int_0^{+\infty} \theta\left(\frac{1}{1+V_d\rho(x) n\log(n)\, u}-W\right)\,du,\\
    &\sim& \frac{1}{n\ln(n)W}\quad\implies   P(W)\sim \frac{1}{n\ln(n)W^2},\label{argtailP}
\end{eqnarray}
where $\theta(.)$ is the Heaviside function. 
This heuristic result is again perfectly consistent with our guess of the previous section that  $P(W)=\frac{1}{W_n} p\left(\frac{W}{W_n} \right)$, with the scaling function $p$ having the universal tail,  $p(w)\underset{w\to +\infty  }{\sim}w^{-2}$, and a scale $W_n\sim \frac{1}{n\ln(n)}$. Indeed, in this case and in the limit $n\to+\infty$, we  obtain that  $P(W)\sim\frac{1}{W_n}\left(\frac{W_n}{W}\right)^2\sim\frac{W_n}{W^2}\sim \frac{1}{n\ln(n)W^2}$, which is identical to the result of \eq{argtailP}.

\subsubsection*{The variance term}

A simple application of the result of Theorem~\ref{theo1} for $\beta=2$  (see appendix~\ref{proofvar}) allows us to bound the variance term ${\cal V}(x)=\E\Bigl[\sum_{i=0}^n w_i^2(x) [y_i-f(x_i)]^2\Bigr]$ for a bounded variance function $\sigma^2$:
\begin{theorem}\label{theovar1}
For $x\in\Omega^\circ$, $\rho$ continuous at $x$, $\sigma^2\leq\sigma_0^2$, and for any $\varepsilon>0$,
there exists a constant $N_{x,\varepsilon}$ such that for $n\geq N_{x,\varepsilon}$, we have
\begin{equation}
   {\cal V}(x)\leq (1+\varepsilon)\frac{\sigma_0^2}{\ln(n)}.
\end{equation}
\end{theorem}

Relaxing the boundedness condition for $\sigma$, but assuming the continuity of $\sigma^2$ at $x$ along with a growth condition, allows us to obtain a precise asymptotic equivalent of ${\cal V}(x)$, when $n\to+\infty$:

\begin{theorem}\label{theovar2}
For $x\in\Omega^\circ$, $\sigma(x)>0$, $\rho\sigma^2$ continuous at $x$,  and assuming the condition $C_{\rm Growth}^\sigma$, i.e., $\int \rho(y)\frac{\sigma^2(y)}{1+\|y\|^{2d}}\,d^dy<\infty$, we have
\begin{equation}\label{eq:varn}
   {\cal V}(x)\underset{n\to +\infty  }{\sim}\frac{\sigma^2(x)}{\ln(n)}.
\end{equation}
\end{theorem}

Note that if the mean variance $\int \rho(y)\sigma^2(y)\,d^dy<\infty$, which is in particular the case when $\sigma^2$ is bounded over $\Omega$, the condition $C_{\rm Growth}^\sigma$ is, in fact, automatically satisfied. Eq.~(\ref{eq:varn}) shows that the effective number of data over which the noise is averaged is ${\cal N}\sim \ln(n)$, as mentioned in the discussion of our heuristic arguments, below Theorem~\ref{theo1}.

\subsubsection*{The bias term}

In appendix~\ref{proofbias}, we prove the following three theorems for the bias term ${\cal B}(x)= \E_X\Bigl[\Bigl(\sum_{i=0}^n w_i(x) [f(x_i)-f(x)]\Bigr)^2\Bigr]$.
\begin{theorem}\label{theobias}
For $x\in\Omega^\circ$ (so that $\rho(x)>0$), we assume that  $\rho$ is continuous at $x$, and the conditions 
\begin{itemize}
\item $C_{\rm Growth}^f$: $\int \rho(y)\frac{f^2(y)}{1+\|y\|^{2d}}\,d^dy<\infty$,
\item $C_{\rm Holder}^f$: there exist $\alpha_x>0$, $K_x>0$, and $\delta_x>0$, such that we have the local Hölder condition for $f$:
\\$x'\in\Omega$ and  $\|x-x'\|\leq \delta_x\implies |f(x) - f(x')| \leq K_x\,\|x-x'\|^{\alpha_x}$.
\end{itemize}
Moreover, we define $\kappa(x)= \int \rho(x+y)\frac{f(x+y)-f(x)}{||y||^{d}}\,d^dy$, with $|\kappa(x)|<\infty$ under conditions $C_{\rm Growth}^f$ and $C_{\rm Holder}^f$.

Then, for $\kappa(x)\neq 0$, the bias term ${\cal B}(x)$ satisfies
\begin{eqnarray}\label{biaskne0}
    {\cal B}(x)\underset{n\to +\infty }{\sim}\left(\E\left[\hat f(x)\right]-f(x)\right)^2,\\
 \qquad {\rm with}\quad  \E\left[\hat f(x)\right]-f(x) \underset{n\to +\infty  }{\sim}\frac{\kappa(x)}{V_d\rho(x)\ln(n)}.\label{limefhat}
\end{eqnarray}
In the non-generic case $\kappa(x)=0$, we have the weaker result
\begin{equation}\label{Bcasetheo}
{\cal B}(x)=\left\{\begin{array}{lr}
        O\left(n^{-\frac{2\alpha_x}{d}}(\ln(n))^{-1-\frac{2\alpha_x}{d}}\right), & \text{for }\, d>2\alpha_x\vspace{0.25cm}\\ 
       O\left(n^{-1}(\ln(n))^{-1}\right), & \text{for }\,d=2\alpha_x\vspace{0.35cm}\\
        O\left(n^{-1}(\ln(n))^{-2}\right), & \text{for }\, d<2\alpha_x
        \end{array}\right.
\end{equation}
\end{theorem}

Note that $\kappa(x)=0$ is non-generic but can still happen, even if $f$ is not constant. For instance, if $\Omega$ is a sphere centered at $x$ or $\,\Omega=\R^d$, if $\rho(x+y)=\hat\rho(||y||)$ is isotropic around $x$, and if $f_x:y\mapsto f(x+y)$ is an odd function of $y$, then we indeed have $\kappa(x)=0$ at this symmetric point $x$.

Interestingly, for $\kappa(x)\neq 0$, \eq{biaskne0} shows that the bias ${\cal B}(x)$ is asymptotically dominated by the square of $\E[\hat f(x)] - f(x)$, showing that the fluctuations of $\E[\hat f(x)]-\sum_{i=0}^n w_i(x) f(x_i)$ are negligible compared to $\E[\hat f(x)]-f(x)$, in the limit $n\to+\infty$ and for $\kappa(x)\neq 0$.

One can relax the local Hölder condition, but at the price of a weaker estimate for ${\cal B}(x)$, which will however be enough to obtain strong results for the regression and classification risks (see below):

\begin{theorem}\label{theobiasbis}
For $x\in\Omega^\circ$, we assume $\rho$ and $f$ continuous at $x$, and the growth condition $C_{\rm Growth}^f$: $\int \rho(y)\frac{f^2(y)}{1+\|y\|^{2d}}\,d^dy<\infty$. Then, the bias term satisfies  
\begin{equation}
    {\cal B}(x)=o\left( \frac{1}{\ln(n)}\right),
\end{equation}
or equivalently, for any $\varepsilon>0$, there exists  $N_{x,\varepsilon}$, such that 
\begin{equation}
    {\cal B}(x)\leq  \frac{\varepsilon}{\ln(n)}, \quad {\rm for }~ n\geq N_{x,\varepsilon}.
\end{equation}
\end{theorem}

Let us now consider a point $x\in\partial\Omega$ for which we have $\rho(x)=0$ (note that $x\in\partial\Omega$ does not necessarily imply $\rho(x)=0$).
In appendix~\ref{proofbias}, we show the following theorem for the expectation value of the estimator $\hat f(x)$ in the limit $n\to+\infty$:

\begin{theorem}\label{theorhonon0}
For $x\in\partial\Omega$ such that $\rho(x)=0$, we assume that $f$ and $\rho$ satisfy the conditions
\begin{itemize}
\item $C_{\rm Growth}^f$: $\int \rho(y)\frac{|f(y)|}{1+\|y\|^{d}}\,d^dy<\infty$,
\item $C_{\rm Holder}^f$: there exist $\alpha_x^f>0$, $K_x^f>0$, and $\delta_x^f>0$, such that we have the local Hölder condition for $f$:
\\$x'\in\Omega$ and  $\|x-x'\|\leq \delta_x^f\implies |f(x) - f(x')| \leq K_x^f\,\|x-x'\|^{\alpha_x^f}$,
\item $C_{\rm Holder}^\rho$: there exist $\alpha_x^\rho>0$, $K_x^\rho>0$, and $\delta_x^\rho>0$, such that we have the local Hölder condition for $\rho$:
\\$x'\in\Omega$ and  $\|x-x'\|\leq \delta_x^\rho\implies |\rho(x')| \leq K_x^\rho\,\|x-x'\|^{\alpha_x^\rho}$.
\end{itemize}
Moreover, we define $\kappa(x)= \int \rho(x+y)\frac{f(x+y)-f(x)}{||y||^{d}}\,d^dy$ ~($|\kappa(x)|<\infty$ under conditions $C_{\rm Growth}^f$ and $C_{\rm Holder}^f$), and
$\lambda(x)= \int \frac{\rho(x+y)}{||y||^{d}}\,d^dy$ ~($0<\lambda(x)<\infty$ under condition $C_{\rm Holder}^\rho$). 
Then,
\begin{eqnarray}
   \lim_{n\to+\infty}\E[\hat f(x)]-f(x)=
   \frac{\kappa(x)}{\lambda(x)}.
\end{eqnarray}
\end{theorem}

Hence, in the generic case $\kappa(x)\ne 0$ (see Theorem~\ref{theobias} and the discussion below it) and under condition $C_{\rm Holder}^\rho$, we find that the bias does not vanish when $\rho(x)=0$, and that the estimator $\hat f(x)$ does not converge to $f(x)$. When $\rho(x)=0$, the scarcity of data near the point $x$ indeed prevents the estimator from converging to the actual value of $f(x)$. In appendix~\ref{proofbias}, we show an example of a  density $\rho$ continuous at $x$ and such that $\rho(x)=0$, but not satisfying the condition $C_{\rm Holder}^\rho$, and for which  $\lim_{n\to+\infty}\E[\hat f(x)]=f(x)$, even if $\kappa(x)\neq 0$.

\subsection*{Asymptotic equivalent for the regression risk}

In appendix~\ref{prooftheorisk}, we prove the following theorem establishing the asymptotic rate at which the excess risk goes to zero with large sample size $n$ for Hilbert kernel regression, under mild conditions that do not require $f$ or $\sigma$ to be bounded, but only to satisfy some growth conditions:

\begin{theorem}
\label{regressionrisk}
For $x\in\Omega^\circ$, we assume $\sigma(x)>0$, $\rho$, $\sigma$, and $f$ continuous at $x$, and the growth conditions~ 
$C_{\rm Growth}^\sigma$: $\int \rho(y)\frac{\sigma^2(y)}{1+\|y\|^{2d}}\,d^dy<\infty$ ~and~
$C_{\rm Growth}^f$: $\int \rho(y)\frac{f^2(y)}{1+\|y\|^{2d}}\,d^dy<\infty$.

Then the following statements are true: 
\begin{itemize}
    \item The excess regression risk at the point $x$ satisfies
\begin{equation}
    \E[(\hat{f}(x)-f(x))^2]\underset{n\to +\infty  }{\sim} \frac{\sigma^2(x)}{\ln(n)}.
\end{equation}

\item The Hilbert kernel estimate converges pointwise to the regression function in probability. More specifically, for any $\delta>0$, there exists a constant $N_{x,\delta}$, such that  for any $\varepsilon>0$, we have the following Chebyshev bound, valid for $n\geq N_{x,\delta}$
\begin{equation}
\P[|\hat{f}(x)-f(x)| \geq \varepsilon]\leq \frac{1+\delta}{\varepsilon^2}\, \frac{\sigma^2(x)}{\ln(n)}.
\end{equation}
\end{itemize}

\end{theorem}

This theorem is a consequence of the corresponding asymptotically equivalent forms of the variance and bias terms presented above. Note that as long as $\rho(x)>0$, the variance term dominates over the bias term and the regression risk has the same form as the variance term.

\subsection*{Rates for the plugin classifier}
In appendix~\ref{proofclassrisk}, we prove the following theorem bounding the asymptotic rate at which the classification risk $\delta\risk(x)=\E[\risk(\hat F(x))] - \risk(F(x))$ goes to zero with large sample size $n$ for Hilbert kernel regression:
\begin{theorem}
\label{theoclass}
For $x\in\Omega^\circ$, we assume $\sigma(x)>0$, $\rho$, $\sigma$, and  $f$ continuous at $x$. Then, the classification  risk $\delta\risk(x)=\E[\risk(\hat F(x))] - \risk(F(x))$ vanishes for $n\to+\infty$. 

More precisely, for any $\varepsilon>0$, there exists $N_{x,\varepsilon}$, such that for any $n\geq N_{x,\varepsilon}$, 
\begin{equation}\label{alpha0risk}
    0\leq \delta\risk(x)\leq 2(1+\varepsilon) \frac{\sigma(x)}{\sqrt{\ln(n)}},\\
\end{equation}

In addition, for any $0<\alpha<1$, the general inequality 
\begin{equation}
\delta\risk(x) \leq 2|f(x)-1/2|^{1-\alpha}\, {\E\left[|\hat f(x) - f(x)|^{2}\right]^\frac{\alpha}{2}}, 
\end{equation}
holds unconditionally and, for $n\geq  N_{x,\varepsilon}$, leads to
\begin{equation}\label{alpharisk}
0\leq \delta\risk(x) \leq 2|f(x)-1/2|^{1-\alpha}\,(1+\varepsilon)^\alpha\frac{\sigma^\alpha(x)}{(\ln(n))^\frac{\alpha}{2}}.
\end{equation}
\end{theorem}

For $0<\alpha<1$, \eq{alpharisk} is weaker than \eq{alpha0risk} in terms of its dependence on $n$, but explicitly shows that the classification  risk vanishes for $f(x)=1/2$. This theorem does not require any growth condition for $f$ or $\sigma$, since both functions take values in $[0,1]$ in the classification context.

\subsection*{Extrapolation behavior outside the support of $\rho$}

We now take the point $x$ outside the closed support $\bar\Omega$ of the distribution $\rho$ (which excludes the case  $\Omega=\R^d$). We are interested in the behavior of $\E\left[\hat f(x)\right]$ as $n\to+\infty$. In  appendix~\ref{proofout} we prove:
\begin{theorem}
\label{theoout}
For $x\notin\bar\Omega$, we assume the growth condition $\int \rho(y)\frac{|f(y)|}{1+\|y\|^{d}}\,d^dy<\infty$. Then,  
\begin{equation}
\hat f_\infty(x):=\lim_{n\to+\infty} \E\left[\hat f(x)\right]=\frac{\int{\rho(y)f(y)\|x-y\|^{-d}} \,d^dy}{\int\rho(y)\|x-y\|^{-d} \,d^dy},
\end{equation}
and $\hat f_\infty$ is continuous at all $x\notin\bar\Omega$.

In addition, if  $\int\rho(y)|f(y)|\,d^dy<\infty$, and defining $d(x,\Omega)>0$ as the distance between $x$ and $\Omega$, we have 
\begin{equation}\label{larged}
\lim_{d(x,\Omega)\to+\infty}\hat  f_\infty(x)=\int\rho(y)f(y)\,d^dy.
\end{equation}
Finally, we consider $x_0\in\partial \Omega$ such that  $\rho(x_0)>0$ (i.e., $x_0\in \partial \Omega\cap\Omega$), and  assume that $f$ and $\rho$, seen as functions restricted to $\Omega$, are continuous at $x_0$, i.e., $\lim_{y\in\Omega\to x_0} \rho(y)=\rho(x_0)$ and $\lim_{y\in\Omega\to x_0} f(y)=f(x_0)$. We also assume  that the local solid angle  $\omega_0=\lim_{r\to 0}\frac{1}{V_d\rho(x_0)r^d}\int_{\|x_0-y\|\leq r}\rho(y)\,d^dy$   exists and satisfies $\omega_0>0$.
 Then,
\begin{equation}\label{contfinf}
\lim_{x\notin \bar\Omega\to x_0}\hat f_\infty(x)=f(x_0).
\end{equation}
\end{theorem}

\eq{larged} shows that far away from $\Omega$ (which is possible to realize, for instance, when $\Omega$ is bounded), $\hat f_\infty(x)$ goes smoothly to the $\rho$-mean of $f$. Moreover, \eq{contfinf} establishes a continuity property for the extrapolation $\hat f_\infty$ at $x_0\in\partial \Omega\cap\Omega$ under the stated conditions (remember that for $x\in \Omega^\circ$, we have $\lim_{n\to+\infty} \E\left[\hat f(x)\right]=f(x)$; see Theorem~\ref{theobias}, and in particular \eq{limefhat}).

\newpage

\subsection*{Fish trajectory training data set details}

In this work, we have exploited the fish data published in~\cite{papaspyros2023biohybrid,papaspyros2023biogap}. The fish are swimming in shallow water, and only their position and heading in the horizontal plane of the tank have been recorded so that the system is effectively 2-dimensional. All trajectories have been subsampled from the original 30 FPS, resulting in a considered time-step of $\Delta t = 0.1$\,s. In addition, a Gaussian smoothing with a window of 0.15\,s has been applied to the trajectories, in order to reduce the noise in the computation of the velocities and accelerations of the fish, using centered difference. Finally, we augmented the data by partly exploiting the rotational symmetry of the system: $x\to \pm x$, $y \to \pm x$, $x\leftrightarrow y$.

\subsection*{Permutations among fish}
As noted earlier, there is a subtlety in matching the identity of the $N$ Hilbert agents with the identity of the $N$ real fish in the training data set. A first option, which would only be reasonable for very small values of $N$, consists in augmenting the data by including the $N!$ permutations of the identities of the $N$ real fish (hence resulting in $n\to N!\times n$ data). However, for $N=5$, the number of identity permutations is already $N! = 120$, and this procedure would be too time-consuming, as the sums in Eqs.~(\ref{WeightFish},\ref{HilbertFish}) are running over the number of data $n$. Hence, for $N=5$, we define the distance  $\|x-x_i\|$ between the input vector $x$ for the Hilbert agents and a given data vector for the real fish $x_i$, as the minimal distance over the $N!$ permutations of the identities of the $N$ real fish. For larger values of $N$, the optimal permutation of the $N$ real fish best matching the configuration of the Hilbert agents can be obtained by using the Hungarian algorithm that solves this assignment problem in polynomial time in $N$~\cite{Hungarian1955}. Alternatively, for large group size $N$, the prediction of the acceleration of a focal individual could only require an input vector involving its  position and velocity and that of $N'\ll N$ nearest neighbors. For instance, in two dimensions, like for fish swimming in shallow water like considered here, the first Voronoi layer has, on average, 6 neighbors, and taking $N'=6$ should be reasonable for large group size.

\subsection*{Treatment of the arena boundary}

In some instances, the updated position of a Hilbert agent would fall outside the circular arena of radius $R=25$\,cm. In that case, we introduce a rejection procedure to enforce the presence of the tank wall, which is also necessary for mathematical models~\cite{calovi2018disentangling,lei2020computational} and machine learning models~\cite{papaspyros2023ML,papaspyros2023biogap}. In the case of an invalid move at time $t+\Delta t$, we reject the updated position and velocity, and update the heading (but not the module) of the velocity at time $t$ so that it becomes almost parallel to the wall (we add a small angle of $\pi/20=9^\circ$). Then, the predicted acceleration, and ultimately, the velocity and position of the faulty agent are recomputed at time $t+\Delta t$. 

We have evaluated the stability of the Hilbert scheme by assessing whether it can learn the presence of the wall even without applying this rejection procedure (see Extended Data Fig.~\ref{extfig:2fish_nowall}).

\subsection*{Hilbert transfer function predictor and control}

The prediction process takes as input a series $u_t$, and generates an output series $x_{t}$. To initiate the process, initial conditions ($x^{pred}_1,x^{pred}_2,..,x^{pred}_d$) and ($u_1,u_2,..,u_{d-1}$) have to be specified, following which the values $u_t$ are iteratively appended to the input series, and the output at the next time step $x^{pred}_{t+1}$ computed using Eq.~(\ref{eq:ARMAmodel}). The initial prediction period ($x^{pred}_1,x^{pred}_2,..,x^{pred}_d$) may be chosen randomly or set to be equal to a previously observed period from the training data. 

During offline prediction, at each time step $t$, the current input value $u_t$ is appended to the stored input time series $u_{t-1},u_{t-2},..$ and the next output $x^{pred}_{t+1}$ is predicted using Eq.~(\ref{eq:ARMAmodel}) using a fixed training period of inputs and outputs:

\begin{equation}
 x^{pred}_{t+1} = f_H(x^{pred}_t,x^{pred}_{t-1},..,x^{pred}_{t-d+1}, u_{t},u_{t-1},..,u_{t-d+1}).
\end{equation}
    
 In the online prediction process, each time a prediction $x^{pred}_{t+1}$ is made, the corresponding true value $x_{t+1}$ is appended to the training data set, together with the corresponding control input $u_t$. 

If acceleration is being predicted using Eq.~(\ref{eq:ARMAmodel}), followed by a double integration to obtain $x_{t+1}$, the following equations is used to make the prediction (note that finite differences of the training data are used to obtain training values for the predictor variable $a_t$): 

\begin{eqnarray}
    a^{pred}_{t+1}&=&f_H(x^{pred}_t,x^{pred}_{t-1},..,x^{pred}_{t-d+1},u_t,u_{t-1},..,u_{t-d+1}),\\
    v^{pred}_{t+1}&=& v^{pred}_t+a^{pred}_{t+1},\\
    x^{pred}_{t+1}&=&x^{pred}_t+v^{pred}_{t+1}.
\end{eqnarray}

In Fig.~\ref{figcontrol1a} the two-state linear dynamical system used with output $x_t$ and input $u_t$ is defined by the equations (in the plots shown, $-k_1=-k_2=k_3=0.1$): 

\begin{eqnarray}
    a_{t+1}&=&k_1 x_t + k_2 v_t + k_3 u_t,\\
    v_{t+1}&=& v_t+a_{t+1},\\
    x_{t+1}&=&x_t+v_{t+1}.
\end{eqnarray}

To generate training data, $u_t$ was chosen to be i.i.d with a standard normal distribution. 

In Fig.~\ref{figcontrol1b} the torque-driven, damped pendulum is defined by the discrete-time equations ($\gamma = 0.5$, $dt = 0.2$): 

\begin{eqnarray}
    v_{t+1}&=& v_t-(\gamma v_t+sin(\theta_t))dt+u_t,\\
    x_{t+1}&=&x_t+v_{t+1} dt.
\end{eqnarray}

To generate training data, $u_t$ was chosen to be i.i.d with a standard normal distribution. During the control phase, a quadratic cost function in the output $C = \frac{1}{T}\sum_{t=1}^T (x_t-x_{target})^2$ was used. 

For the directed stochastic gradient procedure used to optimize the control cost in the forecast period of a finite time horizon $T$ in MPC, we used the following update procedure \cite{nesterov2017random}: 

\begin{equation}
U_{i+1} = U_i - g \frac{C(U_i + \epsilon dU_i)-C(U_i)}{\epsilon} dU_i.
\end{equation}

Here $U_i(t) = (u^i_t,u^i_{t+1},..,u^i_{t+T-1})$ denotes a trial control vector of length $T$, which is iteratively updated for a number of Monte Carlo iterations $i=1..M$ (we chose $M=100$). $C(X(U),U)$ is the $U$-dependent cost function, which is computed by first predicting the output vector $x^{pred}_{t+1}(U),..,x^{pred}_{t+T}(U)$, either using the Hilbert procedure, or the true system model as a baseline. Note, that in this case the offline Hilbert procedure has to be used, since these are Monte Carlo iterations and the true value of the output is not known.

For the first iteration for the first time step, $U_1$ is initialized randomly. For subsequent time steps, The first $T-1$ components of the vector $U_1(t+1)$ are initialized using last $T-1$ components of $U_M(t)$. The stochastic steps $dU_i$ are chosen to be random directions on the $T$-dimensional unit sphere. The gain $g$ was chosen to be 10 and the step size $\epsilon = 0.3$ in the control applications shown in Fig.~\ref{fig:ControlsFig2}. 

After $M$ stochastic directional gradient steps, and following the MPC procedure, we take {\it only} the first time step $u_t$ from the vector $U_M(t)$ and apply this as the control input. We then observe the real system output $x_{t+1}$, and append both $u_t$ and $x_{t+1}$ to the training data before iterating the MPC loop. 

\subsection*{Hilbert and Transformer predictors for the zeros of the Riemann function}

A list of the first 2\,001\,052 zeros of the Riemann zeta function, accurate to within $4\times 10^{-9}$ has been retrieved from~\cite{odlyzko_zeros}, but we have not exploited the first million zero in order to lie deep in the “high-energy” universal regime. For both predictors, the training set consists of 100\,000 ratios, and the test set is a disjoint set of 200\,000 consecutive ratios.
The PDF of the ratios, $P(r)$, has a fat power-law tail for large $r$ (see~\cite{Atas2013ratio}), resulting in the  values of the observed ratios $r$ spanning several orders of magnitude. Hence, we have exploited the fact that $P(r) = \frac{1}{r^2}P(\frac{1}{r})$, so that the PDF of $\log(r)$ is an even function, which decays exponentially for large negative or positive $\log(r)$. The variable $\log(r)$ is then more adapted as an input and output variable for the Hilbert scheme. Similarly, the Transformer predictor is fed with sequences of the log-ratio $\log(r)$ instead of sequences of the ratio $r$.
The code and data to regenerate 
Fig.~\ref{fig:Riemann1} and Fig.~\ref{fig:Riemann2} are provided (see Data availability). In particular, the parameters of the Transformer are detailed in the \texttt{Log-Transformer.py code}. In addition, it is possible to interact with Fig.~\ref{fig:Riemann2} to change the viewing direction by readily executing the \texttt{Treat-Riemann\_zeros.py} code.

\backmatter

\section*{Supplementary Information}

\subsection*{\bf Supplementary Movies}
\noindent\textbf{Movie S1:}
\label{mov:S1} 2-minute collective dynamics of 2 Hilbert agents for a memory $M=2$, and enforcing the presence of the wall by means of the rejection procedure.

\vskip 0.4cm\noindent\textbf{Movie S2:}
\label{mov:S2} 2-minute collective dynamics of 2 Hilbert agents for a memory $M=2$, without enforcing the presence of the wall. Note the occurrence of small and short excursions of the fish outside the limits of the tank.

\vskip 0.4cm\noindent\textbf{Movie S3:}
\label{mov:S3} 1-minute collective dynamics of 5 Hilbert agents for a memory $M=2$, and enforcing the presence of the wall by means of the rejection procedure.

\subsection*{\bf Extended Data}
The Extended Data document contains 1 table and 7 figures mentioned in the main text.

\subsection*{\bf Proof Appendix}
The Proof Appendix includes the detailed proof of the 10 theorems mentioned in the main text.

\section*{Acknowledgments} C.\,S. is especially grateful to Guy Theraulaz (CRCA, Toulouse) and Vaios Papaspyros (EPFL, Lausanne) for an inspiring collaboration and fruitful discussions on collective phenomena in animal groups. Moreover, the authors are grateful to Vaios Papaspyros for invaluable discussions of machine learning methods in this context.

\section*{Declarations}

\begin{itemize}
\item \textbf{Funding:} P.M. gratefully acknowledges support from the Crick-Clay Professorship.
\item \textbf{Competing interests:} The authors declare no conflict of interest/competing interests.
\item \textbf{Ethics approval:} Not applicable.
\item \textbf{Consent to participate:} Not applicable.
\item \textbf{Consent for publication:} Not applicable.
\item \textbf{Data availability:} The data from~\cite{papaspyros2023biogap,papaspyros2023ML} for $N=2$ and $N=5$ fish have been downsampled at 10\,FPS and smoothed using a Gaussian kernel of width $0.15$\,s. The resulting fish trajectory time series are available at \url{https://figshare.com/projects/Data_and_Code_for_the_article_AI_without_networks/188349}.
\item \textbf{Code availability:} The Fortran codes to generate Hilbert fish trajectories for $N=2$ and $N=5$ agents are available at \url{https://figshare.com/projects/Data_and_Code_for_the_article_AI_without_networks/188349}.
\item \textbf{Authors' contributions}: P.\,M. and C.\,S. designed the research;  P.\,M. and C.\,S. obtained the main theorems from heuristic arguments; C.\,S. wrote the proofs of the theorems with inputs from P.\,M.; C.\,S. applied the Hilbert scheme to fish data and developed the application to Zeta function zeros; P.\,M. developed the Hilbert predictive control application;  P.\,M. and C.\,S. wrote the article.
\end{itemize}

\newpage
\bibliography{sn-bibliography}
\newpage
\begin{center}
    {\LARGE\bf Supplementary Information for\\
    \textit{AI without networks},\\  \vskip 0.2cm   by P.\,P. Mitra and C.~Sire}
\end{center}
\vskip 2cm
\section*{Legends of Supplementary Movies}

\noindent\textbf{Movie S1:}
2-minute collective dynamics of 2 Hilbert agents for a memory $M=2$, and enforcing the presence of the wall by means of the rejection procedure.

\vskip 0.4cm\noindent\textbf{Movie S2:}
2-minute collective dynamics of 2 Hilbert agents for a memory $M=2$, without enforcing the presence of the wall. Note the occurrence of small and short excursions of the fish outside the limits of the tank.

\vskip 0.4cm\noindent\textbf{Movie S3:}
1-minute collective dynamics of 5 Hilbert agents for a memory $M=2$, and enforcing the presence of the wall by means of the rejection procedure.

\vskip 1.5cm
\section*{Extended Data Table}

\setcounter{figure}{0}  
\renewcommand{\figurename}{Extended Data Table}

\begin{figure}[ht]
\centering
        \includegraphics[width=0.9\textwidth]{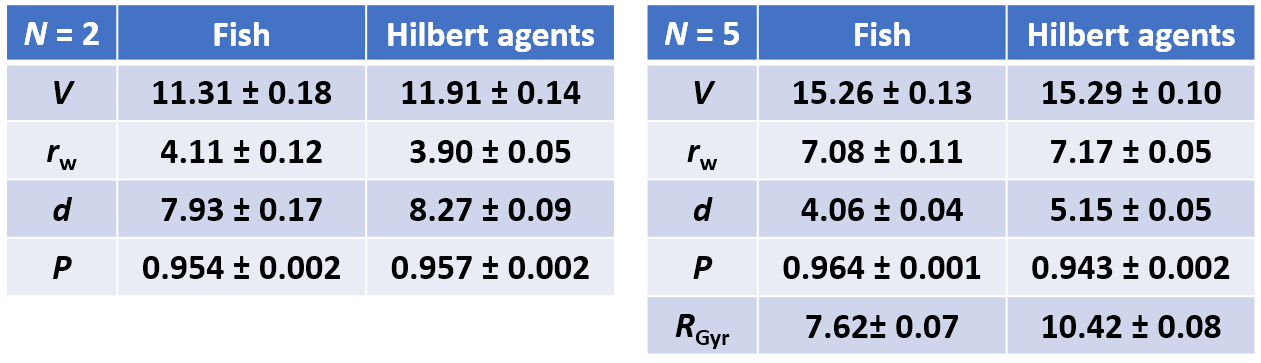}
        \caption{\small We report the mean and standard error for the PDF of the observables appearing in Fig.~3 (for $N=2$ individuals) and Fig.~4 (for $N=5$ individuals). The speed $V$ is expressed in cm/s, while the distance to the wall, $r_w$, the distance between two nearest neighbors, $d$, and the gyration radius, $R_{\rm Gyr}$ (only for $N=5$), are expressed in cm. Finally, the polarization, $P$, is without unit and between 0 and 1.}
         \label{Table1}   
\end{figure}

\newpage

\section*{Extended Data Figures}

\setcounter{figure}{0}   
\renewcommand{\figurename}{Extended Data Fig.}

\begin{figure}[ht]
     \centering
        \includegraphics[width=\textwidth]{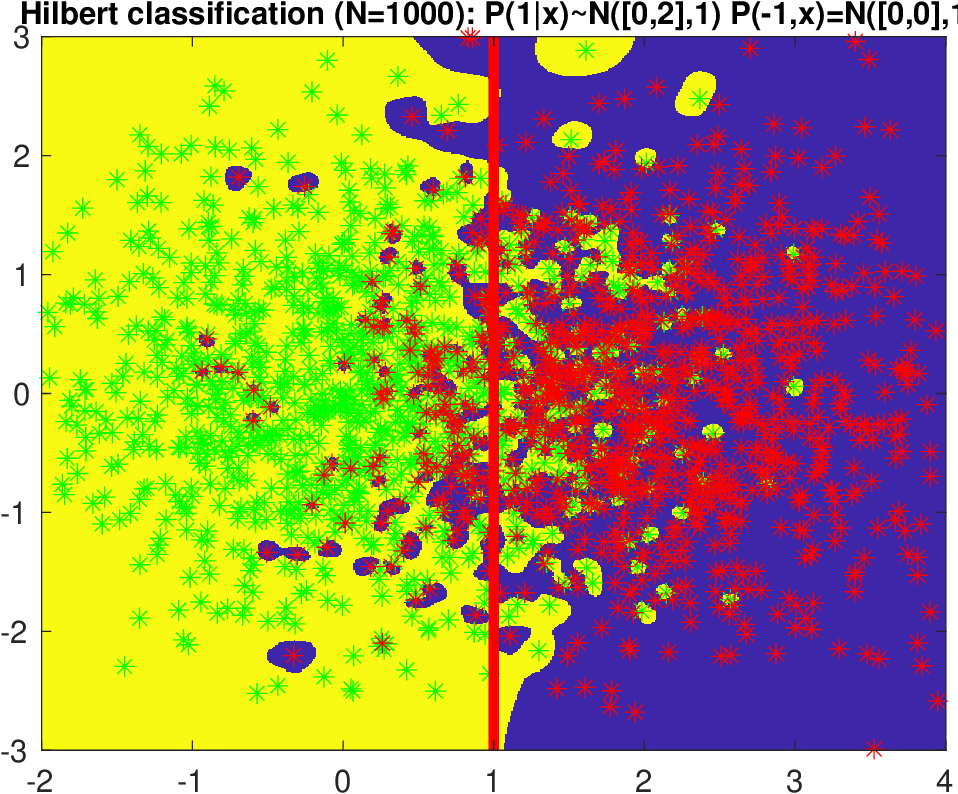}
         \caption{\small Classification using the Hilbert kernel: A simple example is shown, with two classes of points drawn from a mixture of 2D unit Normal distributions, with mean separated by 2. The points are shown in green and red colors (1000 points of each class). The red vertical line is the Bayes classification boundary. The yellow and blue colored regions are the Hilbert-predicted classification regions for the green and red points. The islands of blue in yellow (and vice versa) are due to the interpolative nature of the classifier, and correspond to the phenomenon of adversarial examples which are guaranteed for interpolating classifiers on noisy data.}
         \label{fighilbertclassification}

\end{figure}

\begin{figure}[ht]
     \centering
        \includegraphics[width=\textwidth]{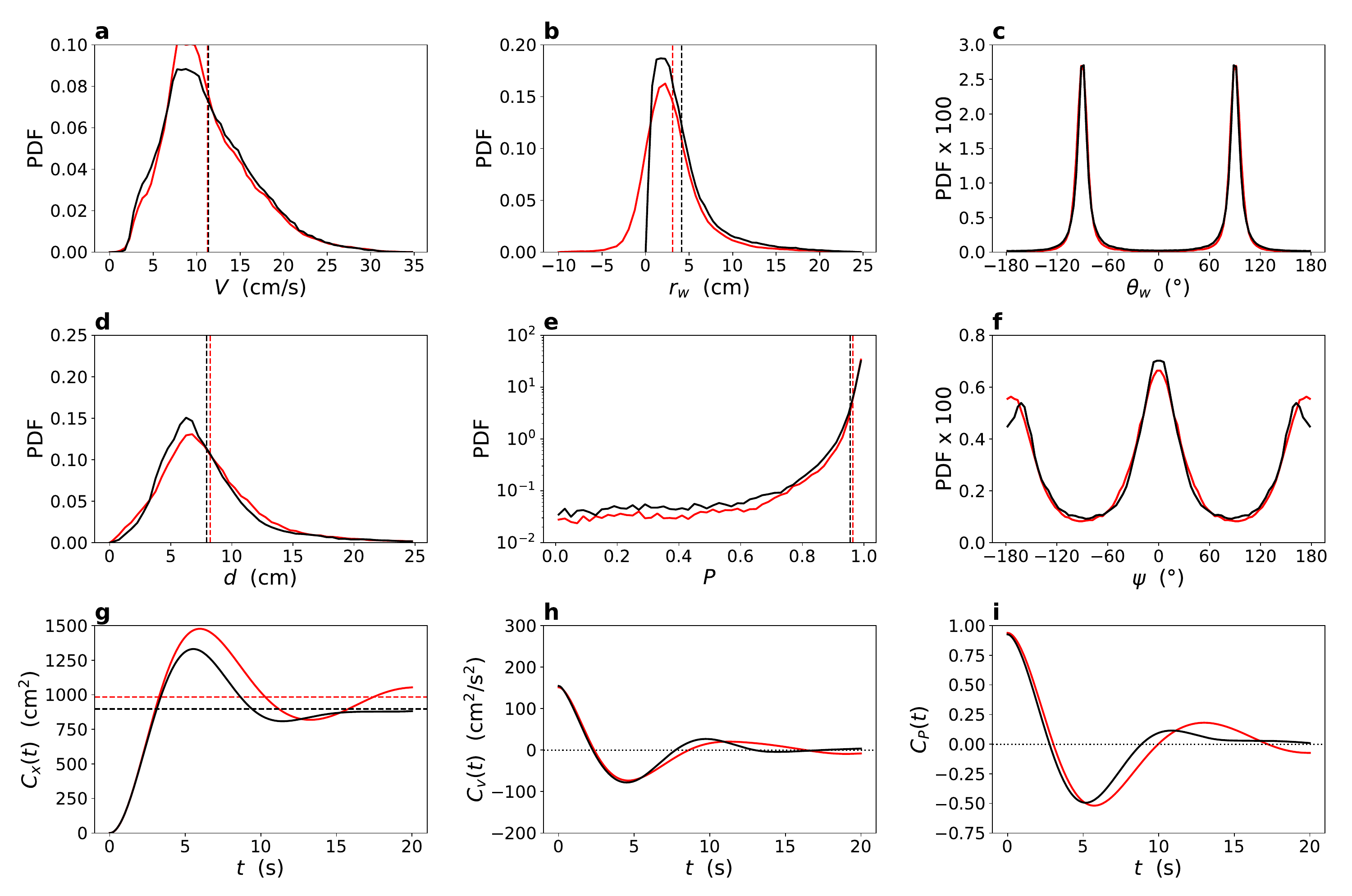}
         \caption{\small \textbf{Behavior of 2 Hilbert fish without the tank wall.} This figure is the analog of Fig.~3 in the main text (also for a memory $M=2$), but in the case where the presence of the tank wall is not enforced in the Hilbert model. The different panels show the 9 observables used to characterize the individual ({\bf a}-{\bf c}) and collective ({\bf d}-{\bf f}) behavior, and the time correlations in the system ({\bf g}-{\bf i}): {\bf a},~PDF of the speed, $V$; {\bf b}, PDF of the distance to the wall, $r_{\rm w}$; {\bf c}, PDF of the heading angle relative to the normal to the wall, $\theta_{\rm w}$; {\bf d}, PDF of the distance between the pair of individuals, $d$; {\bf e}, PDF of the group polarization, $P=\left|\cos({\Delta\phi}/{2})\right|$, where $\Delta\phi $ is the relative heading angle; {\bf f},~PDF of the viewing angle at which an individual perceives the other individual,~$\psi$. See Fig.~\ref{fig:flowchart}a and b in the main text for a visual representation of the main variables. {\bf g},~Mean squared displacement, $C_x(t)$, and its asymptotic limit, $C_x(\infty) =2\langle r^2\rangle$ (dotted lines); {\bf h}, Velocity autocorrelation,  $C_v(t)$; {\bf i}, Polarization autocorrelation, $C_P(t)$. The black PDFs correspond to experiments, while the red PDFs correspond to the predictions of the Hilbert generative model. The plots are on the same scale as in Fig.~3 in the main text, except for $r_{\rm w}$, for which the horizontal axis has been extended to negative values of $r_{\rm w}$ corresponding to instances where an individual is observed outside the limits of the experimental circular tank. Yet, the Hilbert fish spend 87\,\% of the time strictly within the tank limits, and when they wander outside the tank, their average excursion distance from the wall is only 1.3\,cm. These excursions are responsible for the upward and rightward shift of the peak of $C_x(t)$ and for the larger asymptotic limit, $C_x(\infty) =2\langle r^2\rangle\approx 980$\,cm$^2$  (compared to $C_x(\infty)\approx 900$\,cm$^2$ for fish or for the Hilbert model implementing the rejection procedure enforcing the presence of the tank wall).} 
         \label{extfig:2fish_nowall}
\end{figure}

\begin{figure}[ht]
     \centering
        \includegraphics[width=\textwidth]{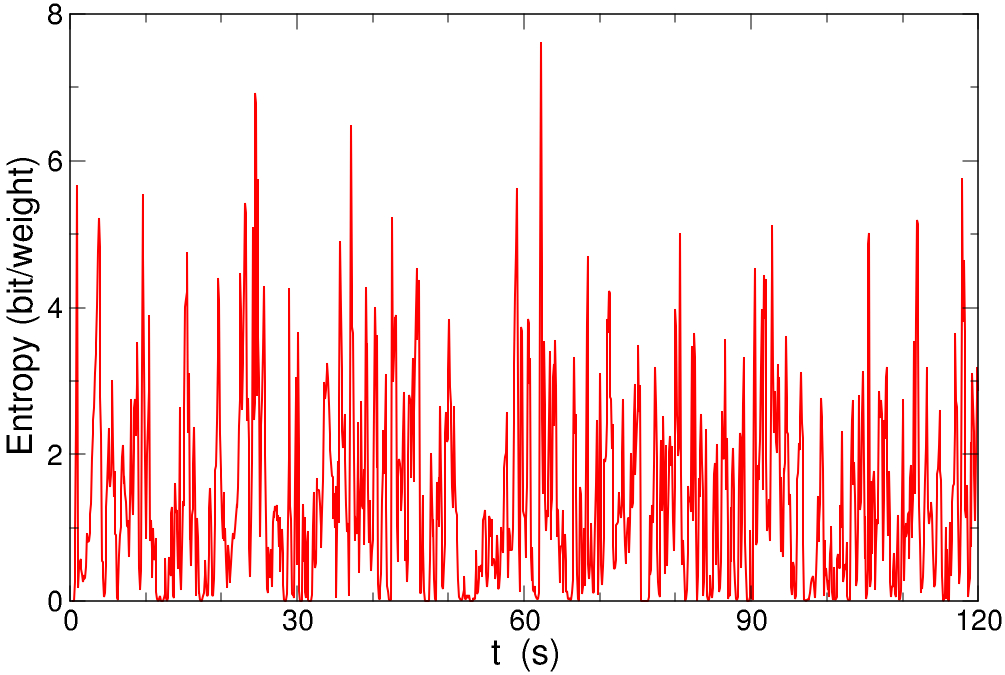}
         \caption{\small \textbf{Entropy time series for 2 Hilbert fish.} We plot a 2-minute time series of the entropy for 2 Hilbert fish, and for a memory $M=2$. The entropy $S$ can be interpreted as $\log_2 {\cal N}$, where ${\cal N}$ is the effective number of real fish configurations used to predict the acceleration of the Hilbert fish. The time series exhibits  short periods  where ${\cal N}\approx 1$ ($S\approx 0$), when the Hilbert scheme has essentially selected a unique real fish configuration (“copying”). This short time series also presents three short periods when ${\cal N}>64$ ($S>6$). The PDF of the entropy computed over much longer time is shown in Extended Data Fig.~4.}
         \label{extfig:2fish_entropy_time}
\end{figure}

\begin{figure}[ht]
     \centering
        \includegraphics[width=\textwidth]{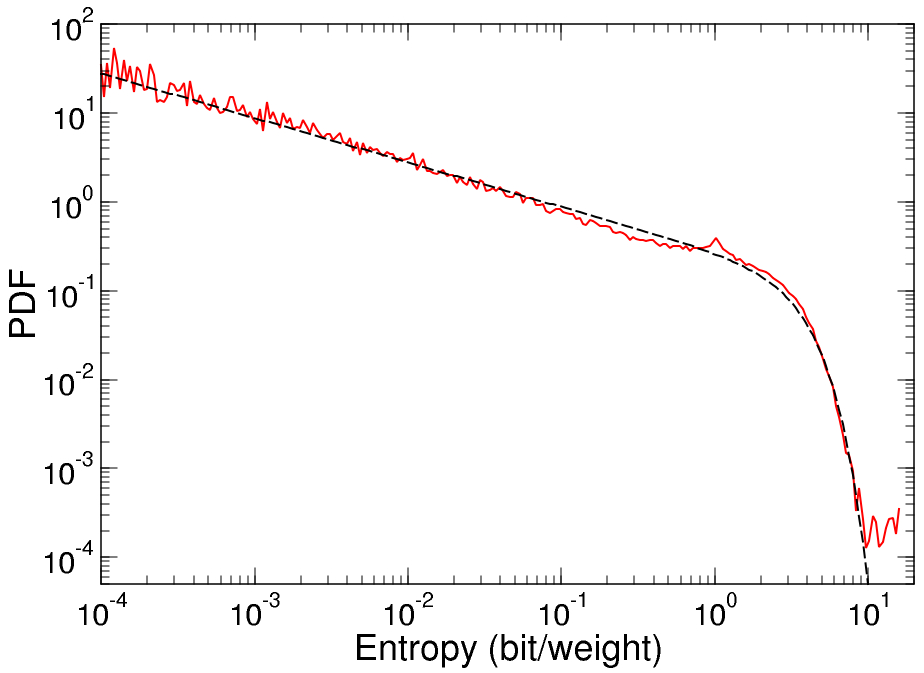}
         \caption{\small \textbf{Entropy distribution for 2 Hilbert fish.} We plot the PDF of the entropy for 2 Hilbert fish (for a memory $M=2$) resulting from an effective simulation time of 3 hours. The PDF of the entropy is reasonably well fitted by the normalized functional form $\rho(S)= (2\,\Gamma[5/4])^{-1} \,(S_c\, S)^{-1/2}\,\exp\left(-(S/S_c)^2\right)$, although the fit does not capture some outliers at $S>10$. The fitted cut-off entropy scale, $S_c\approx 3.61$, corresponds to ${\cal N}_c= 2^{S_c}\approx 12.2$ real fish configurations contributing to the acceleration prediction, while the mean entropy $\langle S\rangle\approx 1.37$ corresponds to $2^{\langle \log_2{\cal N}\rangle} =2^{\langle S\rangle}\approx 2.6$ configurations. The mean number of configurations used for a prediction is $\langle {\cal N}\rangle = \langle 2^S\rangle$, and is dominated by outliers. If this average is restrained to instance where $S\leq 10$, one finds $\langle {\cal N}\rangle_{S\leq 10} \approx 5.2$ (our fit $\rho(S)$ would predict $\langle {\cal N}\rangle_{S\leq 10} \approx 4.7$), whereas the average including all data is $\langle {\cal N}\rangle \approx 35$. Also note the small peak in the PDF near $S=1$, corresponding to ${\cal N}=2$ relevant configurations contributing almost equally to the Hilbert prediction. Yet, during the simulation, entropies as high as $S\sim 15$ were recorded, corresponding to ${\cal N}\sim 32768$ fish configurations effectively considered by the Hilbert kernel. Compared to $k$NN methods, the Hilbert interpolation scheme is hence able to adapt the effective number of used data for the prediction to the properties of the input vector. 
         See also Extended Data Fig.~3 for a short time series of the entropy.}  
         \label{extfig:2fish_entropy_pdf}
\end{figure}

\begin{figure}[ht]
     \centering
        \includegraphics[width=\textwidth]{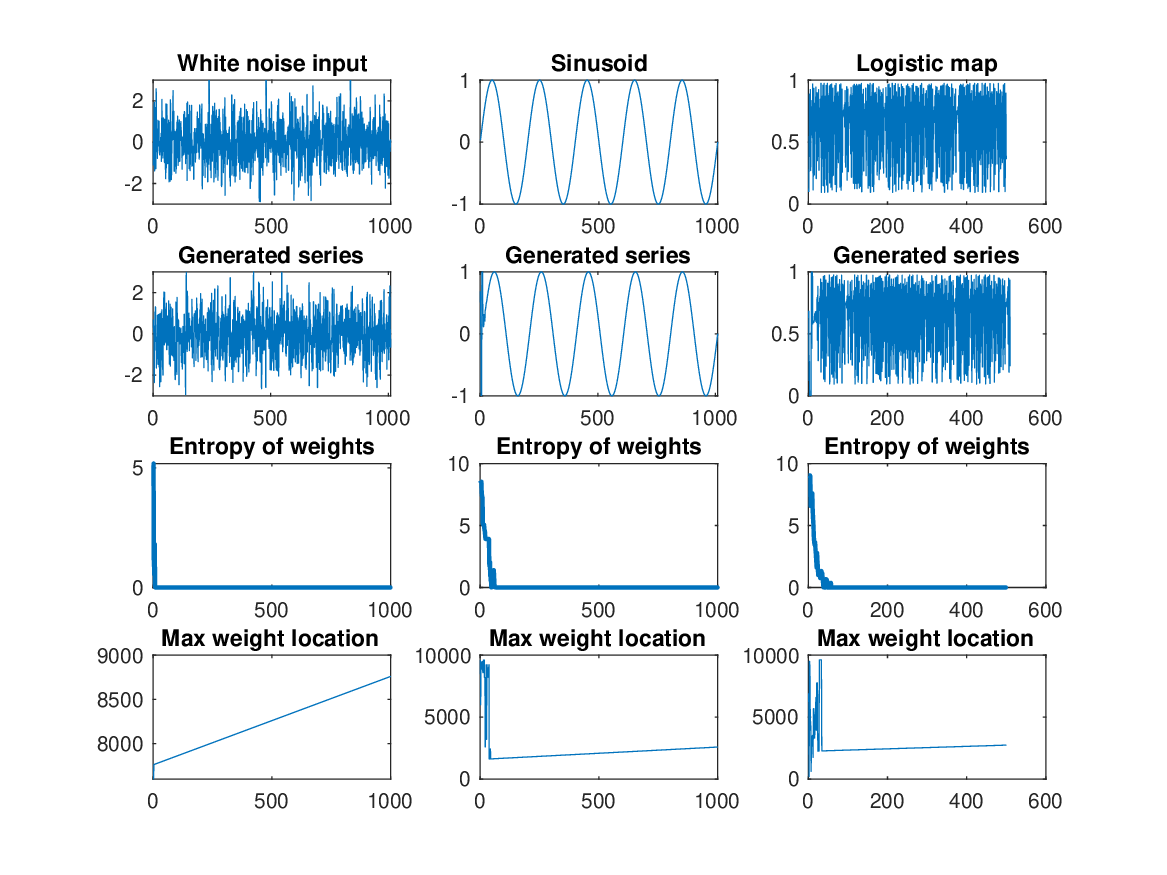}
         \caption{\small Autoregressive generative modeling of time series data: three examples are shown of signals generated by Eq.~(\ref{eq:ARmodel}) with simple training data consisting of $N=10\,000$ samples of a single training signal. The three columns respectively show results corresponding to a training signal generated by white Gaussian noise, a sinusoid, and iterates of a logistic regression equation $x_{n+1}=\lambda x_n (1-x_n)$ for $\lambda=3.9$. A lag window size of $T=10$ is used (see the next figure for a lag window size of $T=40$, and the signal generation is initiated by random initial conditions consisting of $T$ samples of a standard normal distributed variable. The second row shows the generated signal. The third row shows the entropy of the generative weights as a function of generation time, and the fourth row shows the position of the maximum weight in the training signal, also as a function of generation time. Note the “copying” behavior, where the generated signal starts some fragment of the training signal after an initial transient. During the “copying” phase, the weight entropy falls to zero, and the index of the maximum weight increments linearly with time.}
         \label{extfig:autoregression1}

\end{figure}

\begin{figure}[ht]
     \centering
        \includegraphics[width=\textwidth]{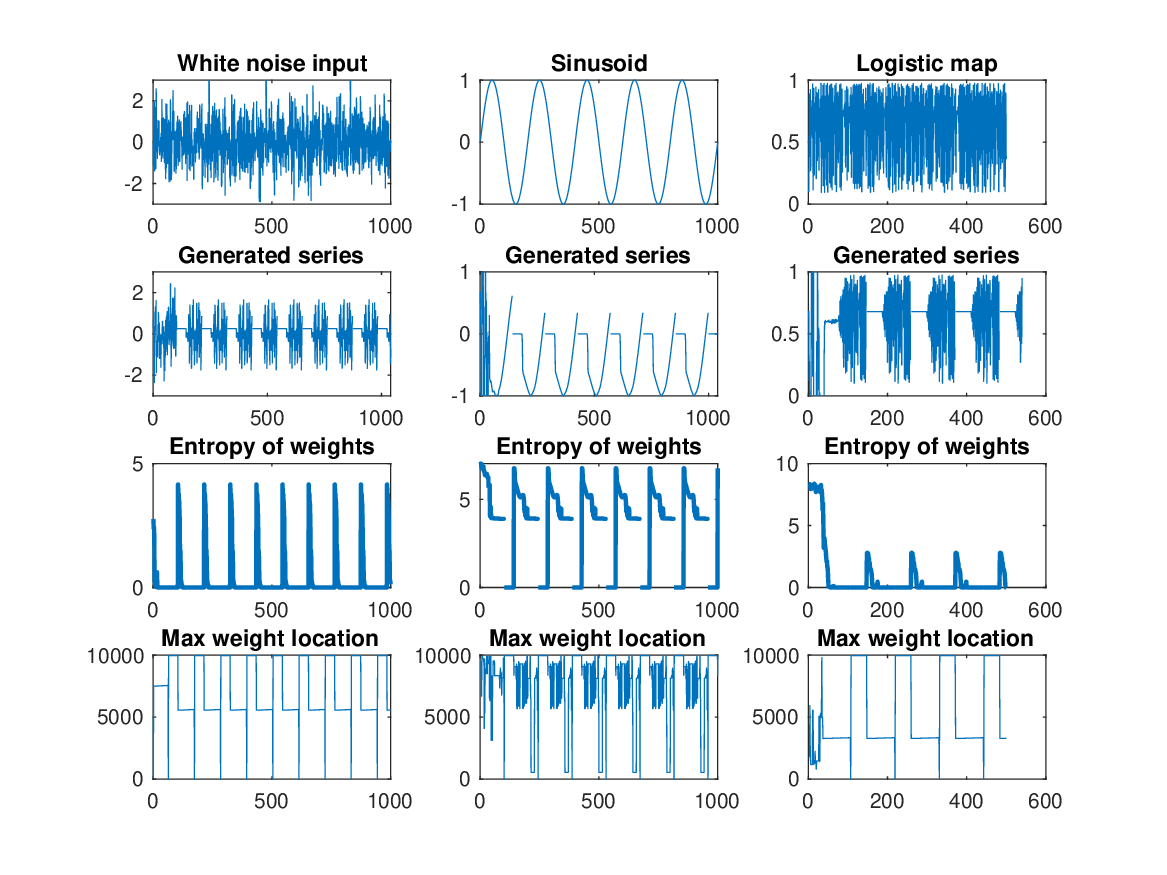}
         \caption{\small Autoregressive generative modeling of time series data: the same signal examples as in the previous extended data figure (Extended Data Fig.~5)~ are shown, this time with a lag window $T=40$. In this case, periodic behavior is observed after an initial transient. The periods themselves contain short episodes of “copying” where the entropy falls to zero, and also show short episodes of close to fixed-point behavior of the dynamics, where the generated signal has an almost constant value. The extent of the “copying”, periodic or constant behaviors depends on the initial conditions as well as on $d$.} 
         \label{extfig:autoregression2}

\end{figure}

\begin{figure}[ht]
     \centering
        \includegraphics[width=\textwidth]{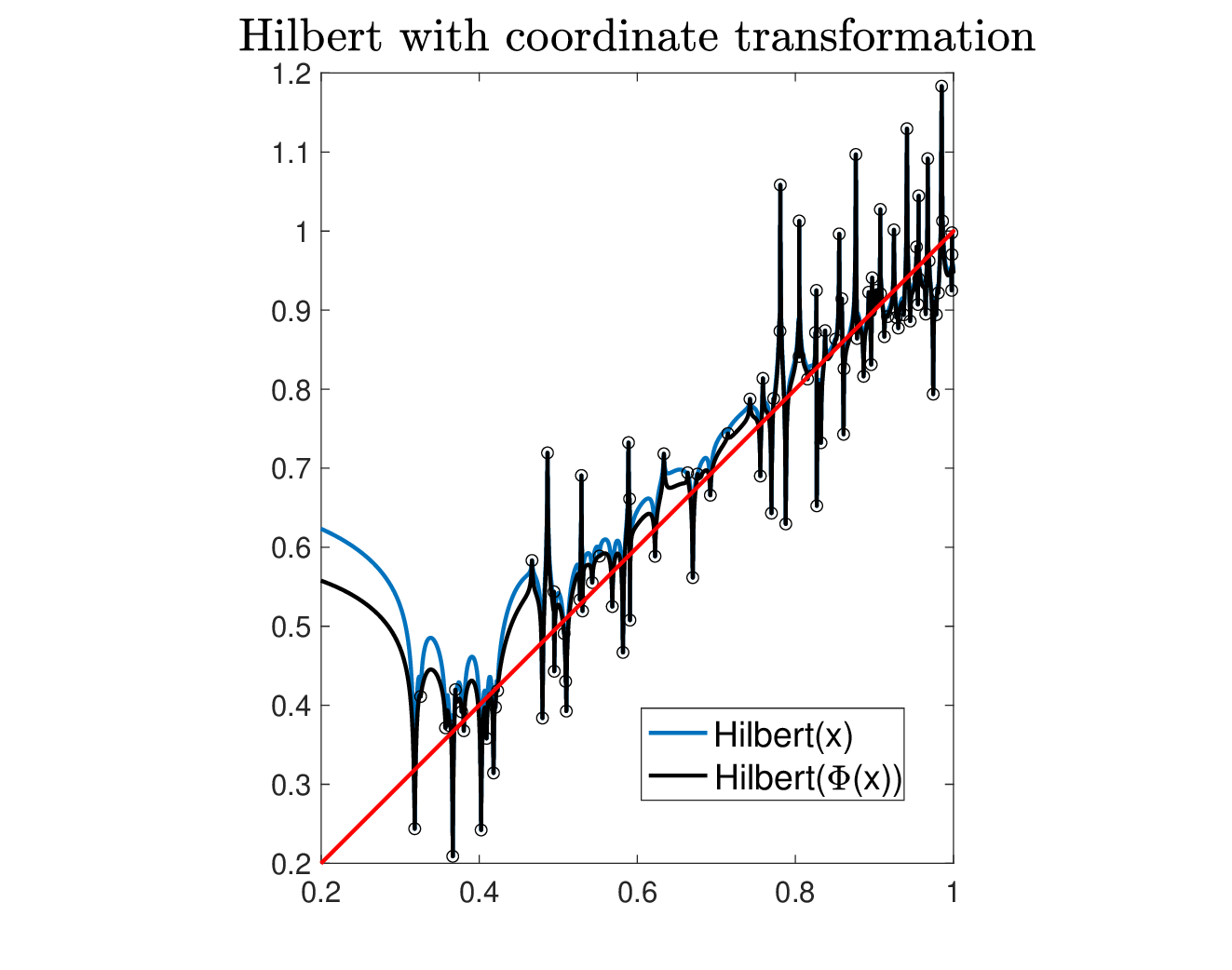}
         \caption{\small Impact of an initial coordinate transformation $x\rightarrow \Phi(x)$. This example shows two Hilbert kernel fits, with the black curve corresponding to the original kernel, and the blue curve corresponding to the generalized kernel corresponding to replacing $x\rightarrow \Phi(x)$ in the weights. The samples $x_i$ ($i=1..100$) are chosen so that the transformed coordinates $\Phi(x_i)$ have a uniform distribution. The red line corresponds to $y=x$ and uncorrelated Gaussian noise with $\sigma = 0.1$ is added to produce the noisy samples. The two regression functions both interpolate, but show slight differences, especially in the data-sparse region, with the transformed weights (that produce uniform sampling of $x$) being a bit closer to the noise-free function. As proven in the paper, both estimates are statistically consistent and have the same large-sample asymptotics in the leading order, but the sub-leading order behavior will generally depend on $\Phi$ in conjunction with the other details of the problem.} 
         \label{extfig:coordinatetransform}

\end{figure}

\begin{appendices}

\onecolumn
\section{Proofs of the theorems}
\label{sec:proofs}
\vskip 0.5cm

\subsection{Preliminaries}

In the following, $x\in \Omega^\circ$ so that $\rho(x)>0$, and we will assume for simplicity that the distribution $\rho$ is continuous at $x$. 

For the proof of our results, we will often exploit the following integral relation, valid for $\beta>0$ and $z>0$,
\begin{equation}\label{trick}
	\frac{1}{\Gamma(\beta)}\int_0^{+\infty}t^{\beta-1}{\rm e}^{-t\,z} \,dt=z^{-\beta}.
\end{equation}

In addition, we define 
\begin{equation}\label{psi}
	\psi(x,t)\coloneqq\int \rho(x+y){\rm e}^{-\frac{t}{||y||^d}}\,d^dy,
\end{equation}
which will play a central role. We note that $\psi(x,0)=1$, and that $t\mapsto\psi(x,t)$ is a continuous and strictly decreasing function of $t$. It is even infinitely differentiable at any $t>0$, but not necessarily at $t=0$. In fact, for a fixed $x$, controlling the behavior of $1-\psi(x,t)$ when $t\to 0$ will be essential to obtain our results.

\subsection{Moments of the weights: large $n$ behavior}  \label{prooftheo1}
In this section, we provide a complete proof of Theorem~\ref{theo1}. Several other theorems will use the same method of proof, and some basic steps will not be repeated in their proof.

Using \eq{trick} for $\beta>0$, we can express powers of the weight function as
\begin{equation}\label{mom1}
	w_0^\beta(x)=\frac{1}{||x-x_0||^{\beta d}}  \frac{1}{\Gamma(\beta)}  \int_0^{+\infty}t^{\beta-1}{\rm e}^{-t\,||x-x_0||^{-d}-t\,\sum_{i=1}^n||x-x_i||^{-d}}\,dt.
\end{equation}
By taking the expected value over the $n+1$ independent random variables  $X_i$, we obtain
\begin{equation}\label{mom2}
\E\left[w_0^\beta(x)\right]=\frac{1}{\Gamma(\beta)}  \int_0^{+\infty}t^{\beta-1}\psi^n(x,t)\phi_\beta(x,t)\,dt,
\end{equation}
with 
\begin{equation}\label{phibeta}
	\phi_\beta(x,t)\coloneqq\int \rho(x+y)\frac{{\rm e}^{-\frac{t}{||y||^d}}}{||y||^{\beta d}}\,d^dy,
\end{equation}
which is also a strictly decreasing function of $t$, continuous at any $t>0$ (in fact, infinitely differentiable for $t>0$). 

Note that the exchange of the integral over $t$ and over $\vec x=(x_0,x_1,..., x_n)$ used to obtain \eq{mom2} is justified by the Fubini theorem, by first noting that the function  $\vec x\mapsto w_0^\beta(x)\prod_{i=0}^n\rho(x_i) $ is in $L^1(\R^d)$, since $0\leq w_0^\beta(x) \leq 1$, and since $\rho$ is obviously in $L^1(\R^d)$. Moreover, the function $t\mapsto t^{\beta-1}\psi^n(x,t)\phi_\beta(x,t)>0$ is also in $L^1(\R)$. Indeed, we will show below that it decays fast enough  when $t\to +\infty$ (see Eqs.~(\ref{fubi3}-\ref{I2fin})), ensuring the convergence of its integral at $+\infty$, and that it is bounded (and continuous) near $t=0$ (see Eqs.~(\ref{fubi1}-\ref{fubi2})), ensuring that this function is integrable at $t=0$.

For $\beta=1$, $\phi_1=-\partial_t \psi$, and we obtain $\E\left[w_0 (x)\right]=\frac{1}{n+1}$, as expected. In the following, we first focus on the case $\beta>1$, before addressing the cases $0<\beta<1$ and $\beta<0$ at the very end of this section.

We now introduce $t_1$ and $t_2$ (to be further constrained later) such that $0<t_1<t_2$. We then express the integral of \eq{mom2} as the sum of corresponding integrals $I_1+I_{12}+I_2$. $I_1$ is the integral between 0 and $t_1$, $I_{12}$ the integral between $t_1$ and $t_2$, and $I_2$ the integral between $t_2$ and $+\infty$. Thus, we have
\begin{equation}\label{boundno1}
 I_1\leq	\E\left[w_0^\beta(x)\right]\leq I_1+I_{12}+I_2,
\end{equation}
provided these integral exists, which we will show below, by providing upper bounds for $I_2$ and $I_{12}$, and tight lower and upper bound for the leading term $I_1$.

\vskip 2cm

\noindent{\textit{Bound for} $I_2$}

For any $R\geq 1$, we can write  the integral defining  $\psi(x,t)$ 
\begin{eqnarray}\label{fubi3}
	\psi(x,t)&=&\int_{||y||\leq R}+\int_{||y||\geq R}\\
	&\leq & {\rm e}^{-\frac{t}{R^d}}+\int_{||y||\geq R}\rho(x+y)\frac{||y||^2}{R^2}\,d^dy,\\
	&\leq & {\rm e}^{-\frac{t}{R^d}} +\frac{C_x}{R^2},
\end{eqnarray}
with $C_x=\sigma^2_\rho+||x-\mu_\rho||^2$ depending on the mean $\mu_\rho$ and variance $\sigma^2_\rho$ of the distribution $\rho$. Similarly, for $\phi_\beta(x,t)$, we obtain the bound 
\begin{equation}\label{fubiphi}
    \phi_\beta(x,t) \leq  \frac{1}{R^{\beta d}}{\rm e}^{-\frac{t}{R^d}}+\frac{C_x}{R^{2+\beta d}},
\end{equation}
valid for $t\geq\max (1,\beta)$ and   $R\leq r_t$, where  $r_t= (t/\beta)^{1/d}\geq 1$ is the location of the maximum of the function  $r\mapsto\frac{{\rm e}^{-\frac{t}{r^d}}}{r^{\beta d}}$.

We now set $R=t^{\frac{s}{d}}$, with $0<s<1$,   and take $T'_2\geq \max (1,\beta, \beta^{1/(1-s)})$ (so that $1\leq R\leq r_t$) is large enough such that the following conditions are satisfied for $t\geq t_2\geq T'_2$,
\begin{eqnarray}
{\rm e}^{-\frac{t}{R^d}}={\rm e}^{-t^{1-s}}&\leq & \frac{C_x}{t^{\frac{2s}{d}}},\\
 \frac{1}{R^{\beta d}}{\rm e}^{-\frac{t}{R^d}}=\frac{1}{t^{\beta s}} {\rm e}^{-t^{1-s}}   &\leq & \frac{C_x}{t^{\frac{2s}{d}+\beta s}}.
\end{eqnarray}
Hence, for $t\geq t_2\geq T'_2$, we obtain
\begin{eqnarray}
	\psi(x,t)&\leq & \frac{2C_x}{t^{\frac{2s}{d}}},\label{boundpsi12}\\
\phi_\beta(x,t) &\leq &  \frac{2C_x}{t^{\frac{2s}{d}+\beta s}}.\label{boundphi12}
\end{eqnarray}
In addition, we also impose $t_2\geq T''_2=(4C_x)^{d/(2s)}$, so that  $\frac{2C_x}{t^{\frac{2s}{d}}}\leq \frac{1}{2}$, for any $t\geq T_2=\max(T'_2,T''_2)$.
We can now exploit the resulting bounds for $\psi(x,t)$ and $\phi_\beta(x,t)$ in \eq{boundpsi12} and \eq{boundphi12} 
to compute an explicit bound for $I_2$, for any given $t_2\geq T_2$:
\begin{equation}\label{I2fin0}
	I_2=\frac{1}{\Gamma(\beta)} \int_{t_2}^{+\infty}t^{\beta-1}\psi^n(x,t)\phi_\beta(x,t)\,dt
	\leq  \frac{1}{\Gamma(\beta)} \int_{t_2}^{+\infty}t^{\beta(1-s)-1} \left(\frac{2C_x}{t^{\frac{2s}{d}}}\right)^{n+1}\,dt.
\end{equation}
The integral in the right-hand side of \eq{I2fin0} only converges for $s>\frac{1}{1+\frac{2(n+1)}{\beta d}}$ (remember that we also impose $s<1)$, and we then set $s=\frac{1}{1+\frac{2}{\beta d}}$, which ensures its convergence for any $n\geq 1$. Performing this integral and using the fact that $\frac{2C_x}{t_2^{\frac{2s}{d}}}\leq \frac{1}{2}$, we finally obtain
\begin{equation}
	I_2\leq C_x\frac{d+\frac{2}{\beta}}{\Gamma(\beta)}\times
	\frac{1}{n\,2^{n}}.\label{I2fin}
\end{equation}
We hence obtain the convergence of $I_2$, which, along with the bounds for $I_1$ and $I_{12}$ below, justifies our use of Fubini theorem to obtain \eq{mom2}. Note that the above bound essentially decays exponentially with $n$, under the stated conditions.

\vskip 0.2cm
\noindent{\textit{Bound for} $I_{12}$}

Again, exploiting the fact that $\psi(x,t)$ and $\phi_\beta(x,t)$ are strictly decreasing functions of $t$, we obtain
\begin{equation}\label{I12}
	I_{12}\leq\frac{\phi_\beta(x,t_1)t_2^{\beta}}{\Gamma(\beta)}\times\psi^n(x,t_1),
\end{equation}
where we note that $\psi(x,t_1)<1$, for any $t_1>0$, implying that this bound decays exponentially with $n$.

\vskip 0.2cm
\noindent{\textit{Bound for} $I_{1}$}

We first want to obtain bounds for $1-\psi(x,t)$, where $0\leq t\leq t_1$, with $t_1>0$ to be constrained below. In addition, exploiting the continuity of $\rho$ at $x$ and the fact that $\rho(x)>0$, we introduce $\varepsilon$ satisfying $0<\varepsilon <1/4$, and define $\lambda>0$ small enough so that the ball $B(x,\delta)\subset\Omega^\circ$, and  $||y||\leq \lambda\implies |\rho(x+y)-\rho(x)|\leq \varepsilon\rho(x)$. Exploiting this definition, we obtain the following lower and upper bounds
\begin{eqnarray}
	1-\psi(x,t) &\geq& (1-\varepsilon)\rho(x)\int_{||y||\leq \lambda}\left( 1-{\rm e}^{-\frac{t}{||y||^d}}\right) \,d^dy, \label{psib1}\\
	1-\psi(x,t)  &\leq& (1+\varepsilon)\rho(x)\int_{||y||\leq \lambda}\left( 1-{\rm e}^{-\frac{t}{||y||^d}}\right) \,d^dy\\ &&\,\,+\int_{||y||\geq \lambda}\rho(x+y)\left( 1-{\rm e}^{-\frac{t}{\lambda^d}}\right)\,d^dy,\\
	&\leq& (1+\varepsilon)\rho(x)\int_{||y||\leq \lambda}\left( 1-{\rm e}^{-\frac{t}{||y||^d}}\right) \,d^dy + \frac{t}{\lambda^d}.\label{psib3}
\end{eqnarray}
The integral appearing in these bounds can be simplified by using radial coordinates:
\begin{eqnarray}
\int_{||y||\leq \lambda}\left( 1-{\rm e}^{-\frac{t}{||y||^d}}\right) \,d^dy,
 	 &=&S_d \int_0^\lambda \left( 1-{\rm e}^{-\frac{t}{r^d}}\right) r^{d-1}\,dr,\\
 	&=&V_d t \int_{\frac{t}{\lambda^d}}^{+\infty}\frac{1-{\rm e}^{-u}}{u^2}\,du,
\end{eqnarray} 	
where $S_d$ and $V_d=\frac{S_d }{d}$ are respectively the surface and the volume of the $d$-dimensional unit sphere, and we have used the change of variable $u=\frac{t}{r^d}$.

We note that for $0<z\leq 1$, we have
\begin{equation}\label{intlog}
	 \int_{z}^{+\infty}\frac{1-{\rm e}^{-u}}{u^2}\,du=-\ln(z)+\int_z^1\frac{1-u-{\rm e}^{-u}}{u^2}\,du+\int_{1}^{+\infty}\frac{1-{\rm e}^{-u}}{u^2}\,du.
\end{equation}
Exploiting this result and now imposing $t_1\leq \lambda^d$, we have, for any $t\leq t_1$ 
\begin{eqnarray} 	
	\ln \left( \frac{C_-}{t}\right) &\leq&\int_{\frac{t}{\lambda^d}}^{+\infty}\frac{1-{\rm e}^{-u}}{u^2}\,du ~~\leq~~	\ln \left( \frac{C_+}{t}\right),  \\
\ln (C_+)&=&d\ln(\lambda) +\int_{1}^{+\infty}\frac{1-{\rm e}^{-u}}{u^2}\,du,\\
\ln (C_-)&=&\ln (C_+)+\int_0^1\frac{1-u-{\rm e}^{-u}}{u^2}\,du.
\end{eqnarray} 	
Combining these bounds with \eq{psib1} and \eq{psib3}, we have shown the existence of two $x$-dependent constants $D_\pm$ such that, for $0\leq t\leq t_1\leq \lambda^d$, we have
\begin{equation}\label{boundpsi1}
(1-\varepsilon)V_d \rho(x)\,t\ln\left( \frac{D_-}{t}\right)	\leq
1-\psi(x,t) \leq (1+\varepsilon)V_d \rho(x)\,t\ln\left( \frac{D_+}{t}\right).
\end{equation}
In addition, we will also choose $t_1< D_\pm/3$, such that  the two functions $t\ln\left( \frac{D_\pm}{t}\right)$	are positive and strictly increasing for $0\leq t\leq t_1$. $t_1$ is also taken small enough such that the two bounds in \eq{boundpsi1} are always less than 1/2, for $0\leq t\leq t_1$ (both bounds vanish when $t\to 0$). 

We now obtain efficient bounds for $\phi_\beta(x,t)$, for $0\leq t\leq t_1$. Proceeding similarly as above, we obtain 
\begin{eqnarray}\label{fubi1}
	\phi_\beta(x,t)&\geq &(1-\varepsilon)\rho(x)  \int_{||y||\leq \lambda}\frac{{\rm e}^{-\frac{t}{||y||^d}}}{||y||^{\beta d}} \,d^dy,\label{phib1}\\
	\phi_\beta(x,t)&\leq&
	(1+\varepsilon)\rho(x)\int_{||y||\leq \lambda}\frac{{\rm e}^{-\frac{t}{||y||^d}}}{||y||^{\beta d}}\,d^dy 
	+\frac{1}{\lambda^{\beta d}}\label{phib2}.	
\end{eqnarray}	
Again, the integral appearing in these bounds can be rewritten as
\begin{equation}	
	\int_{||y||\leq \lambda}\frac{{\rm e}^{-\frac{t}{||y||^d}}}{||y||^{\beta d}} \,d^dy \label{intphi1}
        = S_d \int_0^\lambda    r^{d(1-\beta)-1}{\rm e}^{-\frac{t}{r^d}} \,dr.
\end{equation} 
For $0<\beta<1$, the integral of \eq{intphi1} is finite for $t=0$, ensuring the existence of $\phi_\beta(x,0)$ and the  fact that $t\mapsto t^{\beta-1}\psi(x,t)\phi_\beta(x,t)$ belongs to $L^1(\R)$ (hence, justifying our use of Fubini theorem for $0<\beta<1$).
For $\beta>1$, we have
\begin{eqnarray}	
	\int_{||y||\leq \lambda}\frac{{\rm e}^{-\frac{t}{||y||^d}}}{||y||^{\beta d}} \,d^dy
       &=& V_d \,t^{1-\beta} \int_{\frac{t}{\lambda^d}}^{+\infty} u^{\beta-2}{\rm e}^{-u}  \,du.\\
       &\underset{t\to 0}{\sim}& V_d \Gamma(\beta-1)t^{1-\beta}.\label{intphi2}
\end{eqnarray} 
This integral diverges when $t\to 0$ and the constant term $\lambda^{-\beta d}$ in \eq{phib2} can be made as small as necessary  (by a factor less than $\varepsilon$) compared to this leading integral term, for a small enough $t_1$. Similarly, we can choose $t_1$ small enough so that the integral \eq{intphi1} is approached by the asymptotic result of \eq{intphi2} up to a factor $\varepsilon$.
Thus, we find  that for $0\leq t\leq t_1$, one has
\begin{equation}\label{fubi2}
 (1-2\varepsilon) V_d\rho(x)\Gamma(\beta-1)t^{1-\beta}	\leq	\phi_\beta(x,t)\leq (1+3\varepsilon) V_d\rho(x)\Gamma(\beta-1)t^{1-\beta}.
\end{equation}
This shows that $t^{\beta-1}\phi_\beta(x,t)$ has a smooth limit equal to $V_d\rho(x)\Gamma(\beta-1)$, when $t\to 0$, so that, combined with the finite upper bound for $I_2$, $t\mapsto t^{\beta-1}\psi(x,t)\phi_\beta(x,t)$ belongs to $L^1(\R)$, for $\beta>1$, and hence for all $\beta>0$. Hence, the use of the Fubini theorem to derive \eq{mom2} has been  justified.

Now combining the bounds for $\psi(x,t)$ and $\phi_\beta(x,t)$, we obtain
\begin{eqnarray}
	I_1 &\geq &(1-2\varepsilon)\frac{1}{\beta-1} V_d\rho(x)\int_0^{t_1}  \left(  1-(1+\varepsilon)V_d \rho(x)\,t\ln\left( \frac{D_+}{t}\right)              \right)^n\,dt,\\
   	I_1 &\leq &(1+3\varepsilon)\frac{1}{\beta-1} V_d\rho(x)\int_0^{t_1}  \left(  1-(1-\varepsilon)V_d \rho(x)\,t\ln\left( \frac{D_-}{t}\right)              \right)^n \,dt.
\end{eqnarray}    	

\vskip 0.2cm
\noindent{\textit{Asymptotic behavior of} $I_{1}$ \textit{and} $\E\left[w_0^\beta(x)\right]$}

We will show below that 
\begin{equation}\label{eqmain}
	\int_0^{t_1}  \left(  1-E_\pm t\ln\left( \frac{D_\pm}{t}\right)\right)^n \,dt\underset{n\to +\infty}{\sim}\frac{1}{E_\pm n\ln(n)},
\end{equation}
where $E_\pm=(1\mp\varepsilon)V_d \rho(x)$. For a given $x$, and for $t_1$ and $t_2$ satisfying the requirements mentioned above, the upper bounds for $I_{12}$ (see \eq{I12}) and $ I_2$ (see \eq{I2fin}) appearing in \eq{boundno1} both decay exponentially with $n$ and can hence be made arbitrarily small compared to $I_1$ which decays as $1/(n\,\ln(n))$.

Finally, assuming for now the result of \eq{eqmain} (to be proven below), we have obtained the exact asymptotic result
\begin{equation}\label{momfin}
\E\left[w_0^\beta(x)\right]\underset{n\to +\infty  }{\sim}\frac{1}{(\beta-1) n\ln(n)}.
\end{equation}

\vskip 0.2cm
\noindent{\textit{Proof of \eq{eqmain}}}
	
We are then left to prove the result of \eq{eqmain}. First, we will use the fact that, for $0\leq z\leq z_1<1$, one has
\begin{equation}\label{expbound}
	{\rm e}^{-\mu z}\leq 1-z \leq {\rm e}^{- z}, 
\end{equation}
where $\mu=-\ln(1-z_1)/z_1$. We can apply this result to the integral of  \eq{eqmain}, using $z_1^\pm=E_\pm t_1\ln(D_\pm/t_1)>0$. Note that $0<t_1<D_\pm/3$ and hence $z_1^\pm>0$ can be made  as close to 0 as desired, and the corresponding $\mu_\pm>1$ can be made as close to 1 as desired. Thus, in order to prove \eq{eqmain}, we need to prove the following equivalent 
\begin{equation}\label{eqmain2}
I_n=	\int_0^{t_1}  {\rm e}^ {-n E t\ln\left( \frac{D}{t}\right) } \,dt\underset{n\to +\infty  }{\sim}\frac{1}{E n\ln(n)},
\end{equation}
for an integral of the form appearing in \eq{eqmain2}. Let us mention again that $t_1$ has been taken small enough so that the function $t\mapsto t\ln\left(\frac{D}{t}\right)$ is positive and strictly increasing (with its maximum at $t_{\max}=D/e<t_1$), for   $0\leq t\leq t_1$.

We now take $n$ large enough so that $\frac{\ln(n)}{n}<t_1$ and $E\ln(n)>1$. One can then write
\begin{eqnarray}
I_n&=&\frac{1}{n}\int_0^{\ln(n)}   {\rm e}^ {-E u\ln\left( \frac{Dn}{u}\right) } \,du+	\int_{\frac{\ln(n)}{n}} ^{t_1}  {\rm e}^ {-n E t\ln\left( \frac{D}{t}\right) }   \,dt=J_n+K_n,   \\
J_n&\leq &\frac{1}{n}\int_0^{1/E}   {\rm e}^ {-E u\ln\left(DEn\right) } \,du+\frac{1}{n}\int_{1/E}^{\ln(n)}   {\rm e}^ {-E u\ln\left( \frac{Dn}{\ln(n)}\right) }\,du, \\
&\leq & \frac{1}{E \,n\ln\left(D\,En\right)}+ \frac{\ln(n)}{D\,E \,n^2\ln\left( \frac{Dn}{\ln(n)}\right)},\\
K_n&\leq & 	\int_{\frac{\ln(n)}{n}} ^{+\infty}  {\rm e}^ {-n E t\ln\left( \frac{D}{t_1}\right) }  \,dt
\leq \frac{1}{E \,n^{1+E\ln\left( \frac{D}{t_1}\right) }\ln\left( \frac{D}{t_1}\right) }.
\end{eqnarray}	
When $n\to+\infty$, we hence find that the upper bound $I^+_n$ of $I_n$ satisfies
\begin{equation}\label{inplus}
	I_n^+\underset{n\to +\infty  }{\sim}\frac{1}{E \,n\ln\left(DEn\right)}\underset{n\to +\infty  }{\sim}\frac{1}{E \,n\ln\left(n\right)}.
\end{equation}	

Let us now prove a similar result for a lower bound of $I_n$ by considering $n$ large enough so that $n E t_1>1$, and by introducing $\delta$ satisfying $0\leq \delta<1/e$:
\begin{eqnarray}
	I_n&=&\frac{1}{nE}\int_0^{n E t_1}   {\rm e}^ {- u\ln\left( DEn\right) +u\ln(u)} \,du,\\
		&\geq& \frac{1}{nE}\int_0^{\delta}   {\rm e}^ {- u\ln\left(DEn \right) +\delta\ln(\delta)} \,du,\\
			&\geq& \frac{{\rm e}^{ \delta\ln(\delta)} }{nE\ln\left( DEn \right)}\left(1- \left( DEn\right)^{-\delta} \right) =	I_n^-(\delta).		
\end{eqnarray}				
Hence, for any $0\leq \delta<1/e$ which can be made arbitrarily small, and for $n$ large enough, we find that $I_n\geq  I_n^-(\delta)$, with    
\begin{equation}\label{inminus}
	 I_n^-(\delta) \sim \frac{{\rm e}^{ \delta\ln(\delta)}}{E \,n\ln\left(DEn\right)}\sim \frac{{\rm e}^{ \delta\ln(\delta)}}{E \,n\ln\left(n\right)}.
\end{equation}
\eq{inminus} combined with the corresponding result of \eq{inplus} for the upper bound $I_n^+$ finally proves \eq{eqmain2}, and ultimately,  \eq{momfin} and Theorem~\ref{theo1} for the asymptotic behavior of the moment $\E\left[w_0^\beta(x)\right]$, for $\beta>1$.

\vskip 0.2cm
\noindent\textit{Entropy} (moment for ``$\beta= 1^-$'')

We define the information entropy, $S(x)$, by 
\begin{equation}
    S(x)= -\sum_{i=0}^n w_i(x)\log[w_i(x)].
\end{equation}
If the weights are equidistributed over ${\cal N}$ data, one obtains $S=-{\cal N}\times 1/{\cal N}\log(1/{\cal N}) =\log({\cal N})$, and ${\rm e}^S={\cal N}$ indeed represents the number of contributing data. The expectation value of the entropy reads
\begin{equation}\label{entw0}
\E[S(x)]= -(n+1)\,\E[ w_0(x)\ln(w_0(x))].
\end{equation}

In order to evaluate \eq{entw0}, we use an integral representation in the spirit of \eq{trick}, valid for any $z>0$,
\begin{equation}\label{trick0}
	\int_0^{+\infty}(\ln(t)+\gamma)\,{\rm e}^{-t\,z} \,dt=-\frac{\ln(z)}{z},
\end{equation}
where $\gamma$ is Euler's constant.
Using \eq{trick0}, we find
\begin{eqnarray}\label{momlog}
	-w_0(x)\ln(w_0(x))=&& -\frac{1}{||x-x_0||^{-d}}   \int_0^{+\infty}{\rm e}^{-t\,||x-x_0||^{-d}-t\,\sum_{i=1}^n||x-x_i||^{-d}}\nonumber\\
   &&\times\left( \ln(||x-x_0||^{-d})+\ln(t)+\gamma  \right) \,dt.
\end{eqnarray}
By taking the expected value over the $n+1$ independent random variables  $X_i$, we obtain
\begin{equation}\label{momentlog}
\E\left[-w_0(x)\ln(w_0(x))\right]=  -\int_0^{+\infty}\psi^n(x,t)\left(\Phi_1(x,t)+(\ln(t)+\gamma)\phi_1(x,t)\right)\,dt,
\end{equation}
with 
\begin{equation}\label{philog}
	\Phi_1(x,t)\coloneqq\int \rho(x+y)\,{\rm e}^{-\frac{t}{||y||^d}}\frac{\ln \left( ||y||^{-d}\right)  }{||y||^{d}}\,d^dy,
\end{equation}
which is  continuous at any $t>0$ (in fact, infinitely differentiable for $t>0$). In addition, $\phi_1(x,t)=-\partial_t\psi(x,t)$ has been defined in \eq{phibeta}. 

By exploiting the same method used to bound $\phi_\beta(x,t)$ (see \eq{fubi2} and above it), we find that
\begin{eqnarray}
\Phi_1(x,t)\underset{t\to 0}{\sim} \frac{1}{2}V_d\rho(x)\ln^2(t),\label{boundbigphi1p}\\
\phi_1(x,t)\underset{t\to 0}{\sim} - V_d\rho(x)\ln(t),\label{boundphi1p}
\end{eqnarray}
where \eq{boundphi1p} is fully consistent with \eq{boundpsi1} (by naively differentiating \eq{boundpsi1}).

Finally, exploiting Eqs.~(\ref{boundbigphi1p},\ref{boundphi1p}), the integral of \eq{momentlog} can be evaluated with the same method as in the previous section, leading to 
\begin{eqnarray}\label{momentlogfinal}
\E\left[-w_0(x)\ln(w_0(x)\right]&\underset{n\to +\infty  }{\sim}& \frac{1}{2}V_d\rho(x)\int_0^{t_1} {\rm e}^ {-n V_d\rho(x) t\ln\left( \frac{D_\pm}{t}\right) }\ln^2(t)\,dt,\\
&\underset{n\to +\infty  }{\sim}&  \frac{1}{2}\frac{\ln(n)}{n}.
\end{eqnarray}
This last result proves the second part of Theorem~\ref{theo1} (see also the heuristic discussion below Theorem~\ref{theo1}) for the expected value of the entropy:
\begin{equation}\label{entw0final}
\E[S(x)]= -(n+1)\,\E[ w_0(x)\ln(w_0(x))]\underset{n\to +\infty  }{\sim}  \frac{1}{2}\ln(n).
\end{equation}

\vskip 0.2cm
\noindent\textit{Moments of order} $0<\beta<1$

The integral representation \eq{trick} allows us to also explore moments  of order $0<\beta<1$. In that case $\kappa_\beta(x)=\phi_\beta(x,0)<\infty$ is finite, with
\begin{equation}\label{phibeta0}
	\kappa_\beta(x)=\int \frac{\rho(x+y)}{||y||^{\beta d}}\,d^dy.
\end{equation}

By retracing the different steps of our proof in the case $\beta>1$, it is straightforward to show that
\begin{eqnarray}
\E\left[w_0^\beta(x)\right]&\underset{n\to +\infty  }{\sim}& \frac{\kappa_\beta(x)}{\Gamma(\beta)} 	\int_0^{t_1} t^{\beta-1}{\rm e}^ {-n V_d\rho(x) t\ln\left( \frac{D_\pm}{t}\right) }\,dt,\\
&\underset{n\to +\infty  }{\sim}&\frac{\kappa_\beta(x)}{(V_d\rho(x) n\ln(n))^\beta},
\end{eqnarray}
where the equivalent for the integral can be obtained by exploiting the very same method used in our proof of \eq{eqmain} above, hence proving the third part of Theorem~\ref{theo1}. 

We observe that contrary to the universal result of \eq{momfin} for $\beta$, the asymptotic equivalent for the moment of order $0<\beta<1$ is non-universal and explicitly depends on $x$ and the distribution $\rho$. 

\vskip 0.2cm
\noindent\textit{Moments of order} $\beta<0$

Finally, moments of order $\beta<0$ are unfortunately inaccessible to our methods relying on the integral relation \eq{trick}, which imposes $\beta>0$. However, we can obtain a few rigorous results for these moments (see also the heuristic discussion just after Theorem~\ref{theo1}).

Indeed, for $\beta=-1$, we have
\begin{equation}
\frac{1}{w_0(x)}=1+\|x-x_0   \|^d\sum_{i=1}^n \frac{1}{\|x-x_i   \|^d}.
\end{equation}
But since we have assumed that $\rho(x)>0$, $\E[\|x-x_i   \|^{-d}]=\int \frac{\rho(x+y)}{||y||^{d}}\,d^dy$ is infinite and moments of order $\beta\leq -1$ are definitely not defined.

As for the moment of order $-1<\beta<0$, it can be easily bounded, 
\begin{equation}
\E\left[w_0^\beta(x)\right]\leq 1+n\int {\rho(x+y)}{||y||^{|\beta|d}}\,d^dy\int \frac{\rho(x+y)}{||y||^{|\beta|d}}\,d^dy,
\end{equation}
and a sufficient condition for its existence is $\kappa_\beta(x)=\int {\rho(x+y)}{||y||^{|\beta|d}}\,d^dy<\infty$ (the other integral, equal to $\kappa_{|\beta|}(x)$, is always finite for $|\beta|<1$), which proves the last part of Theorem~\ref{theo1}.

\vskip 0.2cm
\noindent\textit{Numerical distribution of the weights} 

In the main text below Theorem~\ref{theo1}, we presented a heuristic argument showing that the results of Theorem~\ref{theo1} and  Theorem~\ref{theolag} (for the Lagrange function; that we prove below) were fully consistent with the weight $W=w_0(x)$ having a long-tailed scaling distribution,
\begin{equation}
P_n(W)=\frac{1}{W_n}p\left(\frac{W}{W_n} \right).
\end{equation}
The scaling function $p$ was shown to have a universal tail $p(w)\sim w^{-2}$ and the scale $W_n$ was shown to obey the equation
$-W_n\ln(W_n)=n^{-1}$. To the leading order for large $n$, we have $W_n\sim\frac{1}{n\ln(n)}$, and we can solve this equation recursively to find the next order approximation, $W_n\sim\frac{1}{n\ln(n\ln(n))}$. In Fig.~\ref{figdist} in the main text, we present numerical simulations for the scaling distribution $p$ of the variable $w=W/W_n$, for $n=65536$, using the estimate  $W_n\approx\frac{1}{n\ln(n\ln(n))}$. We observe that $p(w)$ is very well approximated by the function $\hat p(w)=\frac{1}{(1+w)^2}$, confirming our non-rigorous results.

The data were generated by drawing random values of $r_i^d=||x-x_i||^d$ using $(n+1)$ \textit{i.i.d.} random variables $a_i$ uniformly distributed in $[0,1[$, with the relation $r_i=[a_i/(1-a_i)]^{1/d}$, and by computing the  resulting weight $W=r_i^{-d}/\sum_{j=0}^n r_j^{-d}$. This corresponds to a distribution of $||x-x_i||$ given by $\rho(x-x_i)=1/V_d/(1+||x-x_i||^d)^2$.

\subsection{Lagrange function: scaling limit}\label{prooflag}

In this section, we prove Theorem~\ref{theolag} for the scaling limit of the Lagrange function $L_0(x)=\E_{X|x_0} [w_0(x)]$.
Exploiting again \eq{trick}, the expected Lagrange function  can be written as
\begin{equation}\label{lag}
L_0(x)=\|x-x_0\|^{-d} \int_0^{+\infty}\psi^n(x,t){\rm e}^{-t\|x-x_0\|^{-d}}\,dt,
\end{equation}
where $\psi(x,t)$ is again given by \eq{psi}.

For a given $t_1>0$, and remembering that $\psi(x,t)$ is a strictly decreasing function of $t$, with $\psi(x,0)=1$, we obtain
\begin{equation}
  L_1  \leq L_0(x)\leq L_1+ L_2,\label{l01}
\end{equation}
with 
\begin{eqnarray}
  L_1 &=&\|x-x_0\|^{-d} \int_0^{t_1}\psi^n(x,t){\rm e}^{-t\|x-x_0\|^{-d}}\,dt,\label{l1} \\
  L_2&=&{\rm e}^{-t_1\|x-x_0\|^{-d}}. \label{l2} 
\end{eqnarray}

For $\varepsilon>0$ and a sufficiently small $t_1>0$ (see section~\ref{prooftheo1}), we can use the bound for $\psi(x,t)$ obtained in section~\ref{prooftheo1}, to obtain
\begin{eqnarray}
	L_1 &\geq &(1-2\varepsilon)\frac{1}{\|x-x_0\|^{d}}\int_0^{t_1}  \left(  1-(1+\varepsilon)V_d \rho(x)\,t\ln\left( \frac{D_+}{t}\right)              \right)^n            {\rm e}^{-\frac{t}{\|x-x_0\|^{d}}}\,dt,\\
   	L_1 &\leq &(1+3\varepsilon)\frac{1}{\|x-x_0\|^{d}}\int_0^{t_1}  \left(  1-(1-\varepsilon)V_d \rho(x)\,t\ln\left( \frac{D_-}{t}\right)              \right)^n    
   	{\rm e}^{-\frac{t}{\|x-x_0\|^{d}}}\,dt.
\end{eqnarray}    
Then, proceeding exactly as in section~\ref{prooftheo1}, it is straightforward to show that $L_1$ can be bounded (up to factors $1+O(\varepsilon)$) by the two integrals $L_1^\pm$ 
\begin{equation}
	L_1^\pm=\frac{1}{\|x-x_0\|^{d}}\int_0^{t_1} {\rm e}^{-n\,V_d \rho(x)\,t\ln\left( \frac{D_\pm}{t}\right) -\frac{t}{\|x-x_0\|^{d}}}\,dt.
\end{equation}
Like in section ~\ref{prooftheo1}, we impose $t_1< D_\pm/3$, such that  the two functions $t\ln\left( \frac{D_\pm}{t}\right)$	are positive and strictly increasing for $0\leq t\leq t_1$.

We now introduce the scaling variable $z_x(n,x_0)=V_d\rho(x)\|x-x_0\|^d n\log(n)$, so that 
\begin{equation}\label{l1pmfin}
	L_1^\pm=\frac{1}{\|x-x_0\|^{d}}\int_0^{t_1} {\rm e}^{-\frac{t}{\|x-x_0\|^{d}}\left(1+z\frac{\ln\left( D_\pm/t\right)}{\ln(n)} \right)}\,dt=
	\int_0^{\frac{t_1}{\|x-x_0\|^{d}}} {\rm e}^{- u\left(1+z\frac{\ln\left( D_\pm\|x-x_0\|^{-d}/u\right)}{\ln(n)} \right)}\,du,
\end{equation}
where we have used the shorthand notation $z\equiv z_x(n,x_0)$.

For a given real $Z\geq 0$, we now want to study the (scaling) limit of $L_0(x)$ when $n\to\infty$, $\|x-x_0\|^{-d}\to +\infty$  (i.e., $x_0\to x$), and such that $z_x(n,x_0)\to Z$, which we will simply denote $\lim_Z L_0(x)$. We note that $\lim_Z L_2=0$ (see \eq{l01} and \eq{l2}), so that we are left to show that $\lim_Z L^\pm_1=\frac{1}{1+Z}=\lim_Z L_0(x)$, which will prove Theorem~\ref{theolag}.

Exploiting the fact that $u\ln(u)\geq -1/e$, for $u\geq 0$, we obtain
\begin{eqnarray}
L_1^\pm &\geq & {\rm e}^{-\frac{z}{e\ln(n)}} \int_0^{\frac{t_1}{\|x-x_0\|^{d}}} {\rm e}^{- u\left(1+z\frac{\ln\left( D_\pm\|x-x_0\|^{-d}\right)}{\ln(n)} \right)}\,du,\\
 &\geq &\frac{1}{1+z\frac{\ln\left( D_\pm\|x-x_0\|^{-d}\right)}{\ln(n)}}{\rm e}^{-\frac{z}{e\ln(n)}}\left(1-{\rm e}^{ -\frac{t_1}{\|x-x_0\|^{d}}  })\right).
\end{eqnarray}
Since we have $\lim_Z\frac{\ln\left( D_\pm\|x-x_0\|^{-d}\right)}{\ln(n)}=1$, $\lim_Z \frac{z}{e\ln(n)}=0$, and $\lim_Z\frac{t_1}{\|x-x_0\|^{d}}=+\infty$,
we find that $L_1^\pm$ is bounded from below by a term for which the $\lim_Z$ is $\frac{1}{1+Z}$, with a relative difference of order $1/\ln(n)$ for finite $n$.

Since we will ultimately take the $\lim_Z$ and hence the limit $x_0\to x$, we can impose that the upper limit of the last integral in \eq{l1pmfin} satisfies $\frac{t_1}{\|x-x_0\|^{d}}>1$.
Let us now consider $K>0$, such that $K<\frac{t_1}{\|x-x_0\|^{d}}$. We then obtain,
\begin{eqnarray}
\hspace{-0.5cm}L_1^\pm &\leq &\int_0^{K} {\rm e}^{- u\left(1+z\frac{\ln\left( D_\pm\|x-x_0\|^{-d}/K\right)}{\ln(n)} \right)}\,du+\int_K^{+\infty}{\rm e}^{- u}\,du,\\
&\leq &\frac{1}{1+z\frac{\ln\left( D_\pm\|x-x_0\|^{-d}/K\right)}{\ln(n)} }+{\rm e}^{-K}.
\label{l1upper} 
\end{eqnarray} 
We can now take $K$ such that $\ln(K)=\left[\ln\left(\frac{t_1}{\|x-x_0\|^{d}}\right)\right]^\alpha$, for some fixed $\alpha$ satisfying $0<\alpha<1$. It is clear that $K$ satisfies $K<\frac{t_1}{\|x-x_0\|^{d}}$. In addition, we have $\lim_Z K=+\infty$ and 
$\lim_Z \frac{\ln(K)}{\ln(n)}=0$, implying that the $\lim_Z$ of the upper bound in \eq{l1upper} is also $\frac{1}{1+Z}$, with a relative difference of order $1/[\ln(n)]^{1-\alpha}$ for finite $n$ (the closer $\alpha>0$ to 0, the more stringent this bound will be).

Finally, since $\lim_Z L_2=0$, 
we have shown that for any real $Z\geq 0$,  $\lim_Z L_1^\pm=\lim_Z L_0(x)=\frac{1}{1+Z}$, which proves Theorem~\ref{theolag}. Note that the two bounds obtained suggest that the relative error between $L_0(x)$ and $\frac{1}{1+Z}$ for finite large $n$ and large $\|x-x_0\|^{-d}$ with $z(n,x_0)$ remaining close to $Z$ is of order $1/\ln(n)$, or equivalently, of order $1/\ln(\|x-x_0\|)$.

\vskip 0.2cm
\noindent\textit{Numerical simulations for the Lagrange function at finite $n$}

In Fig.~\ref{fighilbertlag}, we illustrate numerically the scaling result of Theorem~\ref{theolag}.

Note that, exploiting Theorem~\ref{theolag}, we can use a simple heuristic argument to estimate the tail of the distribution of the random variable $W=w_0(x)$.
Indeed, approximating $L_0(x)$ for finite but large $n$ by its asymptotic form $\frac{1}{1+z_x(n,x_0)}$, with $z_x(n,x_0)=V_d\rho(x) n\log(n)\|x-x_0\|^d$, we obtain 
\begin{eqnarray}
\int_W^{1}P(W')\,dW' &\sim&\int \rho(x_0)\,\theta\left(\frac{1}{1+V_d\rho(x) n\log(n)\|x-x_0\|^d}-W\right)\,d^dx_0,\\
    &\sim& V_d\rho(x)\int_0^{+\infty} \theta\left(\frac{1}{1+V_d\rho(x) n\log(n)\, u}-W\right)\,du,\\
    &\sim& \frac{1}{n\ln(n)W}\quad\implies   P(W)\sim \frac{1}{n\ln(n)W^2},\label{argtail}
\end{eqnarray}
where $\theta(.)$ is the Heaviside function. 
This heuristic result is again perfectly consistent with our guess (see the discussion below Theorem~\ref{theo1}) that $P(W)=\frac{1}{W_n} p\left(\frac{W}{W_n} \right)$, with the scaling function $p$ having the universal tail, $p(w)\underset{w\to +\infty  }{\sim}w^{-2}$, and a scale $W_n\sim \frac{1}{n\ln(n)}$. Indeed, in this case and in the limit $n\to+\infty$, we obtain that $P(W)\sim\frac{1}{W_n}\left(\frac{W_n}{W}\right)^2\sim\frac{W_n}{W^2}\sim \frac{1}{n\ln(n)W^2}$, which is identical to the result of \eq{argtail}.

\subsection{The variance term}\label{proofvar}

We define the variance term ${\cal V}(x)$ as
\begin{equation}
   {\cal V}(x) = \E\Bigl[\sum_{i=0}^n w_i^2(x) [y_i-f(x_i)]^2\Bigr] = \E_X\Bigl[\sum_{i=0}^n w_i^2(x) \sigma^2(x_i) \Bigr]=(n+1)\,\E\left[w_0^2(x) \sigma^2(x_0)\right].
\end{equation}
If we first assume that $\sigma^2(x)$ is bounded by $\sigma_0^2$, we can readily bound ${\cal V}(x) $ using Theorem~\ref{theo1} with $\beta=2$:
\begin{equation}
   {\cal V}(x) \leq (n+1)\sigma_0^2\,\E\left[w_0^2(x)\right].
\end{equation}
Hence, for any $\varepsilon>0$, there exists a constant $N_{x,\varepsilon}$, such that for $n\geq N_{x,\varepsilon}$, we obtain Theorem~\ref{theovar1}
\begin{equation}
   {\cal V}(x) \leq (1+\varepsilon)\frac{\sigma_0^2}{\ln(n)}.
\end{equation}

However, one can obtain an exact asymptotic equivalent for ${\cal V}(x) $ by assuming that $\sigma^2$ is continuous at $x$ (with $\sigma^2(x)>0$), while relaxing the boundedness condition. Indeed, we now assume the growth condition $C_{\rm Growth}^\sigma$ 
\begin{equation}
    \int \rho(y)\frac{\sigma^2(y)}{1+\|y\|^{2d}}\,d^dy<\infty.\label{condsigma}
\end{equation}
Note that this condition can be satisfied even in the case where the mean variance $\int \rho(y)\sigma^2(y)\,d^dy$ is infinite.

Proceeding along the very same line as the proof of Theorem~\ref{theo1} in section~\ref{prooftheo1}, we can write
\begin{equation}\label{var}
\E\left[w_0^2(x)\sigma^2(x_0)\right]=  \int_0^{+\infty}t\psi^n(x,t)\phi(x,t)\,dt,
\end{equation}
with 
\begin{equation}\label{phivar}
	\phi(x,t)\coloneqq\int \rho(x+y)\sigma^2(x+y)\frac{{\rm e}^{-\frac{t}{||y||^d}}}{||y||^{2 d}}\,d^dy,
\end{equation}
which as a similar form as \eq{phibeta}, with $\beta=2$. The condition of \eq{condsigma} ensures that the integral defining $\phi(x,t)$ converges for all $t>0$.

The continuity of $\sigma^2$ at $x$ (and hence of $\rho\sigma^2$) and the fact the $\rho(x)\sigma^2(x)>0$ implies the existence of a small enough $\lambda>0$ such that the ball $B(x,\lambda)\subset\Omega^\circ$ and $||y||\leq \lambda\implies |\rho(x+y)\sigma^2(x+y)-\rho(x)\sigma^2(x)|\leq \varepsilon\rho(x)\sigma^2(x)$, a property exploited for $\rho$ in the proof of Theorem~\ref{theo1} (see \eq{psib1} and the paragraph above it), and which can now be used to efficiently bound $\phi(x,t)$. In addition, using the method of proof of Theorem~\ref{theo1} (see \eq{phib2}) also requires that $\int_{||y||\geq \lambda} \rho(y)\frac{\sigma^2(y)}{\|y\|^{2d}}\,d^dy<\infty$, which is ensured by the condition $C_{\rm Growth}^\sigma$ of \eq{condsigma}. Apart from these details, one can proceed strictly along the proof and Theorem~\ref{theo1}, leading to the proof of Theorem~\ref{theovar2}:
\begin{equation}\label{varfin}
   {\cal V}(x) \underset{n\to +\infty  }{\sim}\frac{\sigma^2(x)}{\ln(n)}.
\end{equation}
Note that if $\sigma^2(x)=0$, one can straightforwardly show that for any $\varepsilon>0$, and for $n$ large enough, one has
\begin{equation}
   {\cal V}(x) \leq\frac{\varepsilon}{\ln(n)},
\end{equation}
while a more optimal estimate can be easily obtained if one specifies how  $\sigma^2$ vanishes at $x$.

\subsection{The bias term}\label{proofbias}

This section aims at proving Theorem~\ref{theobias}, \ref{theobiasbis}, and \ref{theorhonon0}.

\textit{Assumptions}

We first impose the following growth condition $C_{\rm Growth}^f$ for $f(x) := \E[Y \mid X = x]$:
\begin{equation}
    \int \rho(y)\frac{f^2(y)}{(1+||y||^{d})^2}\,d^dy<\infty,
\end{equation}
which is obviously satisfied if $f$ is bounded. Since $\rho$ is assumed to have  a second moment, the condition $C_{\rm Growth}^f$ is also satisfied for any function satisfying $|f(x)|\leq A_f||y||^{d+1}$ for all $y$, such that $||y||\geq R_f$, for some $R_f>0$. Using the Cauchy-Schwartz inequality, we find that the condition $C_{\rm Growth}^f$ also implies that 
\begin{equation}
    \int \rho(y)\frac{|f(y)|}{1+||y||^{d}}\,d^dy<\infty.\label{growthf}
\end{equation}

In addition, for any $x\in\Omega^\circ$ (so that $\rho(x)>0$), we assume that there exists a neighborhood of $x$ such that $f$ satisfies a local Hölder condition. In other words, there exist $\delta_x>0$, $K_x>0$, and  $\alpha_x>0$, such that the ball $B(0,\delta_x)\subset\Omega$, and
\begin{equation}
    ||y||\leq \delta_x\implies |f(x+y)-f(x)|\leq K_x||y||^{\alpha_x},
\end{equation}
which defines condition $C_{\mathrm{Holder} }^f$.

\vskip 0.2cm
\textit{Definition of the bias term and preparatory results}

We define the bias term ${\cal B}(x)$ as
\begin{eqnarray}
{\cal B}(x)&=& \E_X\Bigl[\Bigl(\sum_{i=0}^n w_i(x) [f(x_i)-f(x)]\Bigr)^2\Bigr]=(n+1){\cal B}_1(x)+n(n+1){\cal B}_2(x),\\
{\cal B}_1(x)&=& \frac{1}{n+1}\E_X\Bigl[\sum_{i=0}^n w_i^2(x) [f(x_i)-f(x)]^2\Bigr],\\
&=&\E_X\Bigl[w_0^2(x) [f(x_0)-f(x)]^2\Bigr],\\
{\cal B}_2(x)&=&\frac{1}{n(n+1)}\E_X\Bigl[\sum_{0\leq i<j\leq n} w_i(x)w_j(x) [f(x_i)-f(x)][f(x_j)-f(x)]\Bigr],\\
&=& \E_X\Bigl[ w_0(x)w_1(x) [f(x_0)-f(x)][f(x_1)-f(x)]\Bigr].
\end{eqnarray}

Exploiting again \eq{trick} for $\beta=2$ like we did in section~\ref{prooftheo1}, we obtain
\begin{equation}
   {\cal B}_1(x)=\int_0^{+\infty}t\,\psi^n(x,t)\chi_1(x,t)\,dt,
\end{equation}
where $\psi(x,t)$ is again the function defined in \eq{psi}, and  where 
\begin{equation}\label{chi1}
	\chi_1(x,t)\coloneqq\int \rho(x+y){\rm e}^{-\frac{t}{||y||^d}}\frac{(f(x+y)-f(x))^2}{||y||^{2 d}}\,d^dy.
\end{equation}
For any $t>0$, and under condition $C_{\rm Growth}^f$, the integral defining $\chi_1(x,t)$ is well-defined. Moreover, $\chi_1(x,t)$ is a strictly positive and strictly decreasing function of $t>0$.

Now, defining $u_i=||x-x_i||^{-d}$, $i=0,...,n$ and exploiting again \eq{trick} for $\beta=2$, we can write 
\begin{equation}
    w_0(x)w_1(x)=u_0u_1\int_0^\infty t\,{\rm e}^{-(u_0+u_1)t-\left( \sum_{i=2}^n u_i\right)t}\,dt
\end{equation}
Now taking the expectation value over the $n+1$ independent variables, we obtain
\begin{eqnarray}
   {\cal B}_2(x)=\int_0^{+\infty}t\,\psi^{n-1}(x,t)\chi_2^2(x,t)\,dt,
\end{eqnarray}
where 
\begin{equation}\label{chi2}
	\chi_2(x,t)\coloneqq\int \rho(x+y){\rm e}^{-\frac{t}{||y||^d}}\frac{f(x+y)-f(x)}{||y||^{d}}\,d^dy.
\end{equation}
Again, for any $t>0$, and under condition $C_{\rm Growth}^f$, the integral defining $\chi_2(x,t)$ is well-defined. Note that,  the integral defining $\chi_2(x,0)$ is well-behaved at $y=0$ under condition $C_{\rm Holder}^f$. Indeed, for $||y||\leq \delta_x$, we have $\frac{|f(x+y)-f(x)|}{||y||^{d}}\leq K_x||y||^{-d+\alpha_x}$, which is integrable at $y=0$ in dimension $d$. Note that, if  $f(x+y)-f(x)$  were only  decaying as $const./\ln(||y||)$, then $|\chi_2(x,t)|\sim const.\ln(|\ln(t)|)\to+\infty$, when $t\to 0$, and  $\chi_2(x,0)$ would not exist (see the end of this section where we relax the local Hölder condition). 

From now, we  denote 
\begin{equation}\label{kappa}
	\kappa(x)\coloneqq\chi_2(x,0)=\int \rho(x+y)\frac{f(x+y)-f(x)}{||y||^{d}}\,d^dy.
\end{equation}
Also note that $\kappa(x)=0$ is possible, even if $f$ is not constant. For instance, if $\Omega$ is a sphere centered at $x$ or $\Omega=\R^d$, if $\rho(x+y)=\hat\rho(||y||)$ is isotropic around $x$ and, if $f_x:y\mapsto f(x+y)$ is an odd function of $y$, then we indeed have $\kappa(x)=0$ at the symmetry point $x$.

\vskip 0.2cm
\textit{Upper bound for} ${\cal B}_1(x)$

For $\varepsilon>0$, we define $\lambda$ like in  section~\ref{prooftheo1} and define $\eta=\min(\lambda,\delta_x)$, so that 
\begin{eqnarray}\label{boundchi1}
	\chi_1(x,t)&\leq &(1+\varepsilon)K_x \rho(x)\int_{||y||\leq\eta} {\rm e}^{-\frac{t}{||y||^d}}||y||^{2 (\alpha_x-d)}\,d^dy+\Lambda_x,\\
	\Lambda_x &=&\int_{||y||\geq\eta} \rho(x+y)\frac{(f(x+y)-f(x))^2}{||y||^{2 d}}\,d^dy,
\end{eqnarray}
where the constant $\Lambda_x<\infty$ under condition $C_{\rm Growth}^f$. The integral in \eq{boundchi1}, can be written as
\begin{eqnarray}\label{boundchi2}
    \int_{||y||\leq\eta} {\rm e}^{-\frac{t}{||y||^d}}||y||^{2 (\alpha_x-d)}\,d^dy &=& S_d\int_0^\eta {\rm e}^{-\frac{t}{r^d}}r^{2 \alpha_x-d-1}\,dr,\\
        &=&V_d t^{\frac{2\alpha_x}{d}-1} \int_{\frac{t}{\eta^d}}^{+\infty}u^{-\frac{2\alpha_x}{d}}{\rm e}^{-u}\,du,
\end{eqnarray}
Hence, we find that $\chi_1(x,t)$ is bounded for $\alpha_x> d/2$. For  $\alpha_x< d/2$, and for $t<t_1$ small enough, there exists a constant $M(2\alpha_x/d)$ so that $\chi_1(x,t)\leq M(2\alpha_x/d) t^{\frac{2\alpha_x}{d}-1}$. Finally, in the marginal case $\alpha_x=d/2$ and for $t<t_1$, we have $\chi_1(x,t)\leq M(1)\ln(1/t)$, for some constant $M(1)$.

Now, exploiting again the upper bound of $\psi(x,t)$ obtained in section~\ref{prooftheo1} and repeating the steps to bound the integrals involving $\psi^n(x,t)$, we find that, for $\alpha_x\neq d/2$,  ${\cal B}_1(x)$ is bounded up to a multiplicative constant by
\begin{eqnarray}\label{boundb1fin}
\int_0^{t_1} t^{\min\left(1,{\frac{2\alpha_x}{d}}\right) }\, {\rm e}^ {-n V_d\rho(x) t\ln\left( \frac{D_-}{t}\right) } \,dt\underset{n\to +\infty }{\sim} M'(2\alpha_x/d)
\left(V_d\rho(x)n\ln(n)\right)^{-\min\left(2,{\frac{2\alpha_x}{d}}+1\right) },
\end{eqnarray}
where $M'(2\alpha_x/d)$ is a constant depending only on $2\alpha_x/d$. In the marginal case, $\alpha_x=d/2$,
${\cal B}_1(x)$ is bounded up to a multiplicative constant by $n^{-2}\ln(n)$.

In summary, we find that 
\begin{equation}\label{B1case}
(n+1) {\cal B}_1(x)=\left\{\begin{array}{lr}
        O\left(n^{-\frac{2\alpha_x}{d}}(\ln(n))^{-1-\frac{2\alpha_x}{d}}\right), & \text{for } d>2\alpha_x\vspace{0.25cm}\\ 
       O\left(n^{-1}(\ln(n))^{-1}\right), & \text{for }d=2\alpha_x\vspace{0.35cm}\\
        O\left(n^{-1}(\ln(n))^{-2}\right), & \text{for } d<2\alpha_x
        \end{array}\right.
\end{equation}

\vskip 0.2cm
\textit{Asymptotic equivalent for} ${\cal B}_2(x)$

Let us first assume that $\kappa(x)=\chi_2(x,0)\neq 0$. Then again, as shown in detail in section~\ref{prooftheo1}, the integral defining ${\cal B}_2(x)$ is dominated by the small $t$ region, and will be asymptotically equivalent to
\begin{eqnarray}
   {\cal B}_2(x)&=&\int_0^{+\infty}t\,\psi^{n-1}(x,t)\chi_2^2(x,t)\,dt,\\
   &\underset{n\to +\infty }{\sim}&\kappa^2(x) \int_0^{t_1} t\, {\rm e}^ {-n V_d\rho(x) t\ln\left( \frac{D_\pm}{t}\right) } \,dt,\\
   &\underset{n\to +\infty }{\sim}& \left(\frac{\kappa(x)}{V_d\rho(x)n\ln(n)}\right)^{2}.
\end{eqnarray}
On the other hand, if $\kappa(x)=0$, one can bound $\chi_2(x,t)$ (up to a multiplicative constant) for $t\leq t_1$ by the integral
\begin{eqnarray}\label{boundchi20}
	\int_{||y||\leq\eta}\left( 1-{\rm e}^{-\frac{t}{||y||^d}}\right)||y||^{\alpha_x-d}\,d^dy
	&=& S_d\int_0^\eta\left( 1-{\rm e}^{-\frac{t}{r^d}}\right)r^{\alpha_x-d}\,r^{d-1}\,dr,\\
	&=& V_d t^{\frac{\alpha_x}{d}}\int_{\frac{t}{\eta^d}}^{+\infty}u^{-1-\frac{\alpha_x}{d}}\left(1-{\rm e}^{-u}\right)\,du.
\end{eqnarray}
Hence, for $\kappa(x)=0$, we find that 
\begin{equation}\label{B2case}
n(n+1){\cal B}_2(x)=O\left(n^{-\frac{2\alpha_x}{d}}(\ln(n))^{-2-\frac{2\alpha_x}{d}}\right).
\end{equation}

\vskip 0.2cm
\textit{Asymptotic equivalent for the bias term} ${\cal B}(x)$

In the generic case $\kappa(x)\neq 0$, we find that $(n+1){\cal B}_1(x)$ is always dominated by $n(n+1){\cal B}_2(x)$, and we find the following asymptotic equivalent for ${\cal B}(x)=(n+1){\cal B}_1(x)+n(n+1){\cal B}_2(x)$:
\begin{equation}\label{Bfin1}
    {\cal B}(x)\underset{n\to +\infty  }{\sim}\left(\frac{\kappa(x)}{V_d\rho(x)\ln(n)}\right)^{2}.
\end{equation}
In the non-generic case $\kappa(x)=0$, the bound for $(n+1){\cal B}_1(x)$ in \eq{B1case} is always more stringent than the bound for $n(n+1){\cal B}_2(x)$ in \eq{B2case}, leading to
\begin{equation}\label{Bcase}
{\cal B}(x)=\left\{\begin{array}{lr}
        O\left(n^{-\frac{2\alpha_x}{d}}(\ln(n))^{-1-\frac{2\alpha_x}{d}}\right), & \text{for } d>2\alpha_x\vspace{0.25cm}\\ 
       O\left(n^{-1}(\ln(n))^{-1}\right), & \text{for }d=2\alpha_x\vspace{0.35cm}\\
        O\left(n^{-1}(\ln(n))^{-2}\right), & \text{for } d<2\alpha_x
        \end{array}\right.,
\end{equation}
which proves the statements made in Theorem~\ref{theobias}.

\vskip 0.2cm
\textit{Interpretation of the bias term ${\cal B}(x)$ for $\kappa(x)\neq 0$}

Here, we assume the generic case $\kappa(x)\neq 0$ and define $\bar f(x)=\E\left[\hat f(x)\right]$. We have
\begin{eqnarray}
\Delta(x)&:=&\E\left[\sum_{i=0}^n w_i(x)(f(x_i)-f(x)) \right]=\bar f(x)-f(x),\\
\bar f(x)&=&\E\left[\sum_{i=0}^n w_i(x)f(x_i)\right]=(n+1)\E\left[w_0(x)f(x_0)\right].\label{fbarintro}
\end{eqnarray}

By using another time \eq{trick}, we find that,
\begin{eqnarray}
   \Delta(x)
   &=&(n+1)\int_0^{+\infty}\psi^{n}(x,t)\chi_2(x,t)\,dt,\\
   &\underset{n\to +\infty }{\sim}&n\,\kappa(x) \int_0^{t_1} {\rm e}^ {-n V_d\rho(x) t\ln\left( \frac{D_\pm}{t}\right) } \,dt,\\
   &\underset{n\to +\infty }{\sim}& \frac{\kappa(x)}{V_d\rho(x)\ln(n)}.
\end{eqnarray}
Comparing this result to the one of \eq{Bfin1}, we find that the bias ${\cal B}(x)$ is asymptotically dominated by the square of the difference $\Delta^2(x)$ between $\bar f(x)=\E\left[\hat f(x)\right]$ and $f(x)$:
\begin{equation}
{\cal B}(x)\underset{n\to +\infty }{\sim}\left(\E\left[\hat f(x)\right]-f(x)\right)^2,
\end{equation}
a statement made in Theorem~\ref{theobias}.

\vskip 0.2cm
\textit{Relaxing the local Hölder condition}

We now only assume the condition $C_{\rm Cont.}^f$ that $f$ is continuous at $x$ (but still assuming the growth conditions). We can now define $\delta_x$ such that the ball $B(x,\delta)\subset\Omega^\circ$ and $||y||\leq \delta_x\implies |f(x+y)-f(x)|\leq\varepsilon$. Then, the proof proceeds as above, but by replacing $K_x$ by $\varepsilon$, $\alpha_x$ by 0, and by updating the bounds for $\chi_1(x,t)$ (for which this replacement is safe) and $\chi_2(x,t)$ (for which it is not). We now find that for $0<t\leq t_1$, with $t_1$ small enough
\begin{eqnarray}\label{chiboundcont}
	0\leq\chi_1(x,t)&\leq &\varepsilon(1+2\varepsilon)V_d\rho(x) t^{-1}                     ,\\
	|\chi_2(x,t)|&\leq &\varepsilon(1+2\varepsilon)V_d\rho(x) \ln\left( \frac{1}{t}  \right). 
\end{eqnarray}
As already mentioned below \eq{chi2}, where we provided an explicit  counterexample, we see that relaxing the local Hölder condition does not guarantee anymore that $\lim_{t\to 0 }|\chi_2(x,0)|<\infty$. With these new bounds, and carrying the rest of the calculation as in the previous sections, we ultimately find the following weaker result compared to \eq{Bfin1} and \eq{Bcase}:
\begin{equation}\label{Bfincont}
    {\cal B}(x)=o\left( \frac{1}{\ln(n)}\right),
\end{equation}
or equivalently, that for any $\varepsilon>0$, there exists a constant $N_{x,\varepsilon}$ such that, for $n\geq N_{x,\varepsilon}$, we have
\begin{equation}\label{Bfincontbis}
    {\cal B}(x)\leq\frac{\varepsilon}{\ln(n)},
\end{equation}
which proves Theorem~\ref{theobiasbis}.

\textit{The bias term at a point where $\rho(x)=0$}

This section aims at proving Theorem~\ref{theorhonon0} expressing the lack of convergence of the estimator $\hat f(x)$ to $f(x)$, when $\rho(x)=0$, and under mild conditions.
Let us now consider a point $x\in\partial\Omega$ for which  $\rho(x)=0$, let us assume that there exists constants $\eta_x$, $\gamma_x>0$, and $G_x> 0$, such that $\rho$ satisfies the local Hölder condition at $x$
\begin{equation}
    ||y||\leq \eta_x\implies \rho(x+y)\leq G_x||y||^{\gamma_x}.
\end{equation}
We will also assume that the growth condition of \eq{growthf} is satisfied. With these two conditions, $\kappa(x)$ defined in \eq{kappa} exists. 
The vanishing of $\rho$ at $x$ strongly affects the behavior of $\psi(x,t)$ in the limit $t\to 0$, which is not singular anymore:
\begin{eqnarray}\label{nonsing}
1-\psi(x,t)&\underset{t\to 0 }{\sim}& t\int\rho(y)\|x-y\|^{-d} \,d^dy,
\end{eqnarray}
where the convergence of the integral $\lambda(x):=\int\rho(y)\|x-y\|^{-d} \,d^dy$ is ensured by the local Hölder condition of $\rho$ at $x$. 

Let us now evaluate $\bar f(x)=\lim_{n\to+\infty}\E[\hat f(x)]$, the expectation value of the estimator $\hat f(x)$ in the limit $n\to+\infty$, introduced in \eq{fbarintro}. First assuming, $\kappa(x)=\chi_2(x,0)\neq 0$, we obtain
\begin{eqnarray}
   \bar f(x)- f(x)
   &=&\lim_{n\to+\infty} (n+1)\int_0^{+\infty}\psi^{n}(x,t)\chi_2(x,t)\,dt,\\
   &=&\lim_{n\to+\infty} n\,\chi_2(x,0)\int_0^{t_1}{\rm e}^{n\,t\,\partial_t \psi(x,0)}\,dt,\\
   &=& \frac{\kappa(x)}{\lambda(x)},\label{biasnon0}
\end{eqnarray}
which shows that the bias term does not vanish in the limit $n\to+\infty$. \eq{biasnon0} can be straightforwardly shown to remain valid when $\kappa(x)=0$. Indeed, for any $\varepsilon>0$ chosen arbitrarily small, we can choose $t_1$ small enough such that $|\chi_2(x,t)|\leq\varepsilon$ for $0\leq t\leq t_1$, which leads to $|\bar f(x)- f(x)|\leq \varepsilon/\lambda(x)$.

Note that relaxing the local Hölder condition for $\rho$ at $x$ and only assuming the continuity of $f$ at $x$ and $\kappa(x)\neq 0$ is not enough to guarantee that $\bar f(x)\neq  f(x)$. For instance, if $\rho(x+y)\sim_{y\to 0} \rho_0/\ln(1/||y||)$, and there exists a local solid angle $\omega_x>0$ at $x$,  one can show that $1-\psi(x,t)\sim_{t\to 0} \omega_x S_d\rho_0\,t\ln(\ln(1/t))$, and the bias
would still vanish in the limit $n\to+\infty$, with $\hat f(x)- f(x)\sim_{n\to+\infty} \kappa(x)/[\omega_x S_d\rho_0\ln(\ln(n))]$.

\subsection{Asymptotic equivalent for the regression risk}\label{prooftheorisk}

This section aims at proving Theorem~\ref{regressionrisk}. Under conditions $C_{\rm Growth}^\sigma$, $C_{\rm Growth}^f$, and $C_{\rm Cont.}^f$, the results of \eq{varfin} and  \eq{Bfincont} show that for $\rho(x)\sigma^2(x)>0$ and $\rho$ and $\sigma^2$ continuous at $x$, the bias term ${\cal B}(x)$ is always dominated by the variance term ${\cal V}(x)$ in the limit $n\to+\infty$. Thus, the excess regression risk satisfies
\begin{equation}
    \E[(\hat{f}(x)-f(x))^2]\underset{n\to +\infty  }{\sim} \frac{\sigma^2(x)}{\ln(n)}.
\end{equation}

As a consequence, the Hilbert kernel estimate converges pointwise to the regression function in probability. Indeed, for $\delta>0$, there exists a constant $N_{x,\delta}$, such that 
\begin{equation}
    \E[(\hat{f}(x)-f(x))^2]\leq (1+\delta)\frac{\sigma^2(x)}{\ln(n)},
\end{equation}
for $n\geq N_{x,\delta}$. Moreover, for any $\varepsilon>0$, since $\E[(\hat{f}(x)-f(x))^2]\geq \varepsilon^2\,\P[|\hat{f}(x)-f(x)| \geq \varepsilon]$, we deduce the following Chebyshev bound, valid for $n\geq N_{x,\delta}$
\begin{equation}
\P[|\hat{f}(x)-f(x)| \geq\varepsilon]\leq \frac{1+\delta}{\varepsilon^2}\, \frac{\sigma^2(x)}{\ln(n)}.
\end{equation}

\subsection{Rates for the plugin classifier}\label{proofclassrisk}

In the case of binary classification $Y\in \{0,1\}$ and $f(x) = \P[Y=1 \mid X=x]$. Let $F \colon \R^d \to \{0,1\}$ denote the Bayes optimal classifier, defined by $F(x) := \theta(f(x) - 1/2)$ where $\theta(\cdot)$ is the Heaviside theta function. This classifier minimizes the risk $\risk(h) := \E[\ind{h(X) \neq Y}] = \P[h(X) \neq Y]$ under zero-one loss. Given the regression estimator $\hat{f}$, we consider the plugin classifier $\hat{F} (x)=\theta(\hat f(x)-\frac{1}{2})$, and we will exploit the fact that 
\begin{equation}\label{ineqclass}
0\leq \E[\risk(\hat F(x))] - \risk(F(x)) \leq 2\,\E[|\hat f(x) - f(x)|]\leq 2\sqrt{\E[(\hat f(x) - f(x))^2]}
\end{equation}

\vskip 0.2cm
\textit{Proof of \eq{ineqclass}}

For the sake of completeness, let us briefly prove the result of \eq{ineqclass}. The rightmost inequality is simply obtained from the Cauchy-Schwartz inequality, and we hence focus on proving the first inequality. Obviously, \eq{ineqclass} is satisfied for $f(x)=1/2$, for which 
$\E[\risk(\hat F(x))] = \risk(F(x))=1/2$.

If $f(x)>1/2$, we have $F(x)=1$, $\risk(F(x))=1-f(x)$,  and
\begin{eqnarray}\label{hatr1}
\E[\risk(\hat F(x))]&=&f(x)\P[\hat f(x)\leq 1/2]+(1-f(x))\P[\hat f(x)\geq 1/2],\\
&=&\risk(F(x)) +(2f(x)-1)\P[\hat f(x)\leq 1/2],
\end{eqnarray}
which implies $\E[\risk(\hat F(x))]\geq\risk(F(x))$.
Since $\P[\hat f(x)\leq 1/2]=\E[\theta(1/2-\hat f(x))]$, and using $\theta(1/2-\hat f(x))\leq \frac{|\hat f(x)-f(x)|}{f(x)-1/2}$, valid for any $1/2<f(x)\leq 1$, we readily obtain \eq{ineqclass}.

Similarly, in the case $f(x)<1/2$, we have $F(x)=0$, $\risk(F(x))=f(x)$,  and
\begin{equation}\label{hatr2}
\E[\risk(\hat F(x))]=\risk(F(x)) +(1-2f(x))\P[\hat f(x)\geq 1/2].
\end{equation}
Since $\P[\hat f(x)\geq 1/2]=\E[\theta( \hat f(x)-1/2)]$, and using $\theta(\hat f(x)-1/2)\leq \frac{|\hat f(x)-f(x)|}{1/2-f(x)}$, valid for any $0\leq f(x)<1/2$, we again obtain \eq{ineqclass} in this case.

In fact, for any $\alpha>0$, the inequalities  $\theta(1/2-\hat f(x))\leq \left(\frac{|\hat f(x)-f(x)|}{f(x)-1/2}\right)^\alpha$ and $\theta(\hat f(x)-1/2)\leq \left(\frac{|\hat f(x)-f(x)|}{1/2-f(x)}\right)^\alpha$ hold, respectively, for $f(x)>1/2$ and $f(x)<1/2$. Combining this remark with the use of the Hölder inequality leads to
\begin{eqnarray}\label{ineqclassgen}
\E[\risk(\hat F(x))] - \risk(F(x)) &\leq& 2|f(x)-1/2|^{1-\alpha}\,\E\left[|\hat f(x) - f(x)|^\alpha\right],\\
&\leq& 2|f(x)-1/2|^{1-\alpha}\, {\E\left[|\hat f(x) - f(x)|^{\frac{\alpha}{\beta}}\right]^\beta},
\end{eqnarray}
for any $0<\beta\leq 1$. In particular, for $0<\alpha<1$ and $\beta=\alpha/2$, we obtain
\begin{equation}\label{ineqclassalpha}
    0\leq \E[\risk(\hat F(x))] - \risk(F(x)) \leq 2|f(x)-1/2|^{1-\alpha}\, {\E\left[|\hat f(x) - f(x)|^{2}\right]^\frac{\alpha}{2}}.
\end{equation}
The interest of this last bound compared to the more classical bound of \eq{ineqclass} is to show explicitly the cancellation of the classification risk as $f(x)\to 1/2$, while still involving the regression risk $\E\left[|\hat f(x) - f(x)|^{2}\right]$ (to the power $\alpha/2<1/2$).

\vskip 0.2cm
\textit{Bound for the classification risk}

Now exploiting the results of section~\ref{prooftheorisk} for the regression risk, and the two inequalities  \eq{ineqclass} and \eq{ineqclassalpha}, we readily obtain Theorem~\ref{theoclass}.

\subsection{Extrapolation behavior outside the support of $\rho$}\label{proofout}
This section aims at proving Theorem~\ref{theoout} characterizing the behavior of the regression estimator $\hat f$ outside the closed support $\bar\Omega$ of $\rho$ (extrapolation).

\vskip 0.2cm
\textit{Extrapolation estimator in the limit $n\to\infty$}

We first assume the growth condition $\int \rho(y)\frac{|f(y)|}{1+\|y\|^{d}}\,d^dy<\infty$.
For $x\in \R^d$ (i.e., not necessarily in $\Omega$), we have quite generally
\begin{equation}\label{outside}
   \E\left[\hat f(x)\right]= (n+1)\E\left[w_0(x) f(x)\right]= (n+1)\int_0^{+\infty}\psi^{n}(x,t)\chi(x,t)\,dt,\\
\end{equation}   
where $\psi(x,t)$ is again given by \eq{psi} and 
\begin{equation}\label{chi}
	\chi(x,t)\coloneqq\int \rho(x+y)f(x+y)\frac{{\rm e}^{-\frac{t}{||y||^d}}}{||y||^{d}}\,d^dy,
\end{equation}
which is finite for any $t>0$, thanks to the above growth condition for $f$.

Let us now assume that the point $x$ is not in the closed support $\bar\Omega$ of the distribution $\rho$ (which excludes the case  $\Omega=\R^d$ ). Since the integral in \eq{outside} is again dominated by its  $t\to 0$ behavior, we have to evaluate $\psi(x,t)$ and $\chi(x,t)$ in this limit, like in the different proofs above. In fact, when $x\notin \bar\Omega$, the integral defining $\psi(x,t)$ and $\chi(x,t)$ are not singular anymore, and we obtain 
\begin{eqnarray}\label{smallt}
1-\psi(x,t)&\underset{t\to 0 }{\sim}& t\int\rho(y)\|x-y\|^{-d} \,d^dy,\\
\chi(x,0)&=&\int{\rho(y)f(y)\|x-y\|^{-d}} \,d^dy.
\end{eqnarray}
Note that  $\psi(x,t)$ has the very same linear behavior as in \eq{nonsing}, when we assumed $x\in\partial\Omega$ with $\rho(x)=0$, and a local Hölder condition for $\rho$ at $x$.

Finally, by using the same method as in the previous sections to evaluate the integral of \eq{outside} in the limit $n\to +\infty $, we obtain 
\begin{eqnarray}
\int_0^{+\infty}\psi^{n}(x,t)\chi(x,t)\,dt&\underset{n\to +\infty  }{\sim}&\chi(x,0)\int_0^{t_1}{\rm e}^{n\,t\,\partial_t \psi(x,0)}\,dt,\\
&\underset{n\to +\infty  }{\sim}&\frac{1}{n}\frac{\chi(x,0)}{|\partial_t \psi(x,0) |},
\end{eqnarray}
which leads to the first result of Theorem~\ref{theoout}:
\begin{equation}
\hat f_\infty(x):=\lim_{n\to+\infty} \E\left[\hat f(x)\right]=\frac{\int{\rho(y)f(y)\|x-y\|^{-d}} \,d^dy}{\int\rho(y)\|x-y\|^{-d} \,d^dy}.
\end{equation}
Note that since the function $(x,y)\longmapsto \|x-y\|^{-d}$ is continuous at all points $x\notin\bar\Omega$, $y\in\Omega$, and thanks to the absolute convergence of the integrals defining  $\hat f_\infty(x)$, standard methods show that $\hat f_\infty$ is continuous (in fact, infinitely differentiable) at all  $x\notin\bar\Omega$.

\vskip 0.2cm
\textit{Extrapolation  far from $\Omega$}

Let us now investigate the behavior of $\hat f_\infty(x)$ when the distance $L:=d(x,\Omega)=\inf\{||x-y||,~y\in\Omega\}>0$ between  $x$ and $\Omega$ goes to infinity, which can only happen for certain $\Omega$, in particular, when $\Omega$ is bounded. We now assume the stronger condition, $\langle |f|\rangle:= \int\rho(y)|f(y)|\,d^dy<\infty$, such that the $\rho$-mean of $f$, $\langle f\rangle:= \int\rho(y)f(y)\,d^dy$, is finite. We consider a point $y_0\in\Omega$, so that $ ||x-y_0||\geq L> 0$, and we will exploit the following inequality, valid for any $y\in \Omega$ satisfying $ ||y-y_0||\leq R$, with $R>0$:
\begin{equation}
0\leq 1-\frac{L^d}{||x-y||^d}\leq\frac{||x-y||^d-L^d}{L^d}\leq\frac{(L+R)^d-L^d}{L^d}\leq{\rm e}^\frac{dR}{L}-1.
\end{equation}

Now, for a given $\varepsilon>0$, there exist $R>0$ large enough such that $\int_{\|y-y_0\|\geq R}\rho(y)\,d^dy\leq\varepsilon/2 $ and $\int_{\|y-y_0\|\geq R}\rho(y)|f(y)|\,d^dy\leq\varepsilon/2 $. Then, for such a $R$, we consider $L$ large enough such that the above bound satisfies ${\rm e}^\frac{dR}{L}-1\leq \varepsilon\min(1/ \langle |f|\rangle,1)/2$. We then obtain
\begin{eqnarray}
\left|  L^d\int{\rho(y)f(y)\|x-y\|^{-d}\,d^dy}- \langle f\rangle\right|&\leq& 
\left({\rm e}^\frac{dR}{L}-1\right) \int_{||y-y_0||\leq R}\rho(y)|f(y)|\,d^dy\\ 
\quad &&+ \int_{\|y-y_0\|\geq R}\rho(y)|f(y)|\,d^dy,\\
&\leq& \frac{\varepsilon}{2\langle |f|\rangle}\times \langle |f|\rangle+\frac{\varepsilon}{2}\leq\varepsilon,
\end{eqnarray}
which shows that under the condition $\langle |f|\rangle<\infty$, we have
\begin{equation}
\lim_{d(x,\Omega)\to+\infty} d^d(x,\Omega)\int{\rho(y)f(y)\|x-y\|^{-d}\,d^dy}=\langle f\rangle.
\end{equation}
Similarly, one can show that
\begin{equation}
\lim_{d(x,\Omega)\to+\infty} d^d(x,\Omega)\int{\rho(y)\|x-y\|^{-d}\,d^dy}=\int \rho(y)\,d^dy=1.
\end{equation}
Finally, we obtain the second result of Theorem~\ref{theoout}, 
\begin{equation}
\lim_{d(x,\Omega)\to+\infty} \hat f_\infty(x)=\langle f\rangle.
\end{equation}

\vskip 0.2cm
\textit{Continuity of the extrapolation }

We now consider $x\notin\bar\Omega$ and $y_0\in\partial\Omega$, but such that $\rho(y_0)>0$ (i.e., $y_0\in \partial\Omega\cap\Omega$), and we note $l:=||x-y_0||>0$. 
We assume the continuity at $y_0$ of $\rho$ and $f$ as seen as functions restricted to $\Omega$, i.e., $\lim_{y\in\Omega\to y_0}\rho(y)=\rho(y_0)$ and $\lim_{y\in\Omega\to y_0}f(y)=f(y_0)$. Hence, for any $0<\varepsilon<1$, there exists $\delta>0$ small enough such that $y\in \Omega$ and $||y-y_0||\leq\delta\implies |\rho(y_0)-\rho(y)|\leq \varepsilon$ and $ |\rho(y_0)f(y_0)-\rho(y)f(y)|\leq \varepsilon$. Since we intend to take $l>0$ arbitrary small, we can impose $l<\delta/2$. 

We will also assume that $\partial\Omega$ is smooth enough near $y_0$, such that there exists a strictly positive local solid angle $\omega_0$ defined by
\begin{equation}\label{anglesoldef}
\omega_0=    \lim_{r\to 0}\frac{1}{V_d\rho(y_0)r^d}\int_{\|y-y_0\|\leq r}\rho(y)\,d^dy=
\lim_{r\to 0}\frac{1}{V_dr^d}\int_{y\in\Omega/\|y-y_0\|\leq r}\,d^dy,
\end{equation}
where the second inequality results from the continuity of $\rho$ at $y_0$ and the fact that $\rho(y_0)>0$.
If $y_0\in \Omega^\circ$, we have $\omega_0=1$, while for $y_0\in \partial\Omega$, we have generally $0\leq \omega_0\leq 1$. Although we will assume $\omega_0>0$ for our proof below, we note that $\omega_0=0$ or $\omega_0=1$ can happen for $y_0\in \partial\Omega$. For instance, we can consider $\Omega_0,\,\Omega_1\subset \R^2$ respectively defined by $\Omega_0=\{(x_1,x_2)\in \R^2 / x_1\geq 0,~|x_2|\leq x_1^2\}$ and $\Omega_1=\{(x_1,x_2)\in \R^2 / x_1\leq 0\}\cup\{(x_1,x_2)\in \R^2 / x_1\geq 0,~|x_2|\geq x_1^2\}$. Then, it is clear that the local solid angle at the origin $O=(0,0)$ is respectively $\omega_0=0$ and $\omega_0=1$. Also note that if $x$ is on the surface of a sphere or on the interior of a face of a hypercube (and in general, when the boundary near $x$ is locally a hyperplane; the generic case), we have $\omega_x=\frac{1}{2}$. If $x$ is a corner of the hypercube, we have  $\omega_x=\frac{1}{2^d}$.

Returning to our proof, and exploiting \eq{anglesoldef}, we  consider $\delta$ small enough such that for all $0\leq r\leq\delta$, we have 
\begin{equation}
\left|\int_{y\in\Omega/\|y-y_0\|\leq r}\,d^dy  -\omega_0V_d\,r^d   \right|\leq \varepsilon\,\omega_0V_d\,r^d.
\end{equation}

We can now use these preliminaries to obtain
\begin{eqnarray}
(\rho(y_0)f(y_0)-\varepsilon)J(x)-C\leq \int\rho(y)f(y)\|x-y\|^{-d}\,d^dy\leq (\rho(y_0)f(y_0)+\varepsilon)J(x)+C,\label{Ibound}\\
(\rho(y_0)-\varepsilon)J(x)-C'\leq \int\rho(y)\|x-y\|^{-d}\,d^dy\leq (\rho(y_0)+\varepsilon)J(x)+C',\label{Ipbound}
\end{eqnarray}
with
\begin{eqnarray}
J(x)&:=&\int_{y\in\Omega\,/\,||y-y_0||\leq \delta}\|x-y\|^{-d}\,d^dy,\\
C&=&\left(\frac{2}{\delta}\right)^2\int_{||y-y_0||\geq \delta}\rho(y)|f(y)|\,d^dy,\\
C'&=&\left(\frac{2}{\delta}\right)^2.
\end{eqnarray}

Let us now show that $\lim_{l\to 0}J(x)=+\infty$. We define $N:=[\delta/l]\geq 2$, where $[\,\bf{.}\,]$ is the integer part, and we have $N\geq 2$, since we have imposed $l<\delta/2$. For $n\in\N\geq 1$, we define, 
\begin{equation}
I_n:=\int_{y\in\Omega/||y-y_0||\leq \delta/n} \,d^dy,
\end{equation}
and note that we have
\begin{eqnarray}
I_n-I_{n+1}=\int_{\substack{y\in\Omega/||y-y_0||\leq \delta/n,\\||y-y_0||\geq \delta/(n+1)}} \,d^dy,\\
\left| I_n-\omega_0V_d\,\left(\frac{\delta}{n}\right)^d\right|\leq \varepsilon\,\omega_0V_d\,\left(\frac{\delta}{n}\right)^d.\label{boundin}
\end{eqnarray}
We can then write
\begin{eqnarray}
J(x)&\geq &\sum_{n=1}^N \frac{1}{\left(l+\frac{\delta}{n}\right)^d}(I_n-I_{n+1}),\\
&\geq &\sum_{n=1}^N \left(\frac{1}{\left(l+\frac{\delta}{n+1}\right)^d}-\frac{1}{\left(l+\frac{\delta}{n}\right)^d}\right)I_{n+1}+ \frac{I_1}{\left(l+{\delta}\right)^d}-\frac{I_{N+1}}{\left(l+\frac{\delta}{N+1}\right)^d}.
\end{eqnarray}
We have
\begin{eqnarray}
\frac{I_1}{\left(l+{\delta}\right)^d}-\frac{I_{N+1}}{\left(l+\frac{\delta}{N+1}\right)^d}&\geq &
\omega_0V_d\left((1-\varepsilon)\frac{1}{\left(1+\frac{l}{\delta}\right)^d}  -(1+\varepsilon) \frac{1}{\left(1+\frac{(N+1)l}{\delta}\right)^d} \right),\\
&\geq & \omega_0V_d\left((1-\varepsilon)\frac{2^d }{3^d}  -(1+\varepsilon)\right)=:C'',
\end{eqnarray}
which defines the constant $C''$.
Now using \eq{boundin}, $l<\delta/2$, $N=[\delta/l]$, and the fact that $(1+u)^d-1\geq d\,u$, for any $u\geq 0$, we obtain 
\begin{eqnarray}
J(x)&\geq &(1-\varepsilon)\,\omega_0V_d\sum_{n=1}^N \frac{1}{\left(1+\frac{(n+1)l}{\delta}\right)^d}
\left(\left(\frac{l+\frac{\delta}{n}}{l+\frac{\delta}{n+1}}\right)^d    -1  \right)+C'',\\
&\geq &(1-\varepsilon)\,\omega_0S_d\sum_{n=1}^N \frac{1}{\left(1+\frac{(n+1)l}{\delta}\right)^{d+1}}\frac{1}{n}+C'',\label{bestbound}\\
&\geq & \frac{(1-\varepsilon)\,\omega_0\,S_d}{\left(1+\frac{(N+1)l}{\delta}\right)^{d+1}}\ln(N-1)+C'',\\
&\geq & (1-\varepsilon)\,\omega_0\,\left(\frac{2}{5}\right)^{d+1}S_d\ln\left(\frac{\delta}{l}-2\right)+C''.
\end{eqnarray}
We hence have shown that  $\lim_{l\to 0}J(x)=+\infty$. Note that we can obtain an upper bound for $J(x)$ similar to \eq{bestbound} in a similar way as above, and with a bit more work, it is straightforward to show that we in fact have $J(x)\sim_{l\to 0}\omega_0\,S_d \ln\left(\frac{\delta}{l}\right)$, a result that we will not need here.

Now, using \eq{Ibound} and \eq{Ipbound} and the fact that $\lim_{l\to 0}J(x)=+\infty$, we find that
\begin{eqnarray}
    \int\rho(y)f(y)\|x-y\|^{-d}\,d^dy\underset{ l\to 0 }{\sim}\rho(y_0)f(y_0)J(x),\\
    \int\rho(y)\|x-y\|^{-d}\,d^dy\underset{ l\to 0 }{\sim}\rho(y_0)J(x),
\end{eqnarray}
for $f(y_0)\ne 0$ (remember that $\rho(y_0)> 0$), while for $f(y_0)=0$, we obtain $\int\rho(y)f(y)\|x-y\|^{-d}\,d^dy=o(J(x))$. 
Finally, we have shown that
\begin{eqnarray}
    \lim_{x\notin \bar\Omega, x\to y_0}   \hat f_\infty(x)=f(y_0),
\end{eqnarray}
establishing the continuity of the extrapolation and the last part of Theorem~\ref{theoout}.

\newpage

\end{appendices}



\end{document}